\documentclass{article}

\pdfoutput=1
\usepackage{subcaption}
\usepackage{rotating}

\PassOptionsToPackage{square,numbers}{natbib} 
\PassOptionsToPackage{
    colorlinks = true,
    linkcolor = black,
    urlcolor  = blue,
    citecolor = black
}{hyperref}

\usepackage[final]{neurips_2024}
\usepackage{amsmath}
\usepackage{subcaption}
\usepackage{graphicx} 
\usepackage{multirow}

\usepackage{doi} 

\usepackage{caption} 
\usepackage[utf8]{inputenc} 
\usepackage[T1]{fontenc}    
\usepackage{url}            
\usepackage{booktabs}       
\usepackage{amsfonts}       
\usepackage{nicefrac}       
\usepackage{microtype}      
\usepackage{xcolor}         
\usepackage{floatpag}
\usepackage{empheq}         

\newcommand{\Gc}{\mathcal{G}}
\newcommand{\Gctilde}{\Tilde{\mathcal{G}}}
\newcommand{\Gct}{\Gc^\theta}
\newcommand{\Gcttilde}{\Gctilde^\theta}
\newcommand{\R}{\mathbb{R}}

\title{
FUSE: Fast Unified Simulation and Estimation for PDEs
}

\author{
  Levi E. Lingsch \\
  Seminar for Applied Mathematics \& AI Center \\
  ETH Zurich \\
  \texttt{levi.lingsch@ai.ethz.ch} \\
  \And
  Dana Grund \\
  Institute for Atmospheric\\
  and Climate Science, ETH Zurich \\
  \texttt{dana.grund@ethz.ch} \\   \And
  Siddhartha Mishra \\
  Seminar for Applied Mathematics \& AI Center \\
  ETH Zurich \\
  \texttt{siddhartha.mishra@ethz.ch} \\
  \And
  Georgios Kissas \\
  AI Center  \\
  ETH Zurich \\
  \texttt{gkissas@ai.ethz.ch} \\
}

\begin{document}

\maketitle

\begin{abstract}
The joint prediction of continuous fields and statistical estimation of the underlying discrete parameters is a common problem for many physical systems, governed by PDEs. Hitherto, it has been separately addressed by employing operator learning surrogates for field prediction while using simulation-based inference (and its variants) for statistical parameter determination. Here, we argue that solving both problems within the same framework can lead to consistent gains in accuracy and robustness. To this end, We propose a novel and flexible formulation of the operator learning problem that allows jointly predicting continuous quantities and inferring distributions of discrete parameters, and thus amortizing the cost of both the inverse and the surrogate models to a joint pre-training step. We present the capabilities of the proposed methodology for predicting continuous and discrete biomarkers in full-body haemodynamics simulations under different levels of missing information. We also consider a test case for atmospheric large-eddy simulation of a two-dimensional dry cold bubble, where we infer both continuous time-series and information about the system conditions. We present comparisons against different baselines to showcase significantly increased accuracy in both the inverse and the surrogate tasks.
\end{abstract}

\section{Introduction}
\label{sec:intro}

Partial Differential Equations (PDEs) describe the propagation of system conditions for a very wide range of physical systems. Parametric PDEs consider different system conditions as well as an underlying solution operator characterized by a set of finite-dimensional parameters. Traditional numerical methods based on different discretization schemes such as Finite Differences, Finite Volumes, and Finite Elements have been developed along with fast and parallelizable implementations to tackle complex problems, such as atmospheric modeling and cardiovascular biomechanics. For parametric PDEs, these methods define maps from the underlying set of discrete parameters, which describe the dynamics and the boundary/initial conditions, to physical quantities such as velocity or pressure that are continuous in the spatio-temporal domain. Despite their successful application, there still exist well-known drawbacks of traditional solvers. To describe a particular physical phenomenon, PDE parameters and solvers need to be calibrated on precise conditions that are not known a priori and cannot easily be measured in realistic applications. Therefore, iterative and thus expensive calibration procedures are considered in the cases where the parameters and conditions are inferred from data \cite{arzani2022machine}. Even after the solvers are calibrated, an ensemble of solutions needs to be generated to account for uncertainties in the model parameters or assess the sensitivity of the solution to different parameters which are computationally prohibitive downstream tasks \cite{quarteroni2015reduced}. 

\textbf{Related work} A variety of deep learning algorithms have recently been proposed for scientific applications, broadly categorized into surrogate and inverse modeling algorithms, to either reduce the computational time of complex forward simulations or infer missing finite-dimensional information from data to calibrate a simulator to precise conditions. 

\emph{Surrogate learning} is a paradigm for accelerating computations. Hence, the cost of evaluating continuous quantities is amortized to an offline training stage.
For functional data, so-called \emph{Neural Operators} \cite{kovachki2023neural,bartolucci2024representation}, generalize across different conditions and discretizations of a complex system (resolution invariance). A multitude of operator learning approaches has been designed \citep{Lu_2021,kissas2022learning,seidman2022nomad,jin2022mionet,raonić2023convolutional,li2023scalable,li2023transformer,hao2023gnot}, such as the Fourier Neural Operator (FNO) \cite{li2021fourier}. However, a priori, neural operators are not designed to process maps between finite-dimensional and continuous quantities.
A different surrogate modeling strategy is constructed by learning a map between parameters of PDEs and solutions, as in traditional Reduced Order Model (ROM) approaches \citep{halder2020deep, chen2023crom, hijazi2023pod}, and their deep learning counterparts \cite{kim2019deep, garom}, including Generative Adversarial ROM (GAROM) \cite{garom}.
In addition, Gaussian Processes and kriging have been re-visited in the context of PDE emulation \cite{gulian2020gaussian,xiong2021clustered}.

Amortizing the cost of parameter inference has also been widely explored in the literature of generative modeling and Simulation Based Inference (SBI). In SBI, Invertible Neural Networks are combined with discrete Normalizing Flows to develop Neural Posterior Estimation \citep{dinh2015nice, dinh2017density, kingma2018glow, rezende2016variational} or continuous Normalizing Flows in Flow Matching Posterior Estimation (FMPE) \citep{dax2023flow, lipman2023flow}. However, they commonly rely on an expensive physical solver to sample continuous predictions, and are hindered by the use of approximate physical models, Gaussian distributions, or a limited sample size. 

Approaches that attempt to \emph{jointly infer parameters and learn a surrogate} are much rarer in the literature. Many are built on Variational Autoencoders (VAEs) \cite{tong2023invaert}, in finite \cite{kingma2022autoencoding} or infinite \cite{seidman2023variational} dimensions. They inherently assume an underlying bijective relation between the continuous and discrete quantities or are restricted to Gaussian latent representations, which limits their applicability to nonlinear problems. Among them, InVAErt \citep{tong2023invaert} extends a deterministic encoder-decoder pair with a variational latent encoder. However, InVAErt is primarily constructed for investigating identifiability of parameters in systems of PDEs, and may not be suitable for learning uncertainty propagation from parameters to continuous fields. 
Making use of the resolution invariance of neural operators, the conditional Variational Neural Operator (cVANO) \cite{kissas2023towards}, based on \cite{seidman2023variational}. cVANO relies on a VAE to learn a low-dimensional manifold of the system constrained by finite-dimensional parameters, but is restricted to Gaussian approximations which are not suitable for problems which lack bijectivity.
As alternative approaches, Simformer \cite{gloeckler2024allinone} estimates both finite-dimensional and function-valued parameters using transformers within the SBI framework, and PAGP \cite{zhang2022physics} leverages the expressive power of Gaussian Processes. However, both are not suited for functional input due to quadratic growth with the input size.
Finally, OpFlow \citep{shi2024universal} brings together operator learning and normalizing flows for quantifying uncertainty in output functions, but it lacks interpretability and inference on the latent space.

\textbf{Our contribution} 
Based on the experience and limitations of the aforementioned approaches, we formulate the following requirements for a unified forward-inverse framework relating functional quantities to a known space of finite-dimensional parameters: First, the interpretability given by the given parameterization should be retained in both the forward and inverse tasks as well as their combination (A). Then, both the forward and inverse model should be discretization invariant with respect to all functional variables (B). Finally, a general probabilistic representation should be chosen for the latent parameter space, allowing for arbitrary parameter distributions (C).
Following these guidelines, the main contributions of our paper are the following.

\begin{figure}[h]
    \centering
    \includegraphics[width=0.8\textwidth]{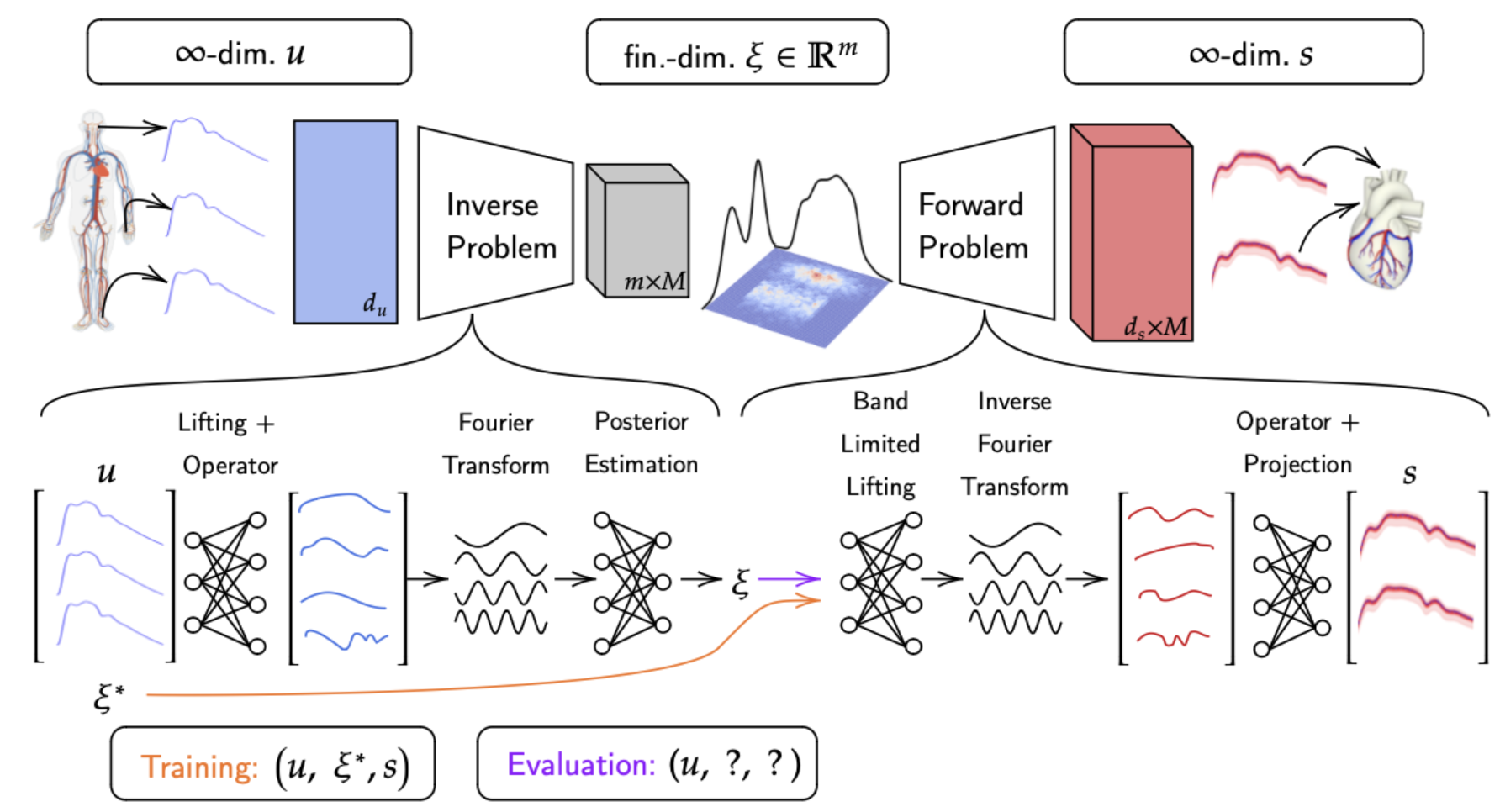}
    \captionsetup{font=normal}
    
    \caption{FUSE models a posterior distribution over finite-dimensional parameters $\xi$ given infinite-dimensional functions $u$ with $d_u$ components (channels). It learns other continuous functions $s$ with $d_s$ channels from parameters $\xi$. Band-limited Fourier transforms and a lifting operator act as a bridge between finite and infinite dimensions for the forward problem. Likewise, as inference models such as FMPE or NPE require fixed-size inputs, the operator layers are conjoined with a band-limited Fourier transform to learn a fixed-size representation of the input function.
    }
    \label{fig:FUSE main diagram}
\end{figure}

\begin{itemize}
    \item We propose \emph{Fast Unified Simulation and Estimation for PDEs} (FUSE), a rigorous framework for unified inverse and forward problems for parametric PDEs. The FUSE framework, illustrated in Figure \ref{fig:FUSE main diagram}, allows to combine different forward and inverse models, and we choose to instantiate it with the FNO and FMPE for the experiments in this work.
    \item We formulate a mathematical framework with a \emph{unified objective} for the forward and inverse problems, which allows us to assess how uncertainty in the inverse problem propagates through the forward problem (\emph{propagated uncertainty}) and evaluate both models together to avoid nonlinear error amplification at model concatenation.
    \item We showcase how to \emph{adapt forward surrogate and inverse estimation models} to fit the FUSE framework, mapping between finite and infinite-dimensional spaces. In particular, we extend Fourier Neural Operators (FNO) with a custom lifting to take finite-dimensional inputs, and we equip Flow-Matching Posterior Estimation (FMPE) with an FNO-based encoding to allow for functional conditional information.
    \item We show that our implementation of FUSE overcomes struggles of baselines (cVANO, GAROM, InVAErt) and ablations (U-Net) on two complex and realistic PDE examples, pulse wave propagation (PWP) in the human arterial network and an atmospheric cold bubble (ACB). The experiments exemplify its ability for accurate and fast surrogate modeling, parameter inference, out-of-distribution generalization, and the flexibility to handle different levels of input information.
\end{itemize}
To our knowledge, the extensions we present to FNO and FMPE have not been explored in the literature yet. We would like to emphasize that we chose to base ourselves on these methods to illustrate the FUSE framework, but other forward and inverse methods may be more suitable for particular test cases.

\section{Methods}

\paragraph{Notation and Assumptions}
Let $\mathcal{U}\subset \mathcal{C}(X, \R^{d_u})$ and $\mathcal{S}\subset \mathcal{C}(Y, \R^{d_s})$ be spaces of continuous functions on compact domains $X \subseteq \R^d$ and $Y \subseteq \R^{d'}$, respectively, and let $\Xi \subseteq \R^m$ be a space of finite-dimensional parameters.
For a map $F: \mathcal{A}\rightarrow \mathcal{B}$, where we take $\mathcal{A}$ and $\mathcal{B}$ to be among the spaces $\mathcal{U}$, $\mathcal{S}$, or $\Xi$, and a probability measure $\pi\in\text{Prob}(\mathcal{A})$, we denote by $F_{\#\pi}(B)\in\text{Prob}(\mathcal{B})$ the push-forward measure that expresses uncertainty on a set $B\subseteq \mathcal{B}$ by the propagation of $\pi$ through $F$, defined as $F_{\#\pi}(B) = \pi(F^{-1}(B))$. Further, we assume all metrics on measures in this paper to be the total variation metric, and denote them by $d$ both on $\text{Prob}(\mathcal{S})$ and $\text{Prob}(\Xi)$. In general, we assume all maps to be Lipschitz continuous, and we omit Lipschitz constants in inequalities (details are provided in the Appendix).

\paragraph{Problem Formulation} 
Consider the setting of a parametric PDE with forward solution mapping $\Gc:\xi \mapsto s$, where $\xi \in \Xi$ is a vector of finite-dimensional parameters and $s \in \mathcal{S}$ is a function-valued output. Since it is the goal to calibrate the input parameters, we omit any functional inputs to $\Gc$, such as initial or boundary conditions, and assume they are kept constant or encoded in the parameters $\xi$.
Given the forward operator $\Gc$, the \emph{forward model uncertainty} is quantified by propagating a given distribution of parameters $\rho \in {\rm Prob}(\Xi)$ through the model, i.e. evaluating the push-forward measure $\Gc_{\#\rho}$. In practice, the parameters $\xi$ are not available and need to be inferred from indirect measurements $u \in \mathcal{U}$, which we assume to be functions in space or time (e.g., time series), and, in general, $u \neq s$. The extended solution operator $\Gctilde: u \mapsto s$ then maps between functional measurements and functional model output, omitting the intermediate parameter space. Since the measurements $u$ are impacted by the uncertain parameters $\xi$, the function inputs $u$ are as well equipped with uncertainty and represented by the measure $\mu \in {\rm Prob}(\mathcal{U})$. Finally, the inverse problem consists in estimating the distribution $\rho(\xi|u)$ of parameters given measurements $u$. The corresponding uncertainty on the predicted outcomes $s$ is then given by the \emph{propagated uncertainty} $\Gc_{\#\rho(.|u)}$.

\paragraph{Unified Objective}
Given an approximate forward operator $\Gcttilde \approx \Gctilde$ between function spaces, and an estimated distribution $\mu^\phi\approx \mu$, parameterized by $\theta$ and $\phi$, respectively, we use triangle inequality to observe that 
\begin{equation}
\label{eq:triangle inequality}
    d(\Gcttilde_{\#\mu^\phi}, \Gctilde_{\#\mu}) \leq 
    \underbrace{d(\Gcttilde_{\#\mu^\phi}, \Gcttilde_{\#\mu})}_{\text{Measure matching}} +  
    \underbrace{d(\Gcttilde_{\#\mu}, \Gctilde_{\#\mu})}_{\text{Operator learning}}.
\end{equation}
Thus, we found a unified objective consisting of two steps corresponding to the two terms in the right hand side (rhs) of Equation \eqref{eq:triangle inequality}. In the measure matching step, the objective amounts to learn an approximation $\mu^\phi$ of the measure $\mu$, whereas in the operator learning step, the objective is to learn a Neural Operator $\Gct$ that approximates the underlying ground truth operator $\Gc$.

Our reformulation of operator learning naturally fits into the aim of this paper to propose a method that can act as an operator surrogate (\emph{forward problem}, operator learning objective) as well as performing parameter inference by minimizing distances on measures (\emph{inverse problem}, measure matching objective), unifying the apparently unrelated problems of surrogate modeling and inference. 

\paragraph{Forward Problem: Operator Learning}
The operator $\Gctilde$ resembles the solution operator to a PDE that maps function inputs such as initial and boundary conditions, coefficients, sources etc to (observables of) the solution of the PDE. It is the goal of supervised operator learning \cite{LMK1} to learn this type of operator as a parameterized Neural Operator (NO) $\Tilde{\mathcal{G}}^{\theta}$ on a (training) distribution $\mu\in\text{Prob}(\mathcal{U})$ by minimizing the operator learning objective in \eqref{eq:triangle inequality}. This objective is bound by the \emph{supervised learning} objective on the parameter space $\Xi$,
\begin{equation}
\label{eq:operator objective}
    d(\Gcttilde_{\#\mu}, \Gctilde_{\#\mu}) \leq d(\Gct_{\#\rho}, \Gc_{\#\rho}),
\end{equation}
and further by the corresponding objective (Appendix \ref{sec:appendix derive L1}),
\begin{equation}
\label{eq:operator loss}
    \mathcal{L}_1(\theta) = \int_{\Xi} \| \Gct(\xi) - \Gc(\xi)\|_{L^1} d \rho(\xi).
\end{equation}
The loss is approximated by training samples of the form $\{ \xi^i, \mathcal{G}(\xi^i)\}_{i=1}^N$, sampled from an underlying data distribution $\rho \in {\rm Prob}(\Xi)$.
In order to use a NO on finite-dimensional inputs, we define the \emph{band-limited lifting} $h^{\theta_3} (\xi)$, composed of a lifting increasing the dimension of the parameters and an inverse Fourier transform. We then choose to instantiate the $\mathcal{G}^{\theta}$ based on Fourier Neural Operator (FNO) \cite{li2021fourier} layers $\mathcal{K}^{\theta_2}$, implemented with discrete spectral evaluations as in \cite{lingsch2024regular} to handle irregularly sampled measurements for the output function,
\begin{equation}
    \mathcal{G}^{\theta} (\xi)  =\mathcal{Q}^{\theta_1} \circ \mathcal{K}^{\theta_2} \circ h^{\theta_3} (\xi),
\end{equation}
where $\mathcal{Q}^{\theta_1}$ is a learnable map that projects the channels to the dimensions of the output function and $\theta = [\theta_1, \theta_2, \theta_3]$ are the trainable parameters.

\paragraph{Inverse Problem: Measure Matching}
We would like to approximate the true posterior measure $\rho(\xi|u)$ given measurements $u$ by $\rho^\phi(\xi|u)$.
This task is equivalent to minimizing the measure matching objective in \eqref{eq:triangle inequality} since (Appendix \ref{sec:appendix derive L2})
\begin{equation}
     d (\Gcttilde_{\#\mu^\phi}, \Gcttilde_{\#\mu}) \leq  d (\rho^\phi(\xi|u), \rho(\xi |u)).
     \label{eq:inverse}
\end{equation}
To minimize the latter distance, we adapt Flow-Matching Posterior Estimation (FMPE) to handle function-valued conditional inputs $u$. FMPE trains a flow function that maps samples from a standard normal distribution to those from the target distribution $\rho(\xi|u)$. Unlike diffusion models \cite{Sohl2015diffusion, Ho2020DDPM, wang2023conditional}, which require multiple steps, FMPE achieves this transformation in a single step, enabling faster training and evaluation. Both approaches are similar, but differ in their probability paths: diffusion models use a diffusion-based path, while FMPE employs an optimal-transport path, leading to greater regularity and improved performance in low-data settings. We provide an ablation using diffusion in Table \ref{tab:ddpm vs fmpe}, and examine FMPE’s performance with varying data sizes in Table \ref{tab:fmpe data scaling}.

The function input $u$ is handed to the FMPE flow parameterized by $v^{\phi_0}_{t, \hat{u}}(\xi)$ as finite-dimensional conditional information $\hat{u}$.
We encode the infinite-dimensional $u$ into $\hat{u} = T^{\phi_1,\phi_2}(u)$ as an adapted FNO, mimicking the forward surrogate. We project the functions to the parameters, using a band-limited forward Fourier transform $\Tilde{h}$ to create a fixed-size latent representation of a function regardless of the input dimensions, subsequently projecting the large latent dimension to the smaller dimension $m$ of $\xi$ by FMPE,
\begin{equation}
    \hat{u} = T^{\phi_1,\phi_2}(u) = \Tilde{h} \circ \mathcal{K}^{\phi_1} \circ \mathcal{P}^{\phi_2} (u),
    \label{eq:fmpe-conditional-info}
\end{equation}
where $\mathcal{P}^{\phi_2}$ is lifting layer, and $\mathcal{K}^{\phi_1}$ a concatenation of FNO layers.
For the detailed formulation of the FMPE loss, we refer to the original reference  \cite{dax2023flow} and equation \eqref{eq:fmpe objective} in Appendix \ref{sec:appendix FMPE}, and formulate it simply as
\begin{equation}
    \mathcal{L}_2(\phi) = \mathcal{L}^{FMPE}(\phi),
    \label{eq:fmpe objective short}
\end{equation}
where $\phi = [\phi_0, \phi_1, \phi_2]$ are the joint trainable parameters of the flow itself and the FNO-based conditional information. The flow is then trained on data of the form $\{(u_i, \xi_i)\}_{i=1}^n$.

\paragraph{Unified Training and Evaluation}
The proposed methodology FUSE obtains the optimally trained parameters $\theta$ and $\phi$ by minimizing the respective objectives of Equations \eqref{eq:operator loss} and \eqref{eq:fmpe objective short},
\begin{equation}
    \mathcal{L}_{\text{FUSE}}(\theta, \phi) = \mathcal{L}^{\text{forward}}_{\text{FUSE}}(\theta) + \mathcal{L}^{\text{inverse}}_{\text{FUSE}}(\phi) = \mathcal{L}_1(\theta) + \mathcal{L}_2(\phi),
\end{equation}
thus minimizing the joint objective \eqref{eq:triangle inequality}. For all experiments presented here, the two FUSE objectives are decoupled during training, implying that for each training batch, $\mathcal{L}_1$ and $\mathcal{L}_2$ are calculated and backpropagation is performed separately over each set of network parameters. This strategy aids the training, allowing us to avoid hyperparameter tuning of the loss function. It is possible to backpropagate losses through the entire network as discussed in Appendix \ref{sec:training details appendix}; however, this does not result in improvements at evaluation time, further supporting our claim that decoupled objectives may be unified at evaluation time to draw connections between forward and inverse problems.

When employing the trained FUSE model, we assume that measurements $u$ are available and it is the goal to infer both the distribution of parameters $\xi$ that best fit the data and the distribution of function outputs $s$ that would result from those parameters. We obtain $M$ parameter samples from the trained FMPE model as $\xi_i \sim \rho^\phi(\xi|u),~i=1,..., M$, and pass these to the FNO forward surrogate to obtain $M$ samples of the simulated outputs $s_i=\Gct(\xi_i),~i=1,...,M$, where we assume $M$ sufficiently large to adequately sample the parameter distribution. We hence obtain an ensemble of estimates with mean prediction $\Bar{s} = \frac{1}{M} \sum^M_{i=1} s_i$ and standard deviation $\sigma_s = (\frac{1}{M-1} \sum^M_{i=1} ( s_i  - \Bar{s} )^2)^{1/2}$. While both model parts can also be evaluated separately, it is this \emph{propagated uncertainty} on $s$ that offers most insight in the effect of parametric uncertainty within the emulated models.

\section{Experiments} \label{sec:experiments}

For evaluation, we are given testing data of the form $(u,\xi^*, s)$ obtained by numerical simulation, where we will refer to $\xi^*$ as the \emph{data-generating parameters}. Based on measurements $s^*$, the inverse FMPE model provides samples of the distribution of discrete parameters $\xi_i$, that are compared to the true data-generating parameters $\xi^*$ using the Continuous Ranked Probability Score (CRPS). The performance of the deterministic forward prediction is measured with the relative $L^1$ and $L^2$ errors over the continuous data and predictions. We compare FUSE against cVANO \cite{kissas2023towards}, inVAErt \cite{tong2023invaert}, and GAROM \cite{garom}, and we consider an ablation where we substitute the FNO layers with UNet layers \cite{ronneberger2015unet} in the FUSE model and leave out the Fourier transforms. For details on the training and evaluation metrics, please refer to \ref{sec:evaluation metrics}. 

\textbf{PDE Test Cases}
As test cases, we consider a simulation of the pulse wave propagation (PWP) in the human arterial network and one of an atmospheric cold bubble (ACB). Both experiments are described by a set of PDEs whose solutions are fully parameterized by finite-dimensional parameters encoding model properties, as well as boundary and initial conditions. In both of these experiments, to replicate a patient or downburst observed in the real world, the solver needs to be calibrated for precise conditions, e.g. a specific patient or atmospheric conditions, and forward uncertainty quantification of the calibrated predictions are of interest. Details on both cases are provided in Appendices \ref{sec: appendix pwp} and \ref{sec:appendix acb}.

\textbf{Atmospheric Cold Bubble (ACB)}
In atmospheric modeling, the two-dimensional dry cold bubble case \cite{straka1993} is a well-known test case for numerical simulations resembling downbursts, which cause extreme surface winds in thunderstorms. Within a domain of neutral atmospheric stability without background wind, an elliptic cold air anomaly is prescribed and initiates a turbulent flow as it sinks to the ground, see Figure \ref{fig:bubble_fields} for an example.
Time series of horizontal and vertical velocity, $u$ and $w$, are recorded at eight sensors placed inside the domain. Similar data is obtained by weather stations ($z\approx 2$ m) or turbulence flux towers ($z\approx 10-50$ m), experienced by wind turbines or high-rise buildings ($z\approx 100$ m), or with unmanned aerial vehicles. 
The PDE model PyCLES \cite{pressel2015} used for simulation is parameterized by the turbulent eddy viscosity $\nu_t$ and diffusivity $D_t$ as model parameters, as well as four parameters describing the amplitude and shape of the initial cold perturbation ($a,~x_r,~z_r,~z_c$). The target task for FUSE here is to calibrate these parameters $\xi \in \mathbb{R}^6$ to time series measurements $u: [0,T]\rightarrow \mathbb{R}^{20}$, at ten locations for horizontal and vertical velocity each, where the input measurements used for calibration and the model output used for prediction and uncertainty quantification share the same space $\mathcal{U}=\mathcal{S}$ of time series.

\textbf{Pulse Wave Propagation (PWP) in the Human Body} The pulse wave propagation in the cardiovascular system contains a great deal of information regarding the health of an individual. For this reason, there are efforts to measure and leverage PWP in both wearable devices, e.g. smart-watches, and clinical medicine. Different systemic parameters such as stroke volume (SV), heart rate (HR), and patient age have been shown to affect the morphology of pulse waves \citep{charlton2019modeling}. These are used to parameterize pulse wave time series, measuring pressure, velocity, and photoplethysmography (PPG), at 13 locations of systemic arteries of the human cardiovascular system through a reduced-order PDE model in the data set published by \citep{charlton2019modeling}. See Tables \ref{tab:cv_parameters} and \ref{tab:artery_locations} in Section \ref{sec: appendix pwp} for the full set of parameters and locations. In clinical applications, only measurements at easily accessible locations (such as the wrist) are available, while it is the goal to predict the pulse signal at other locations of interest from the inferred parameters $\xi \in \mathbb{R}^{32}$. The space of input functions $\mathcal{U}$ hence differs from the space of output functions $\mathcal{S}$ in this test case.
In order to account for different clinical scenarios, we train the FUSE model using random masking of locations and evaluate it on different levels of available input information,
\begin{enumerate}
    \item[Level 1.] \textbf{Perfect information:} Pressure, velocity, and PPG at all locations,
    \item[Level 2.] \textbf{Intensive care unit information:} Pressure, velocity, and PPG at the wrist,
    \item[Level 3.] \textbf{Minimal information:} PPG at the fingertip.
\end{enumerate}

\section{Results} \label{sec:results}

Collected results for all experiments are summarized in Table \ref{tab:results inv for both}. While the main tasks of the FUSE model are evaluated for both test cases, we focus on ACB to exemplify its generalization properties and on PWP for different levels of input information. We would like to emphasize that the goal of uncertainty representation here is to capture parametric uncertainty inherent to the data-generating numerical model and its parameterizations. Our model does not provide a notion of uncertainty in the predictions due to imperfect training of the neural networks. Given that these errors are generally low (Table \ref{tab:results inv for both}), we are confident to interpret all spread on $\xi$ and $s$ given by FUSE as parametric model uncertainty. In particular, we expect the true posterior distributions to have positive spread, hence a prediction with small ensemble spread is not necessarily accurate. The CRPS is designed to judge model performance in this probabilistic setting.

\textbf{Inverse Problem}
Given time series measurements $u$, the inverse model samples from the distribution of parameters, $\xi \sim  \rho^{\phi}(\xi | u)$, with more samples in areas of the parameter space that are more likely to match the data-generating parameters $\xi^*$.
We find that FUSE outperforms the other methods when sufficient input information is given. We also found that inVAErt experiences posterior collapse for PWP in this setting, which is represented in high CRPS (Figure \ref{fig:invaert posterior collapse}). Histograms reveal that FUSE captures the expected dependencies of the data on the parameters (Figures \ref{fig:pwp median histograms}, \ref{fig:pwp worst histograms}), such as wider distributions for scarcer information in PWP. In the ACB case, sharp Dirac-like estimates are obtained for most of the samples (Figure \ref{fig:acb worst med crps}). However, a good fit for the numerical model parameters seems to rely on an accurate estimate of the initial condition (Figure \ref{fig:acb worst med histograms}).

\textbf{Forward Problem}
The surrogate $\mathcal{G}^{\theta}(\xi)$ is tasked to assess the ability of FUSE to predict the continuous $s$ from the data-generating parameters $\xi^*$. FUSE consistantly outperforms all other methods in both test cases. In particular, FUSE is able to reproduce very nonlinear dynamics and sharp velocity gradients in the ACB case (Figure \ref{fig:acb time series for different parameters}). Potential errors mostly lie in the bulk velocity of the anomaly, resulting in a stretching or squeezing of the time series at later times. As expected, predictions are slightly worse further away from the perturbation location, where the flow has become more turbulent, especially for the vertical velocity which exhibits strong oscillations (Figure \ref{fig:acb dc box plot errors each location}).
For the PWP, continuous predictions based on the true parameters have an L1 error of well below $0.3\%$ at all locations (Figures \ref{fig:pwp true parameters box plot}, \ref{fig:pwp worst case predictions from true parameters}). Smaller arteries generally exhibit a slightly higher error variance, which fits the physical problem interpretation, as smaller diameters show more variations to pressure than larger system arteries.

\begin{figure}[ht]
    \centering
    \begin{subfigure}[b]{0.19\textwidth}
        \includegraphics[width=\textwidth]{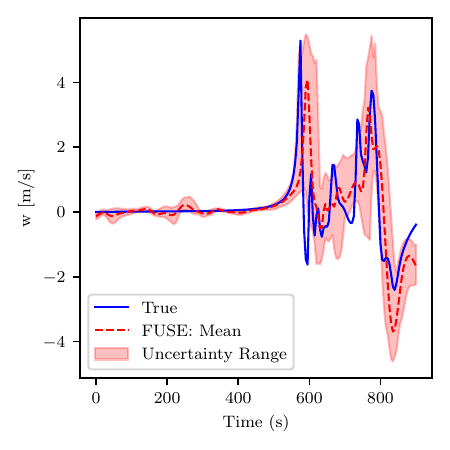}
        \caption{FUSE}
        \label{fig:sub1}
    \end{subfigure}
    \hfill
    \begin{subfigure}[b]{0.19\textwidth}
        \includegraphics[width=\textwidth]{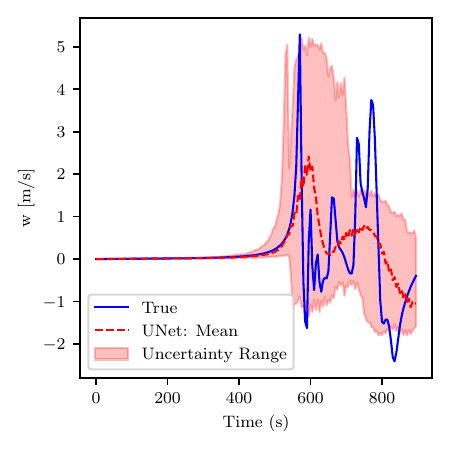}
        \caption{UNet}
        \label{fig:sub2}
    \end{subfigure}
    \hfill
    \begin{subfigure}[b]{0.19\textwidth}
        \includegraphics[width=\textwidth]{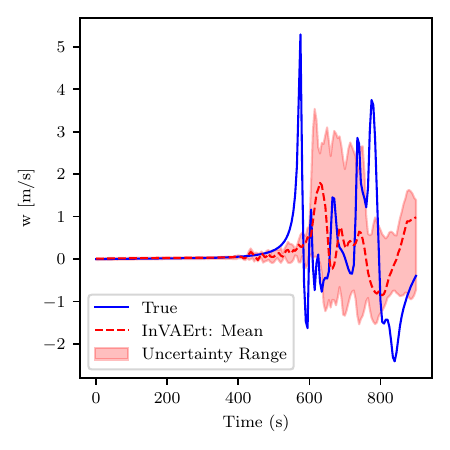}
        \caption{InVAErt}
        \label{fig:sub3}
    \end{subfigure}
    \hfill
    \begin{subfigure}[b]{0.19\textwidth}
        \includegraphics[width=\textwidth]{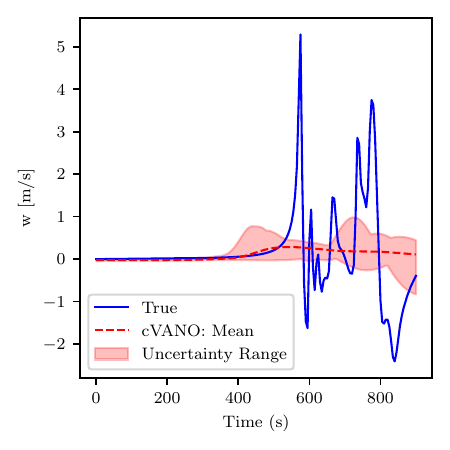}
        \caption{cVANO}
        \label{fig:sub4}
    \end{subfigure}
    \hfill
    \begin{subfigure}[b]{0.19\textwidth}
        \includegraphics[width=\textwidth]{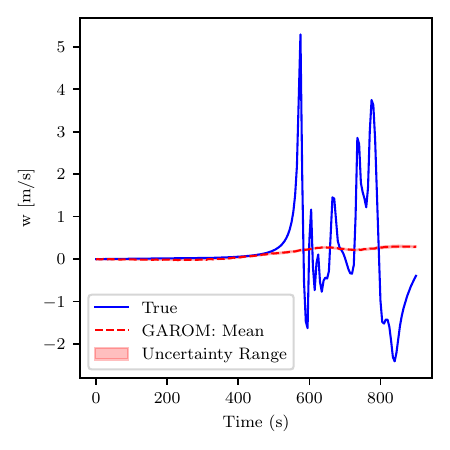}
        \caption{GAROM*}
        \label{fig:sub5}
    \end{subfigure}
    \caption{ACB, propagated uncertainty: Comparison of the benchmark models on the horizontal velocities at location 1, showing the unified ensemble predictions for the sample where FUSE has the greatest $L^1$ error, i.e. a worst-case scenario. *GAROM predicts the velocity from the true parameters, while all other models predict pressure from the posterior distribution over parameters given continuous inputs.}
    \label{fig:acb worst case prediction with uncertainty}
\end{figure}

\begin{figure}[ht]
    \centering
    \begin{subfigure}[b]{0.19\textwidth}
        \includegraphics[width=\textwidth]{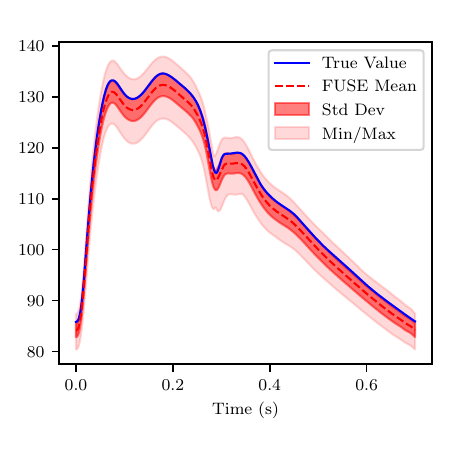}
        \caption{FUSE}
    \end{subfigure}
    \hfill
    \begin{subfigure}[b]{0.19\textwidth}
        \includegraphics[width=\textwidth]{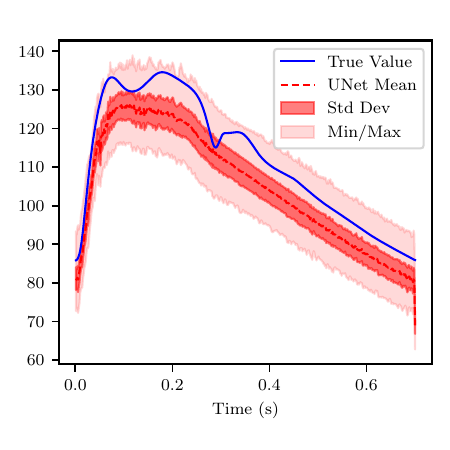}
        \caption{UNet}
    \end{subfigure}
    \hfill
    \begin{subfigure}[b]{0.19\textwidth}
        \includegraphics[width=\textwidth]{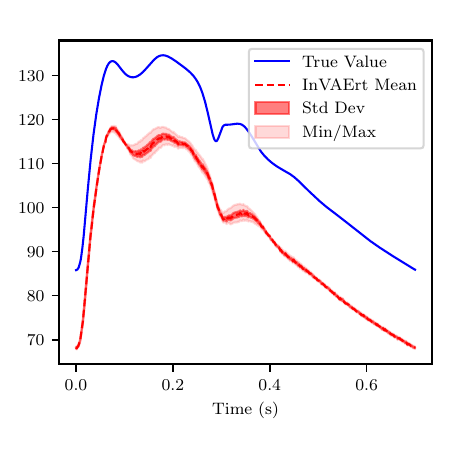}
        \caption{InVAErt}
    \end{subfigure}
    \hfill
    \begin{subfigure}[b]{0.19\textwidth}
        \includegraphics[width=\textwidth]{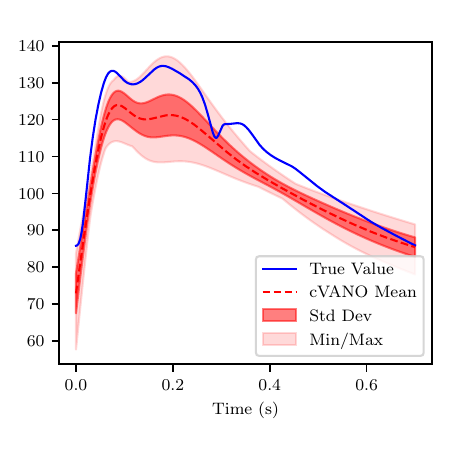}
        \caption{cVANO}
    \end{subfigure}
    \hfill
    \begin{subfigure}[b]{0.19\textwidth}
        \includegraphics[width=\textwidth]{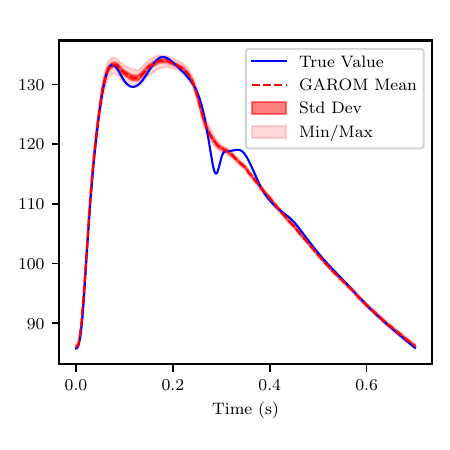}
        \caption{GAROM*}
    \end{subfigure}
    \caption{PWP, Comparison of models ability to predict the Sup. Middle Cerebral pressure and its uncertainty for the sample where FUSE has the greatest $L^1$ error, i.e. a worst-case scenario. *see Figure \ref{fig:acb worst case prediction with uncertainty}.}
    \label{fig:pwp worst case prediction with uncertainty}
\end{figure}

\textbf{Unified Uncertainty Propagation}
Given an input $u$, the concatenated evaluation of the inverse and forward parts of FUSE provides a distribution over the output functions $s$, by first sampling parameters $\xi_i$ and then evaluating the forward emulator on this ensemble, yielding continuous samples $s_i$. This can be interpreted as the uncertainty introduced by simulating the data with the parameterized model FUSE was trained on: Uncertainty is introduced as the parameters are not fully identifiable in the inverse step, and is propagated and expanded onto the continuous output through the ensemble predictions on the parameter samples. Again, FUSE outperforms all other models significantly in all experiments in the $L^1$ and $L^2$ error over the mean predictions $\Bar{s}$. Figures \ref{fig:acb worst case prediction with uncertainty} and \ref{fig:pwp worst case prediction with uncertainty} show how FUSE captures both smooth and rough structures, even on the sample it performs worst on. It only exhibits slight difficulties in predicting a constant velocity close to zero for ACB. At each location of the ACB, the $L^1$ error obtained with the propagated ensemble mean $\Bar{s}$ is comparable to the error by evaluating only the forward model, $s=\mathcal{G}^\theta (\xi^*)$ (Figure \ref{fig:acb cc box plot errors each location}). An inspection of the predictions for all locations for the PWP case shows that the true time series lie within one standard deviation of the FUSE ensemble mean for the median sample, and within the min-max uncertainty bounds for the worst sample (Figures \ref{fig:pwp median waveforms}, \ref{fig:pwp worst waveforms}). We observe that the accurate prediction of model parameters correlates to accurate predictions of pressures (Figure \ref{fig:pwp l1 crps correlation}).

\begin{figure}[ht]
    \centering
    \begin{subfigure}{0.48\textwidth}
        \includegraphics[width=\linewidth]{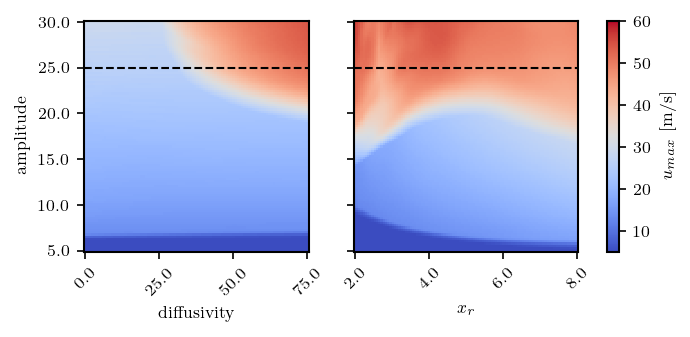}
    \end{subfigure}
    \hfill
    \begin{subfigure}{0.48\textwidth}
        \includegraphics[width=\linewidth]{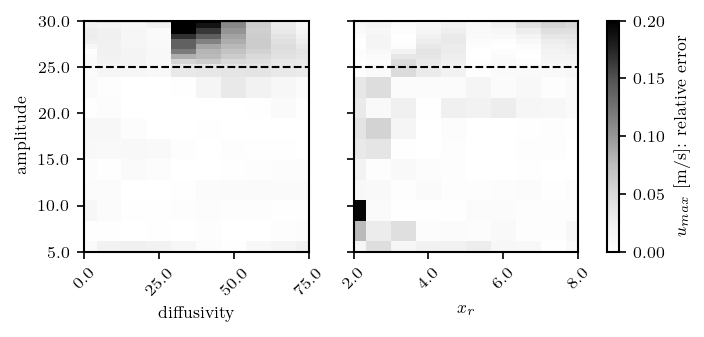}
    \end{subfigure}
    
    \caption{ACB, sensitivity analysis and generalization: Validation of the FUSE model against numeric simulations on peak horizontal velocities $u$ at location 1 ($x=15$ km, $z=50$ m). Samples above the dashed line correspond to amplitudes larger than seen during training. The parameter resolution is $100\times 100$ for FUSE, and $(10+10)\times 10$ for the numerical samples. Figure continued in the appendix, Figure \ref{fig:acb ood}.}
    \label{fig:tt pariwise validation}
\end{figure}

\textbf{Sensitivity Analysis}
Based on the good approximative properties showcased above, the forward operator $\mathcal{G}^\theta$ of FUSE can be used as a fast surrogate model to assess the sensitivity of the underlying PDE to varying the parameters $\xi$. Such analysis is usually constrained to a small sample size due to the large computational costs, and the dense sampling enabled by the surrogate helps to explore the parameter space for parameters that optimally fit patient data in PWP, or exhibit extreme winds in ACB. Varying one parameter at a time, while keeping all others fixed at their default value, exhibits the single effect on the data (\emph{fingerprint}). 
For ACB, an increased amplitude of the cold anomaly mainly results in a speed-up and hence a squeezing of the time series, while other parameters have more nonlinear effects (Figure \ref{fig:acb forward sensitivity}). As the FUSE model provides no access to the full velocity fields, only numerical simulations reveal that these sharp sensitivities are mainly caused by certain features (eddies) of the flow reaching or missing a sensor depending on the parameter value. A validation of this sensitivity analysis for an estimate of the maximal horizontal velocity over the time series is provided in Figure \ref{fig:tt pariwise validation} for a pairwise fingerprint on $100\times100$ samples, with consistently low errors. Only for one sample with a weak and small perturbation, the signal is not captured accurately. An additional evaluation of 60 samples at the margins of the parameter space shows that the peak velocities are consistently represented well (Figures \ref{fig:acb acb pairwise validation z2000}, \ref{fig:acb acb pairwise validation z50}).
For PWP, the sensitivity results (Figure \ref{fig:pwp fingerprints blood flow}) are consistent with observations reported in the literature \cite{charlton2019modeling}.

\textbf{Out of Distribution Generalization for ACB}
We test the generalization capabilities of the FUSE forward model on test samples with amplitudes larger than seen in training in the pairwise sensitivity setup (Figure \ref{fig:tt pariwise validation}). The FUSE model is able to capture the shape of the pairwise parameter dependencies, but struggles to locate the sharp dependence of the local peak velocity on the eddy diffusivity. Averaged errors in the out of distribution range are shown in Table \ref{tab:acb relative errors ood}, with FUSE performing clearly superior to all other models.

\textbf{Levels of Available Information for PWP}
The case of missing input information is modeled by masking certain input components when evaluating the models. While FUSE shows regular performance comparable with the baselines on levels 2 and 3 of missing information, cVANO seems to be better suited for this setting (Table \ref{tab:results inv for both}), probably due to the assumption that the latent dimensions are uncorrelated. In a case without conditional information, FUSE provides the prior distribution of the data (Figure \ref{fig:prior vs predicted}).
When the available input measurement locations of the PWP experiment are known a priori, it is possible to get more accurate results when training FUSE without masking. For example, training to predict pressures and parameters from only the PPG data at the fingertip only, without masking, results in a CRPS of $4.28 \times 10^{-2}$ and a relative $L^1$ error of 3.6\%, roughly half the error of the masked model. If the measurement data is fixed, predictions without masking are much more suitable for analysis.

\begin{table}
\small
    \centering
    \caption{Performance of FUSE and the baseline models in estimating parameters $\xi$ from continuous inputs $u$, quantified by CRPS, and predicting time series data $s$, quantified by a relative ${L}_1$ and ${L}_2$ error. Here, "True parameters" evaluates the forward model part only, and "Estimated parameters" and levels one to three evaluates the sample mean $\Bar{s}$ predicted by the unified model.
    }
    \begin{tabular}{c l l c c c}
    \toprule
        & Model & Experimental Setup & CRPS$\times10^2$ & Rel. ${L}_1$ Error & Rel. ${L}_2$ Error\\[\medskipamount]
        \cmidrule{1-6}
        \multirow{5}{*}[-5ex]{\rotatebox{90}{\textbf{Pulse Wave Propagation}}} 
        &FUSE (\textbf{ours}) & True Parameters & - &  \textbf{0.13 $\pm$ 0.05} \% & \textbf{0.19 $\pm$ 0.08} \%\\
        && Level 1 & \textbf{1.31  $\pm$ 0.69 } & \textbf{0.86 $\pm$ 0.93} \% & \textbf{0.90 $\pm$ 0.92} \% \\
        && Level 2 & \textbf{3.18  $\pm$ 1.40 } & \textbf{3.83 $\pm$ 2.66} \% & \textbf{3.87 $\pm$ 2.64} \% \\
        && Level 3 & \(8.48  \pm 2.55 \) & \textbf{6.56 $\pm$ 3.36} \% & \textbf{7.05 $\pm$ 3.39} \%\\
        \cmidrule{2-6}
        &UNet (\textbf{ablation}) & True Parameters & - & 1.38 $\pm$ 0.34 \% & 2.00 $\pm$ 0.48 \%\\
        && Level 1 & \(3.79 \pm 0.97 \)   & 3.81 $\pm$  2.03\% & 4.25 $\pm$ 1.92 \% \\
        && Level 2 & \(5.75  \pm 1.32 \)  & 6.79 $\pm$ 3.77 \% & 7.19 $\pm$ 3.68 \% \\
        && Level 3 & \(12.72  \pm 2.65 \) & 7.83 $\pm$ 2.64 \% & 8.78 $\pm$ 2.87 \%\\
        \cmidrule{2-6}
        &InVAErt & True Parameters & - & 3.44 $\pm$ 0.52 \% & 4.21 $\pm$ 0.60 \%\\
        && Level 1 & \(11.62 \pm 2.39 \) & 8.28 $\pm$ 4.78 \% & 8.94 $\pm$ 4.91 \% \\
        && Level 2 & \(16.88 \pm 3.67 \) & 7.85 $\pm$ 2.80 \% & 8.64 $\pm$ 2.99 \% \\
        && Level 3 & \(21.22 \pm 4.35 \) & 8.37 $\pm$ 2.73 \% & 9.48 $\pm$ 2.98 \%\\
        \cmidrule{2-6}
        &cVANO & True Parameters & - & 7.49 $\pm$ 2.60 \% & 8.51 $\pm$ 2.58 \%\\
        && Level 1 & 5.61 $\pm$ 0.24 & 7.54 $\pm$ 2.52 \% & 8.52 $\pm$ 2.53 \% \\
        && Level 2 & 4.00 $\pm$ 2.22 & 8.42 $\pm$ 3.11 \% & 9.72 $\pm$ 3.23 \% \\
        && Level 3 & \textbf{3.66 $\pm$ 0.76} &10.79 $\pm$ 3.60 \% &12.54 $\pm$ 4.05 \%\\
        \cmidrule{2-6}
        &GAROM & True Parameters & - & 0.70 $\pm$ 0.13 \% & 1.04 $\pm$ 0.22 \%\\
        && Levels 1-3 & - &  - & - \\[\medskipamount]
        \toprule
        \multirow{5}{*}[-3ex]{\rotatebox{90}{\textbf{Atm. Cold Bubble}}} 
        &FUSE (\textbf{ours}) & True Parameters        & -               & \textbf{0.41 $\pm$ 0.45} \% & \textbf{0.80 $\pm$ 0.89} \% \\
        && Estimated Parameters   & \textbf{1.84 $\pm$ 1.63} & \textbf{0.58 $\pm$ 0.59} \% & \textbf{1.13 $\pm$ 1.11} \% \\
        \cmidrule{2-6}
        &UNet (\textbf{ablation}) & True Parameters        & -               & 1.70 $\pm$ 0.90 \% & 2.86 $\pm$ 1.55 \% \\
        && Estimated Parameters   & 4.47 $\pm$ 2.67 & 2.13 $\pm$ 1.16 \% & 3.69 $\pm$ 1.94 \% \\
        \cmidrule{2-6}
        &InVAErt & True Parameters        & -               & 1.11 $\pm$ 0.85 \% & 1.94 $\pm$ 1.31 \% \\
        && Estimated Parameters   & 6.65 $\pm$ 9.05 & 1.17 $\pm$ 1.35 \% & 2.88 $\pm$ 2.04 \% \\
        \cmidrule{2-6}
        &cVANO & True Parameters        & - & 5.32 $\pm$ 1.57 \% & 8.26 $\pm$ 2.44\% \\
        && Estimated Parameters   & 19.84 $\pm$ 4.73 & 8.77 $\pm$ 3.04 \%& 12.33 $\pm$ 4.00\% \\
        \cmidrule{2-6}
        &GAROM & True Parameters        & - & 3.99 $\pm$ 1.67 \% & 6.35 $\pm$ 2.43 \% \\
        && Estimated Parameters   & - & - & - \\
        \bottomrule
    \end{tabular}
    \label{tab:results inv for both}
\end{table}

\begin{table}
\small
    \centering
    \caption{Performance of FUSE on out-of-distribution samples for ACB, as described in Figure \ref{fig:tt pariwise validation}, averaged over the samples, the velocity components $u$ and $w$, and the measurement locations.}
    \begin{tabular}{c l l c c c}
    \toprule
        & Model & Experimental Setup & CRPS$\times10^2$ & Rel. ${L}_1$ Error & Rel. ${L}_2$ Error\\[\medskipamount]
        \cmidrule{1-6}
        \multirow{5}{*}[-0.5ex]{\rotatebox{90}{\textbf{Out-of-distribution (ACB)}}}
        & Fuse (\textbf{ours})
        & True Parameters        & -               & \textbf{1.83 $\pm$ 1.10} \% & \textbf{3.15 $\pm$ 1.80} \% \\
        & & Estimated Parameters   & \textbf{3.13 $\pm$ 1.13} & \textbf{1.63 $\pm$ 0.78} \% & \textbf{2.77 $\pm$ 1.28} \% \\
        \cmidrule{2-6}
        & UNet (\textbf{ablation})
        & True Parameters        & -               & 2.78 $\pm$ 0.32 \% & 4.37 $\pm$ 0.43 \% \\
        & & Estimated Parameters   & 12.36 $\pm$ 5.47 & 3.43 $\pm$ 0.40 \% & 5.40 $\pm$ 0.58 \% \\
        \cmidrule{2-6}
        & InVAErt
        & True Parameters        & -               & 4.48 $\pm$ 2.65 \% & 6.49 $\pm$ 3.26 \% \\
        & & Estimated Parameters   & 18.80 $\pm$ 19.09 & 4.31 $\pm$ 1.13 \% & 6.92 $\pm$ 1.83 \% \\
        \cmidrule{2-6}
        & cVANO
        & True Parameters        & - & 9.31 $\pm$ 1.22 \% & 13.74 $\pm$ 1.54 \% \\
        & & Estimated Parameters   & 24.76 $\pm$ 1.45 & 15.35 $\pm$ 1.28 \%& 21.20 $\pm$ 1.49\% \\
        \cmidrule{2-6}
        &GAROM
        &True Parameters        & - & 7.22 $\pm$ 1.00 \% & 10.96 $\pm$ 1.22 \% \\
        & & Estimated Parameters   & - & - & - \\
        \bottomrule
    \end{tabular}
    \label{tab:acb relative errors ood}
\end{table}

\section{Discussion} \label{sec:discussion}

\textbf{Summary} We propose FUSE, a framework for unifying surrogate modeling and parameter identification for parametric PDEs by unifying, as an example, operator learning with flow-matching posterior estimation. Both are represented in the FUSE objective through a deterministic forward loss and a probabilistic inverse loss, respectively. The joint architecture allows for inverse estimation of finite-dimensional parameters $\xi$ given continuous measurements $u$, as well as forward predictions of (other) continuous measurements $s$ given $\xi$. Inheriting from the properties of its constitutive components, FUSE is discretization invariant with respect to the continuous inputs and outputs, and able to represent arbitrary distributions in the fully interpretable parameter space. As this latent space is defined through a known parameterization of the underlying PDE used for numerical modeling, FUSE allows for accurate calibration and evaluation of a numerical solver, as well as performing downstream tasks such as parameter uncertainty quantification. On two test cases of parametric PDEs, namely the pulse wave propagation (PWP) in the human cardiovascular system and an atmospheric cold bubble (ACB) with time series measurements, FUSE consistently outperforms four baseline methods designed to perform similar tasks. It also shows good out-of-distribution generalization and great flexibility when trained and evaluated on different levels of available input information.

\textbf{Connection to Existing Methods} 
The proposed methodology and the adapted implementation of the FNO and FMPE offer the following advantages over the existing methods for joint parameter estimation and forward emulation mentioned in the introduction. By incorporating Neural Operators in both the forward and inverse parts, we ensure discretization invariance throughout the model \cite{bartolucci2024representation}.
In contrast to Variational Autoencoders, we present a fully probabilistic formulation and require no restriction on the latent data distribution. This allows us to model arbitrary posterior distributions, whereas cVANO is restricted to Gaussian distributions. Likewise, FUSE allows one to leverage powerful, existing neural operators which improves predictions on the output functions. The experiments have shown that FUSE excels over all baselines in both test cases. In particular, cVANO and GAROM fail to capture even rough characteristics of the flow in the forward emulation for the ACB case with nonlinear measurements, and InVAErt shows a consistent bias towards the mean in the PWP task. Only in the case of missing information in the smoother PWP case, cVANO outperforms FUSE on the inverse problem. In terms of the choice of FNO in both the forward and inverse part of FUSE, the ablation with UNet layers exhibits larger uncertainties and less performance of the predicted mean. An additional ablation, presented in the Appendix \ref{tab:ddpm vs fmpe}, shows that optimal transport probability paths perform better than the diffusion-based paths in a conditional DDPM model. Finally, Gaussian processes are often chosen for their ability to handle unstructured data. We incorporate this property by using the grid-independent FNO implementation by \cite{lingsch2024regular}.

\textbf{Applicability} FUSE is explicitly formulated for finite-dimensional parameters and function measurements and outputs. This setting naturally arises when calibrating a numerical solver, with training data generated synthetically by the solver itself, and was demonstrated with out two test cases. However, it is very common for real measured datasets e.g. in bioengineering \cite{johnson2016mimic}, and more specifically when involving PWP, to contain both time-series data and vectors of parameters available for different patients. FUSE naturally extends to these datasets, where an underlying PDE is assumed but not formulated or simulated explicitly, including parameter inference using real data \cite{simulation-based-inference}, precision medicine or solver calibration \cite{richter2024bayesianwindkesselcalibrationusing}, and fingerprinting to discover parameter-disease correlations \cite{sokolow2024aortic}. In the case that the parameters are non-physical, and hence not measurable, such as the numerical model parameters in the ACB case, this does of course not apply, and the main aim of solving the forward and inverse problem is to explore the solver itself. Moreover, FUSE may help to refine existing parameterizations either through disentanglement \cite{higgins2017beta} or by identifying parameters with little influence on the simulated data.

\textbf{Limitations and Future Work} The FUSE framework aims at emulating a given parameterization of a PDE. If such a parameterization (or a complete dataset) is not available, VAE-based pre-processing \cite{bojanowski2017optimizing} or manifold discovery \cite{brehmer2020flows} may be used. When applying FUSE to real measurements, the parameterization turns to be only an approximation of the dynamical system, whereas it is considered a perfect model in the experiments presented here. This structural model uncertainty has to be added to the interpretation of the unified uncertainties given by the FUSE model, as it has to be for any other inverse algorithm. In practice, measurements are always associated with a measurement error, which is considerable in particular for extreme measurements. The current implementation of FUSE does not allow for errors in the continuous inputs to be taken into account, and it is left for a future extension of the framework to include these into a fully Bayesian formulation of the inverse model. In two or three-dimensional applications, such as atmospheric modeling, the parameters of interest might be themselves space/time-dependent functions, such as surface properties or initial conditions \citep{ruckstuhl2020}. Based on the scaling properties of the FMPE model for high-dimensional data such as images \citep{lipman2023flow}, FUSE is expected to scale well in size for high-dimensional parameter spaces, but is yet to be extended to a resolution-invariant function representation in latent space. All these limitations will be addressed in future research.

\section*{Acknowledgments}
G.K. would like to acknowledge support from Asuera Stiftung via the ETH Zurich Foundation.
\bibliographystyle{abbrvnat} 

\bibliography{references}

\begin{thebibliography}{62}
\providecommand{\natexlab}[1]{#1}
\providecommand{\url}[1]{\texttt{#1}}
\expandafter\ifx\csname urlstyle\endcsname\relax
  \providecommand{\doi}[1]{doi: #1}\else
  \providecommand{\doi}{doi: \begingroup \urlstyle{rm}\Url}\fi

\bibitem[Alastruey et~al.(2012)Alastruey, Parker, Sherwin, et~al.]{alastruey2012arterial}
J.~Alastruey, K.~H. Parker, S.~J. Sherwin, et~al.
\newblock Arterial pulse wave haemodynamics.
\newblock In \emph{11th international conference on pressure surges}, volume~30, pages 401--443. Virtual PiE Led t/a BHR Group Lisbon, Portugal, 2012.

\bibitem[Arzani et~al.(2022)Arzani, Wang, Sacks, and Shadden]{arzani2022machine}
A.~Arzani, J.-X. Wang, M.~S. Sacks, and S.~C. Shadden.
\newblock Machine learning for cardiovascular biomechanics modeling: challenges and beyond.
\newblock \emph{Annals of Biomedical Engineering}, 50\penalty0 (6):\penalty0 615--627, 2022.
\newblock \doi{10.1007/s10439-022-02967-4}.

\bibitem[Bartolucci et~al.(2024)Bartolucci, de~Bezenac, Raonic, Molinaro, Mishra, and Alaifari]{bartolucci2024representation}
F.~Bartolucci, E.~de~Bezenac, B.~Raonic, R.~Molinaro, S.~Mishra, and R.~Alaifari.
\newblock Representation equivalent neural operators: a framework for alias-free operator learning.
\newblock \emph{Advances in Neural Information Processing Systems}, 36, 2024.
\newblock URL \url{https://papers.nips.cc/paper_files/paper/2023/hash/dc35c593e61f6df62db541b976d09dcf-Abstract-Conference.html}.

\bibitem[Bojanowski et~al.(2018)Bojanowski, Joulin, Lopez-Pas, and Szlam]{bojanowski2017optimizing}
P.~Bojanowski, A.~Joulin, D.~Lopez-Pas, and A.~Szlam.
\newblock Optimizing the latent space of generative networks.
\newblock In \emph{Proceedings of the 35th International Conference on Machine Learning}, volume~80 of \emph{Proceedings of Machine Learning Research}, pages 600--609. PMLR, 10--15 Jul 2018.
\newblock URL \url{https://proceedings.mlr.press/v80/bojanowski18a.html}.

\bibitem[Brehmer and Cranmer(2020)]{brehmer2020flows}
J.~Brehmer and K.~Cranmer.
\newblock Flows for simultaneous manifold learning and density estimation.
\newblock In \emph{Advances in Neural Information Processing Systems (NeurIPS)}, 2020.
\newblock URL \url{https://proceedings.neurips.cc/paper/2020/hash/051928341be67dcba03f0e04104d9047-Abstract.html}.

\bibitem[{\v{C}}ani{\'c} et~al.(2006){\v{C}}ani{\'c}, Tamba{\v{c}}a, Guidoboni, Mikeli{\'c}, Hartley, and Rosenstrauch]{vcanic2006modeling}
S.~{\v{C}}ani{\'c}, J.~Tamba{\v{c}}a, G.~Guidoboni, A.~Mikeli{\'c}, C.~J. Hartley, and D.~Rosenstrauch.
\newblock Modeling viscoelastic behavior of arterial walls and their interaction with pulsatile blood flow.
\newblock \emph{SIAM Journal on Applied Mathematics}, 67\penalty0 (1):\penalty0 164--193, 2006.
\newblock \doi{10.1137/060651562}.

\bibitem[Charlton et~al.(2019)Charlton, Mariscal~Harana, Vennin, Li, Chowienczyk, and Alastruey]{charlton2019modeling}
P.~H. Charlton, J.~Mariscal~Harana, S.~Vennin, Y.~Li, P.~Chowienczyk, and J.~Alastruey.
\newblock Modeling arterial pulse waves in healthy aging: a database for in silico evaluation of hemodynamics and pulse wave indexes.
\newblock \emph{American Journal of Physiology-Heart and Circulatory Physiology}, 317\penalty0 (5):\penalty0 H1062--H1085, 2019.
\newblock \doi{10.1152/ajpheart.00218.2019}.

\bibitem[Chen et~al.(2023)Chen, Xiang, Cho, Chang, Pershing, Maia, Chiaramonte, Carlberg, and Grinspun]{chen2023crom}
P.~Y. Chen, J.~Xiang, D.~H. Cho, Y.~Chang, G.~A. Pershing, H.~T. Maia, M.~M. Chiaramonte, K.~T. Carlberg, and E.~Grinspun.
\newblock {CROM}: Continuous reduced-order modeling of {PDE}s using implicit neural representations.
\newblock In \emph{The Eleventh International Conference on Learning Representations}, 2023.
\newblock URL \url{https://iclr.cc/virtual/2023/poster/12094}.

\bibitem[Cleary et~al.(2021)Cleary, Garbuno-Inigo, Lan, Schneider, and Stuart]{cleary2021}
E.~Cleary, A.~Garbuno-Inigo, S.~Lan, T.~Schneider, and A.~M. Stuart.
\newblock Calibrate, emulate, sample.
\newblock \emph{Journal of Computational Physics}, 424:\penalty0 109716, 2021.
\newblock ISSN 0021-9991.
\newblock \doi{10.1016/j.jcp.2020.109716}.
\newblock URL \url{https://www.sciencedirect.com/science/article/pii/S0021999120304903}.

\bibitem[Coscia et~al.(2024)Coscia, Demo, and Rozza]{garom}
D.~Coscia, N.~Demo, and G.~Rozza.
\newblock Generative adversarial reduced order modelling.
\newblock \emph{Sci Rep}, 14\penalty0 (3826), 2024.
\newblock \doi{10.1038/s41598-024-54067-z}.

\bibitem[Dinh et~al.(2015)Dinh, Krueger, and Bengio]{dinh2015nice}
L.~Dinh, D.~Krueger, and Y.~Bengio.
\newblock Nice: Non-linear independent components estimation.
\newblock In \emph{International Conference on Learning Representations}, 2015.
\newblock URL \url{https://arxiv.org/abs/1410.8516}.

\bibitem[Dinh et~al.(2017)Dinh, Sohl-Dickstein, and Bengio]{dinh2017density}
L.~Dinh, J.~Sohl-Dickstein, and S.~Bengio.
\newblock Density estimation using real {NVP}.
\newblock In \emph{The Fifth International Conference on Learning Representations}, 2017.
\newblock URL \url{https://openreview.net/forum?id=HkpbnH9lx}.

\bibitem[Formaggia et~al.(2010)Formaggia, Quarteroni, and Veneziani]{formaggia2010cardiovascular}
L.~Formaggia, A.~Quarteroni, and A.~Veneziani.
\newblock \emph{Cardiovascular Mathematics: Modeling and simulation of the circulatory system}, volume~1.
\newblock Springer Science \& Business Media, 2010.

\bibitem[Gloeckler et~al.(2024)Gloeckler, Deistler, Weilbach, Wood, and Macke]{gloeckler2024allinone}
M.~Gloeckler, M.~Deistler, C.~D. Weilbach, F.~Wood, and J.~H. Macke.
\newblock All-in-one simulation-based inference.
\newblock In \emph{Proceedings of the 41st International Conference on Machine Learning}, volume 235, pages 15735--15766. PMLR, 21--27 Jul 2024.
\newblock URL \url{https://proceedings.mlr.press/v235/gloeckler24a.html}.

\bibitem[Gulian et~al.(2022)Gulian, Frankel, and Swiler]{gulian2020gaussian}
M.~Gulian, A.~Frankel, and L.~Swiler.
\newblock Gaussian process regression constrained by boundary value problems.
\newblock \emph{Computer Methods in Applied Mechanics and Engineering}, 388:\penalty0 114117, 2022.
\newblock \doi{https://doi.org/10.1016/j.cma.2021.114117}.

\bibitem[Halder et~al.(2020)Halder, Damodaran, and Khoo]{halder2020deep}
R.~Halder, M.~Damodaran, and B.~C. Khoo.
\newblock Deep learning based reduced order model for airfoil-gust and aeroelastic interaction.
\newblock \emph{AIAA Journal}, 58\penalty0 (4):\penalty0 1595--1606, 8 2020.
\newblock \doi{10.2514/1.J059027}.

\bibitem[Hao et~al.(2023)Hao, Wang, Su, Ying, Dong, Liu, Cheng, Song, and Zhu]{hao2023gnot}
Z.~Hao, Z.~Wang, H.~Su, C.~Ying, Y.~Dong, S.~Liu, Z.~Cheng, J.~Song, and J.~Zhu.
\newblock {GNOT}: A general neural operator transformer for operator learning.
\newblock In \emph{Proceedings of the 40th International Conference on Machine Learning}, volume 202, pages 12556--12569. PMLR, 23--29 Jul 2023.
\newblock URL \url{https://proceedings.mlr.press/v202/hao23c.html}.

\bibitem[Higgins et~al.(2017)Higgins, Matthey, Pal, Burgess, Glorot, Botvinick, Mohamed, and Lerchner]{higgins2017beta}
I.~Higgins, L.~Matthey, A.~Pal, C.~P. Burgess, X.~Glorot, M.~M. Botvinick, S.~Mohamed, and A.~Lerchner.
\newblock beta-{VAE}: Learning basic visual concepts with a constrained variational framework.
\newblock In \emph{The Fifth International Conference on Learning Representations}, volume~3, 2017.
\newblock URL \url{https://openreview.net/forum?id=Sy2fzU9gl}.

\bibitem[Hijazi et~al.(2023)Hijazi, Freitag, and Landwehr]{hijazi2023pod}
S.~Hijazi, M.~Freitag, and N.~Landwehr.
\newblock {POD}-galerkin reduced order models and physics-informed neural networks for solving inverse problems for the navier–stokes equations.
\newblock \emph{Adv. Model. and Simul. in Eng. Sci.}, 10\penalty0 (5):\penalty0 5, 03 2023.
\newblock \doi{10.1186/s40323-023-00242-2}.

\bibitem[Ho et~al.(2020)Ho, Jain, and Abbeel]{Ho2020DDPM}
J.~Ho, A.~Jain, and P.~Abbeel.
\newblock Denoising diffusion probabilistic models.
\newblock In \emph{Advances in Neural Information Processing Systems}, volume~33, pages 6840--6851. Curran Associates, Inc., 2020.
\newblock URL \url{https://papers.nips.cc/paper_files/paper/2020/hash/4c5bcfec8584af0d967f1ab10179ca4b-Abstract.html}.

\bibitem[Jin et~al.(2022)Jin, Meng, and Lu]{jin2022mionet}
P.~Jin, S.~Meng, and L.~Lu.
\newblock {MIONet}: Learning multiple-input operators via tensor product.
\newblock \emph{SIAM Journal on Scientific Computing}, 44\penalty0 (6):\penalty0 A3490--A3514, 2022.
\newblock \doi{10.1137/22M1477751}.

\bibitem[Johnson et~al.(2016)Johnson, Pollard, Shen, Li-wei, Feng, Ghassemi, Moody, Szolovits, Celi, and Mark]{johnson2016mimic}
A.~E.~W. Johnson, T.~J. Pollard, L.~Shen, H.~L. Li-wei, M.~Feng, M.~Ghassemi, B.~Moody, P.~Szolovits, L.~A. Celi, and R.~G. Mark.
\newblock {MIMIC-III}, a freely accessible critical care database.
\newblock \emph{Scientific data}, 3:\penalty0 160035, 2016.
\newblock \doi{10.1038/sdata.2016.35}.

\bibitem[Kim et~al.(2019)Kim, Azevedo, Thuerey, Kim, Gross, and Solenthaler]{kim2019deep}
B.~Kim, V.~C. Azevedo, N.~Thuerey, T.~Kim, M.~Gross, and B.~Solenthaler.
\newblock Deep fluids: A generative network for parameterized fluid simulations.
\newblock \emph{Computer graphics forum}, 38\penalty0 (2):\penalty0 59--70, 2019.
\newblock \doi{10.1111/cgf.13619}.

\bibitem[Kingma and Dhariwal(2018)]{kingma2018glow}
D.~P. Kingma and P.~Dhariwal.
\newblock Glow: Generative flow with invertible 1x1 convolutions.
\newblock In S.~Bengio, H.~Wallach, H.~Larochelle, K.~Grauman, N.~Cesa-Bianchi, and R.~Garnett, editors, \emph{Advances in Neural Information Processing Systems}, volume~31. Curran Associates, Inc., 2018.
\newblock URL \url{https://papers.nips.cc/paper_files/paper/2018/hash/d139db6a236200b21cc7f752979132d0-Abstract.html}.

\bibitem[Kingma and Welling(2013)]{kingma2022autoencoding}
D.~P. Kingma and M.~Welling.
\newblock Auto-encoding variational bayes.
\newblock In \emph{International Conference on Learning Representations}, 2013.
\newblock URL \url{https://arxiv.org/abs/1312.6114}.

\bibitem[Kissas(2023)]{kissas2023towards}
G.~Kissas.
\newblock \emph{Towards Digital Twins for Cardiovascular Flows: A Hybrid Machine Learning and Computational Fluid Dynamics Approach}.
\newblock PhD thesis, University of Pennsylvania, 2023.

\bibitem[Kissas et~al.(2022)Kissas, Seidman, Guilhoto, Preciado, Pappas, and Perdikaris]{kissas2022learning}
G.~Kissas, J.~H. Seidman, L.~F. Guilhoto, V.~M. Preciado, G.~J. Pappas, and P.~Perdikaris.
\newblock Learning operators with coupled attention.
\newblock \emph{J. Mach. Learn. Res.}, 23\penalty0 (1), Jan. 2022.
\newblock URL \url{https://dl.acm.org/doi/10.5555/3586589.3586804}.

\bibitem[Kovachki et~al.(2023)Kovachki, Li, Liu, Azizzadenesheli, Bhattacharya, Stuart, and Anandkumar]{kovachki2023neural}
N.~Kovachki, Z.~Li, B.~Liu, K.~Azizzadenesheli, K.~Bhattacharya, A.~Stuart, and A.~Anandkumar.
\newblock Neural operator: Learning maps between function spaces with applications to pdes.
\newblock \emph{Journal of Machine Learning Research}, 24\penalty0 (89):\penalty0 1--97, 2023.
\newblock URL \url{http://jmlr.org/papers/v24/21-1524.html}.

\bibitem[Lanthaler et~al.(2022)Lanthaler, Mishra, and Karniadakis]{LMK1}
S.~Lanthaler, S.~Mishra, and G.~E. Karniadakis.
\newblock Error estimates for {D}eep{ON}ets: {A} deep learning framework in infinite dimensions.
\newblock \emph{Transactions of Mathematics and Its Applications}, 6\penalty0 (1):\penalty0 tnac001, 2022.
\newblock \doi{10.1093/imatrm/tnac001}.

\bibitem[Letafati et~al.(2024)Letafati, Ali, and Latva-aho]{letafati2024}
M.~Letafati, S.~Ali, and M.~Latva-aho.
\newblock Conditional denoising diffusion probabilistic models for data reconstruction enhancement in wireless communications, 2024.
\newblock URL \url{https://arxiv.org/abs/2310.19460}.

\bibitem[Li et~al.(2021)Li, Kovachki, Azizzadenesheli, Liu, Bhattacharya, Stuart, and Anandkumar]{li2021fourier}
Z.~Li, N.~Kovachki, K.~Azizzadenesheli, B.~Liu, K.~Bhattacharya, A.~Stuart, and A.~Anandkumar.
\newblock Fourier neural operator for parametric partial differential equations.
\newblock In \emph{International Conference on Learning Representations}, volume 2010.08895, 2021.
\newblock URL \url{https://iclr.cc/virtual/2021/poster/3281}.

\bibitem[Li et~al.(2023{\natexlab{a}})Li, Meidani, and Farimani]{li2023transformer}
Z.~Li, K.~Meidani, and A.~B. Farimani.
\newblock Transformer for partial differential equations{\textquoteright} operator learning.
\newblock \emph{Transactions on Machine Learning Research}, 2023{\natexlab{a}}.
\newblock ISSN 2835-8856.
\newblock URL \url{https://openreview.net/forum?id=EPPqt3uERT}.

\bibitem[Li et~al.(2023{\natexlab{b}})Li, Shu, and Barati~Farimani]{li2023scalable}
Z.~Li, D.~Shu, and A.~Barati~Farimani.
\newblock Scalable transformer for pde surrogate modeling.
\newblock In \emph{Advances in Neural Information Processing Systems}, volume~36, pages 28010--28039. Curran Associates, Inc., 2023{\natexlab{b}}.
\newblock URL \url{https://papers.nips.cc/paper_files/paper/2023/hash/590daf74f99ee85df3d8c007df9c8187-Abstract-Conference.html}.

\bibitem[Lingsch et~al.(2024)Lingsch, Michelis, De~Bezenac, M.~Perera, Katzschmann, and Mishra]{lingsch2024regular}
L.~E. Lingsch, M.~Y. Michelis, E.~De~Bezenac, S.~M.~Perera, R.~K. Katzschmann, and S.~Mishra.
\newblock Beyond regular grids: {F}ourier-based neural operators on arbitrary domains.
\newblock In \emph{Proceedings of the 41st International Conference on Machine Learning}, volume 235, pages 30610--30629. PMLR, 21--27 Jul 2024.
\newblock URL \url{https://proceedings.mlr.press/v235/lingsch24a.html}.

\bibitem[Lipman et~al.(2023)Lipman, Chen, Ben-Hamu, Nickel, and Le]{lipman2023flow}
Y.~Lipman, R.~T.~Q. Chen, H.~Ben-Hamu, M.~Nickel, and M.~Le.
\newblock Flow matching for generative modeling.
\newblock In \emph{International Conference on Learning Representations}, 2023.
\newblock URL \url{https://iclr.cc/virtual/2023/poster/11309}.

\bibitem[Liu et~al.(2022)Liu, Zeman, Sørland, and Schär]{liu2022}
S.~Liu, C.~Zeman, S.~L. Sørland, and C.~Schär.
\newblock Systematic {{Calibration}} of a {{Convection-Resolving Model}}: {{Application Over Tropical Atlantic}}.
\newblock \emph{Journal of Geophysical Research: Atmospheres}, 127\penalty0 (23), 2022.
\newblock \doi{10.1029/2022JD037303}.

\bibitem[Lu et~al.(2021)Lu, Jin, Pang, Zhang, and Karniadakis]{Lu_2021}
L.~Lu, P.~Jin, G.~Pang, Z.~Zhang, and G.~E. Karniadakis.
\newblock Learning nonlinear operators via {DeepONet} based on the universal approximation theorem of operators.
\newblock \emph{Nature Machine Intelligence}, 3\penalty0 (3):\penalty0 218–229, Mar. 2021.
\newblock \doi{10.1038/s42256-021-00302-5}.

\bibitem[Moore(2024)]{moore2024}
A.~Moore.
\newblock Investigating the {{Near-Surface Wind Fields}} of {{Downbursts}} using a {{Series}} of {{High-Resolution Idealized Simulations}}.
\newblock \emph{Weather and Forecasting}, 2024.
\newblock \doi{10.1175/WAF-D-23-0164.1}.

\bibitem[Orf et~al.(2012)Orf, Kantor, and Savory]{orf2012}
L.~Orf, E.~Kantor, and E.~Savory.
\newblock Simulation of a downburst-producing thunderstorm using a very high-resolution three-dimensional cloud model.
\newblock \emph{Journal of Wind Engineering and Industrial Aerodynamics}, 2012.
\newblock \doi{10.1016/j.jweia.2012.02.020}.

\bibitem[Parodi et~al.(2019)Parodi, Lagasio, Maugeri, Turato, and Gallus]{parodi2019}
A.~Parodi, M.~Lagasio, M.~Maugeri, B.~Turato, and W.~Gallus.
\newblock Observational and {{Modelling Study}} of a {{Major Downburst Event}} in {{Liguria}}: {{The}} 14 {{October}} 2016 {{Case}}.
\newblock \emph{Atmosphere}, 10\penalty0 (12):\penalty0 788, 2019.
\newblock \doi{10.3390/atmos10120788}.

\bibitem[Pauluis(2008)]{pauluis2008}
O.~Pauluis.
\newblock Thermodynamic {{Consistency}} of the {{Anelastic Approximation}} for a {{Moist Atmosphere}}.
\newblock \emph{Journal of the Atmospheric Sciences}, 65\penalty0 (8), 2008.
\newblock \doi{10.1175/2007JAS2475.1}.

\bibitem[Pressel et~al.(2015)Pressel, Kaul, Schneider, Tan, and Mishra]{pressel2015}
K.~G. Pressel, C.~M. Kaul, T.~Schneider, Z.~Tan, and S.~Mishra.
\newblock Large-eddy simulation in an anelastic framework with closed water and entropy balances.
\newblock \emph{Journal of Advances in Modeling Earth Systems}, 7\penalty0 (3):\penalty0 1425--1456, 2015.
\newblock ISSN 1942-2466.
\newblock \doi{10.1002/2015MS000496}.
\newblock URL \url{https://onlinelibrary.wiley.com/doi/abs/10.1002/2015MS000496}.

\bibitem[Quarteroni et~al.(2015)Quarteroni, Manzoni, and Negri]{quarteroni2015reduced}
A.~Quarteroni, A.~Manzoni, and F.~Negri.
\newblock \emph{Reduced basis methods for partial differential equations: an introduction}, volume~92.
\newblock Springer, 2015.

\bibitem[Raonic et~al.(2023)Raonic, Molinaro, De~Ryck, Rohner, Bartolucci, Alaifari, Mishra, and de~B\'{e}zenac]{raonić2023convolutional}
B.~Raonic, R.~Molinaro, T.~De~Ryck, T.~Rohner, F.~Bartolucci, R.~Alaifari, S.~Mishra, and E.~de~B\'{e}zenac.
\newblock Convolutional neural operators for robust and accurate learning of pdes.
\newblock In A.~Oh, T.~Naumann, A.~Globerson, K.~Saenko, M.~Hardt, and S.~Levine, editors, \emph{Advances in Neural Information Processing Systems}, volume~36. Curran Associates, Inc., 2023.
\newblock URL \url{https://papers.nips.cc/paper_files/paper/2023/hash/f3c1951b34f7f55ffaecada7fde6bd5a-Abstract-Conference.html}.

\bibitem[Rezende and Mohamed(2015)]{rezende2016variational}
D.~Rezende and S.~Mohamed.
\newblock Variational inference with normalizing flows.
\newblock In F.~Bach and D.~Blei, editors, \emph{Proceedings of the 32nd International Conference on Machine Learning}, volume~37, Lille, France, 07--09 Jul 2015. PMLR.
\newblock URL \url{https://proceedings.mlr.press/v37/rezende15.html}.

\bibitem[Richter et~al.(2024)Richter, Nitzler, Pegolotti, Menon, Biehler, Wall, Schiavazzi, Marsden, and Pfaller]{richter2024bayesianwindkesselcalibrationusing}
J.~Richter, J.~Nitzler, L.~Pegolotti, K.~Menon, J.~Biehler, W.~A. Wall, D.~E. Schiavazzi, A.~L. Marsden, and M.~R. Pfaller.
\newblock Bayesian windkessel calibration using optimized 0d surrogate models, 2024.
\newblock URL \url{https://arxiv.org/abs/2404.14187}.

\bibitem[Ronneberger et~al.(2015)Ronneberger, Fischer, and Brox]{ronneberger2015unet}
O.~Ronneberger, P.~Fischer, and T.~Brox.
\newblock U-net: Convolutional networks for biomedical image segmentation.
\newblock In \emph{Medical Image Computing and Computer-Assisted Intervention -- MICCAI 2015}, pages 234--241, Cham, 2015. Springer International Publishing.
\newblock \doi{10.1007/978-3-319-24574-4_28}.

\bibitem[Ruckstuhl and Janjić(2020)]{ruckstuhl2020}
Y.~Ruckstuhl and T.~Janjić.
\newblock Combined {{State-Parameter Estimation}} with the {{LETKF}} for {{Convective-Scale Weather Forecasting}}.
\newblock \emph{Monthly Weather Review}, 148\penalty0 (4):\penalty0 1607--1628, 2020.
\newblock \doi{10.1175/MWR-D-19-0233.1}.

\bibitem[Segers et~al.(2008)Segers, Rietzschel, De~Buyzere, Stergiopulos, Westerhof, Van~Bortel, Gillebert, and Verdonck]{segers2008three}
P.~Segers, E.~Rietzschel, M.~De~Buyzere, N.~Stergiopulos, N.~Westerhof, L.~Van~Bortel, T.~Gillebert, and P.~Verdonck.
\newblock Three-and four-element windkessel models: assessment of their fitting performance in a large cohort of healthy middle-aged individuals.
\newblock \emph{Proceedings of the Institution of Mechanical Engineers, Part H: Journal of Engineering in Medicine}, 222\penalty0 (4):\penalty0 417--428, 2008.
\newblock \doi{10.1243/09544119JEIM287}.

\bibitem[Seidman et~al.(2022)Seidman, Kissas, Perdikaris, and Pappas]{seidman2022nomad}
J.~Seidman, G.~Kissas, P.~Perdikaris, and G.~J. Pappas.
\newblock {NOMAD}: Nonlinear manifold decoders for operator learning.
\newblock In S.~Koyejo, S.~Mohamed, A.~Agarwal, D.~Belgrave, K.~Cho, and A.~Oh, editors, \emph{Advances in Neural Information Processing Systems}, volume~35. Curran Associates, Inc., 2022.
\newblock URL \url{https://papers.nips.cc/paper_files/paper/2022/hash/24f49b2ad9fbe65eefbfd99d6f6c3fd2-Abstract-Conference.html}.

\bibitem[Seidman et~al.(2023)Seidman, Kissas, Pappas, and Perdikaris]{seidman2023variational}
J.~H. Seidman, G.~Kissas, G.~J. Pappas, and P.~Perdikaris.
\newblock Variational autoencoding neural operators.
\newblock In \emph{Proceedings of the 40th International Conference on Machine Learning}, volume 202. PMLR, 23--29 Jul 2023.
\newblock URL \url{https://proceedings.mlr.press/v202/seidman23a.html}.

\bibitem[Sherwin et~al.(2003)Sherwin, Formaggia, Peiro, and Franke]{sherwin2003computational}
S.~Sherwin, L.~Formaggia, J.~Peiro, and V.~Franke.
\newblock Computational modelling of 1{D} blood flow with variable mechanical properties and its application to the simulation of wave propagation in the human arterial system.
\newblock \emph{International Journal for Numerical Methods in Fluids}, 43\penalty0 (6-7):\penalty0 673--700, 2003.
\newblock \doi{10.1002/fld.543}.

\bibitem[Shi et~al.(2024)Shi, Gao, Ross, and Azizzadenesheli]{shi2024universal}
Y.~Shi, A.~F. Gao, Z.~E. Ross, and K.~Azizzadenesheli.
\newblock Universal functional regression with neural operator flows.
\newblock In \emph{Advances in Neural Information Processing Systems:Workshop on Bayesian Decision-making and Uncertainty}, 2024.
\newblock URL \url{https://openreview.net/forum?id=ZCtnWuaZiI}.

\bibitem[Sohl-Dickstein et~al.(2015)Sohl-Dickstein, Weiss, Maheswaranathan, and Ganguli]{Sohl2015diffusion}
J.~Sohl-Dickstein, E.~Weiss, N.~Maheswaranathan, and S.~Ganguli.
\newblock Deep unsupervised learning using nonequilibrium thermodynamics.
\newblock In \emph{Proceedings of the 32nd International Conference on Machine Learning}, volume~37 of \emph{Proceedings of Machine Learning Research}, Lille, France, 07--09 Jul 2015. PMLR.
\newblock URL \url{https://proceedings.mlr.press/v37/sohl-dickstein15.html}.

\bibitem[Sokolow et~al.(2024)Sokolow, Kissas, Beeche, Swago, Thompson, Viswanadha, Chirinos, Damrauer, Perdakaris, Rader, and Witschey]{sokolow2024aortic}
R.~Sokolow, G.~Kissas, C.~Beeche, S.~Swago, E.~W. Thompson, M.~Viswanadha, J.~Chirinos, S.~Damrauer, P.~Perdakaris, D.~J. Rader, and W.~R. Witschey.
\newblock An aortic hemodynamic fingerprint reduced order modeling analysis reveals traits associated with vascular disease in a medical biobank.
\newblock \emph{bioRxiv}, 2024.
\newblock \doi{10.1101/2024.04.19.590260}.

\bibitem[Straka et~al.(1993)Straka, Wilhelmson, Wicker, Anderson, and Droegemeier]{straka1993}
J.~M. Straka, R.~B. Wilhelmson, L.~J. Wicker, J.~R. Anderson, and K.~K. Droegemeier.
\newblock Numerical solutions of a non-linear density current: {{A}} benchmark solution and comparisons.
\newblock \emph{International Journal for Numerical Methods in Fluids}, 17\penalty0 (1):\penalty0 1--22, 1993.
\newblock \doi{10.1002/fld.1650170103}.

\bibitem[Tong et~al.(2024)Tong, {Sing Long}, and Schiavazzi]{tong2023invaert}
G.~G. Tong, C.~A. {Sing Long}, and D.~E. Schiavazzi.
\newblock {InVAErt} networks: A data-driven framework for model synthesis and identifiability analysis.
\newblock \emph{Computer Methods in Applied Mechanics and Engineering}, 423:\penalty0 116846, 2024.
\newblock \doi{https://doi.org/10.1016/j.cma.2024.116846}.

\bibitem[Wang et~al.(2023)Wang, Chen, Luo, and Wang]{wang2023conditional}
R.~Wang, Z.~Chen, Q.~Luo, and F.~Wang.
\newblock A conditional denoising diffusion probabilistic model for radio interferometric image reconstruction.
\newblock In \emph{Proceedings of the 11th International Conference on Artificial Intelligence and Applications (FAIA)}, 2023.
\newblock \doi{10.3233/FAIA230554}.

\bibitem[Wehenkel et~al.(2024)Wehenkel, Behrmann, Miller, Sapiro, Sener, Cameto, and Jacobsen]{simulation-based-inference}
A.~Wehenkel, J.~Behrmann, A.~C. Miller, G.~Sapiro, O.~Sener, M.~C. Cameto, and J.-H. Jacobsen.
\newblock Simulation-based inference for cardiovascular models.
\newblock In \emph{NeurIPS Workshop}, 2024.
\newblock URL \url{https://arxiv.org/abs/2307.13918}.

\bibitem[Wildberger et~al.(2023)Wildberger, Dax, Buchholz, Green, Macke, and Sch\"{o}lkopf]{dax2023flow}
J.~Wildberger, M.~Dax, S.~Buchholz, S.~Green, J.~H. Macke, and B.~Sch\"{o}lkopf.
\newblock Flow matching for scalable simulation-based inference.
\newblock In \emph{Advances in Neural Information Processing Systems}, volume~36. Curran Associates, Inc., 2023.
\newblock URL \url{https://papers.nips.cc/paper_files/paper/2023/hash/3663ae53ec078860bb0b9c6606e092a0-Abstract-Conference.html}.

\bibitem[Xiong et~al.(2022)Xiong, Cai, and Li]{xiong2021clustered}
J.~Xiong, X.~Cai, and J.~Li.
\newblock Clustered active-subspace based local gaussian process emulator for high-dimensional and complex computer models.
\newblock \emph{Journal of Computational Physics}, 450, 2022.
\newblock \doi{10.1016/j.jcp.2021.110840}.

\bibitem[Zhang et~al.(2022)Zhang, Zhang, and Lin]{zhang2022physics}
J.~Zhang, S.~Zhang, and G.~Lin.
\newblock {PAGP}: A physics-assisted gaussian process framework with active learning for forward and inverse problems of partial differential equations, 2022.
\newblock URL \url{http://arxiv.org/abs/2204.02583}.

\end{thebibliography}

\clearpage
\appendix
\section{Supplementary Material}

\counterwithin{figure}{section}
\setcounter{figure}{0} 

\subsection{Training and Architecture Details}
\label{sec:training details appendix}
All experiments were performed on a Nvidia GeForse RTX 3090 GPU with 24GB of memory. The number of training, validation, and test samples are provided in Table \ref{tab:data details}, along with the data dimensions. Further details regarding the experimental data are provided in Sections \ref{sec: appendix pwp} and \ref{sec:appendix acb}. For each experiment, we performed a hyperparameter sweep when possible to select the learning rate, scheduler rate, weight decay, and network size. Due to the large number of hyperparameters, we selected 64 random combinations of hyperparameters and trained each model for 500 epochs. Following the sweep, we selected the model with the best performance and trained it for 2000 epochs, saving it at the epoch with the best performance on the validation set. Following this, we evaluated its performance on the test samples. A special exception must be made for the VAE-based models, InVAErt and cVANO. InVAErt was particularly susceptible to posterior collapse. Following the advice of the authors in Section 2.3.2 of \citep{tong2023invaert}, we hand-tuned the model to prevent over-parameterization and used early stopping. Similarly, cVANO was susceptible to training instability and exploding gradients, which also necessitated careful tuning and training in order to get the best results. The final size of each model, in number of network parameters, is provided in Table \ref{tab:model sizes}. The results from the model selection revealed all models are fairly consistent in size, typically all on the same order of magnitude. 

We observed some sensitivity to the normalization scheme used on the data. In the PWP experiment, we normalize each input channel to a range between zero and one by dividing by the maximum value across all samples at that channel. In the ACB experiment, we normalize each channel by taking the minimum and maximum value across all samples in that channel and rescaling these to a range between zero and one.

The training time per epoch of all experiments is consistent, ranging from 1.5 to 2 seconds. During the hyperparameter sweep, each model is trained for 500 epochs, taking approximately 12.5 minutes. The best model from the sweep is trained for a total of 2000 epochs, taking approximately 1 hour. Throughout this entire project, we estimate the total number of GPU-hours for all model hyper-parameter sweeps, hand-tuning, and training of the final model to be approximately 180 GPU-hours. 

\paragraph{Additional Training Approaches}
Although the two objectives of FUSE are decoupled during training time, it is also possible to backpropagate losses on the outputs through both the forward and inverse model components. FMPE supports differentiable sampling, therefore the loss on the outputs $s$ could be formulated as a function of continuous inputs $u$, 
\begin{equation}
    \mathcal{L}_3(\theta, \phi) = 
    \mathcal{L}_1 (\theta;~\Gct \circ \rho^\phi(.|.)) 
    = \int_{\mathcal{U}} \left\lVert \mathcal{G}^\theta(\rho^\phi(\xi| u)) - \mathcal{G}(\rho(\xi^* | u)) \right\rVert_{L^{1}} \, du.
\end{equation}
In the case where the inputs and outputs, respectively $u$ and $s$, are identical, a \emph{reconstruction} loss may be applied to further train the inverse model component. This entails predicting outputs $s$ from the data-generating parameters $\xi^*$ and predicting the inverse problem from the predictions $s$, as opposed to the original inputs $u$. This is formulated as 
\begin{equation}
    \mathcal{L}_{r}(\phi, \theta)
    = \mathcal{L}_2(\phi;~\hat{u}=T\circ \Gct(\xi^*))
    = \mathcal{L}^{FMPE}(\phi;~\hat{u}=T\circ \Gct(\xi^*)).
\end{equation}
Although each approach has the potential to aid training it is still required to train each objective separately. Likewise, we did not observe improvements in our experiments by implementing these additional losses. This further supports our claim in Equation \ref{eq:triangle inequality}, that model components trained by decoupled objectives for forward and inverse problems may be \emph{unified} at evaluation to propagate uncertainties in the parameters to outputs and provide interpretable results for complex and dynamic systems of PDEs.

\begin{table}[ht]
    \centering
    \caption{Data details. $N_{train}$, $N_{val}$, and $N_{train}$ provide the number of training, validation, and test samples. The number of scalar parameters of the PDE system is given as $m$. "Channels" describe the number of continuous input components (locations/variables), and the number of points in the time discretization of the continuous inputs and outputs is given by "Sample Points". The "Batch Size" is the number of training samples per mini-batch during training.}
    \begin{tabular}{c c c c c c c c c}
    \toprule
        Experiment & $N_{train}$ & $N_{val}$ & $N_{test}$ & $m$ & Channels & Sample Points & Batch Size \\
        \midrule
        PWP  & 4000 & 128 & 128 & 32 & 52 & 487 & 64 \\
        ACB  & 8000 & 1000 & 1000 & 6 & 20 & 181 & 64 \\
        \bottomrule
    \end{tabular}
    \label{tab:data details}
\end{table}

\begin{table}
    \centering
    \caption{Number of network parameters (millions) per model for the experiments.}
    \begin{tabular}{c c c}
    \toprule
        Model               & PWP      & ACB  \\
        \midrule
        FUSE \textbf{(ours)}& 7.2          & 5.8 \\
        UNet                & 14.1         & 9.8 \\
        cVANO               & 5.7          & 1.8 \\
        GAROM               & 11.1         & 2.7 \\
        InVAErt             & 13.6         & 2.9 \\
        \bottomrule
    \end{tabular}
    \label{tab:model sizes}
\end{table}

\begin{table}[!ht]
    \centering
    \captionsetup{font=normal}
    
    \caption{Evaluating inference time, CRPS, and relative L1 errors over continuous outputs for a trained model as the number of FMPE  and Conditional DDPM samples increase. Conditional DDPM is implemented as described \cite{letafati2024} for a 1D problem using 32 diffusion steps. We observe that the FMPE method results to higher accuracy in both metrics and slight increases in accuracy up until 512 samples, while DDPM is fairly consistent. Sampling times are approximately equivalent between the two models.}
    \resizebox{\textwidth}{!}{%
    {
    \begin{tabular}{c c c cc cc cc}
        \toprule
        & Posterior Samples & Forw. Pass (s) & \multicolumn{2}{c}{Inverse Pass (s)} & \multicolumn{2}{c}{CRPS$\times 10^2$} & \multicolumn{2}{c}{$L_1$ Error} \\[\medskipamount]
        & & FNO & FMPE & DDPM & FMPE & DDPM & FMPE & DDPM \\[\medskipamount]
        \midrule
        \multirow{8}{*}[-0ex]{\rotatebox{90}{\textbf{case: ACB}}}
        & 128    & 0.0018 & 0.179  & 0.108 & $1.84 \pm 1.63$ & $4.96  \pm 2.23$ & $0.58 \pm 0.58$ & $1.61 \pm 0.91$  \\
        & 256    & 0.0030 & 0.194  & 0.114 & $1.64 \pm 1.63$ & $4.75  \pm 2.25$ & $0.58 \pm 0.58$ & $1.60 \pm 0.90$  \\
        & 512    & 0.0059 & 0.211  & 0.114 & $1.54 \pm 1.64$ & $4.61  \pm 2.16$ & $0.58 \pm 0.58$ & $1.60 \pm 0.90$  \\
        & 1024   & 0.0127 & 0.240  & 0.113 & $1.49 \pm 1.64$ & $4.57  \pm 2.20$ & $0.58 \pm 0.58$ & $1.59 \pm 0.89$  \\
        & 2048   & 0.0239 & 0.495  & 0.152 & $1.46 \pm 1.64$ & $4.54  \pm 2.21$ & $0.58 \pm 0.58$ & $1.59 \pm 0.89$  \\
        & 4096   & 0.0608 & 0.973  & 0.285 & $1.45 \pm 1.64$ & $4.53  \pm 2.21$ & $0.58 \pm 0.58$ & $1.60 \pm 0.90$  \\
        & 8192   & 0.1205 & 1.820  & 0.550 & $1.44 \pm 1.64$ & $4.52  \pm 2.20$ & $0.58 \pm 0.58$ & $1.59 \pm 0.89$  \\
        \bottomrule
    \end{tabular}
    }
    }
    \label{tab:ddpm vs fmpe}
\end{table}

\begin{table}[!ht]
    \centering
    {
        \captionsetup{font=normal}
    \caption{Evaluating training time, CRPS, and relative L1 errors for models as the number of training data increases. As more training samples are used, CRPS and relative L1 error improve; however, the model converges after several thousand training samples resulting in smaller performance gains per new training sample.}

    \begin{tabular}{c c c c c }
        \toprule
        & Training Data & Training Time Per Epoch (s) & CRPS$\times 10^2$ & $L_1$ Error \\[\medskipamount]
        \midrule
        \multirow{8}{*}[-0ex]{\rotatebox{90}{\textbf{case: ACB}}}
        & 128  & 0.027 & $16.10 \pm 4.53$ & $5.70 \pm 2.34$  \\
        & 256  & 0.059 & $12.61 \pm 4.85$ & $4.22 \pm 2.29$ \\
        & 512  & 0.118 & $10.23 \pm 4.31$ & $2.51 \pm 1.60$ \\
        & 1024 & 0.213  & $8.48 \pm 4.20$ & $1.71 \pm 1.30$ \\
        & 2048 & 0.450  & $5.88 \pm 3.47$ & $1.13 \pm 0.95$ \\
        & 4096 & 0.973  & $3.90 \pm 2.52$ & $0.84 \pm 0.71$ \\
        & 8192 & 1.698  & $1.84 \pm 1.63$ & $0.58 \pm 0.60$ \\
        \bottomrule
    \end{tabular}
    \label{tab:fmpe data scaling}
    }
\end{table}

\subsection{Evaluation Metrics}
\label{sec:evaluation metrics}
\paragraph{Continuous Ranked Probability Score} We define the \emph{Continuous Ranked Probability Score} (CRPS) and present how it compares a single ground truth value $y$ to a Cumulative Distribution Function (CDF) $F$, quantifying how well a distribution of samples fits a single target. The CRPS is defined as
\begin{equation}
    CRPS(F, y) = \mathbb{E}\left[|X-y|\right] - \frac{1}{2}\cdot \mathbb{E} \left[X-X'\right],
\end{equation}
with independent and identically distributed random variables $X$ and $X'$ following the distribution dictated by $F$. The first expectation computes the Mean Absolute Error of the samples compared to the target value, and the second expectation penalizes tight distributions which are far from the true value. A CRPS of zero indicates that the distribution perfectly fits the target sample by returning a Dirac measure.

\paragraph{$L^p$ Error} We employ the relative $L^1$ and $L^2$ error for evaluating predictions of time series data. This metric is calculated by 
\begin{equation}
    \frac{||\mathcal{G}(\xi) - \mathcal{G}^{\theta}(\xi)||_p}{||\mathcal{G}(\xi)||_p},
\end{equation}
where $\mathcal{G}(\xi)$ is the true function described by the system of PDEs, and $\mathcal{G}^{\theta}(\xi)$ is a prediction produced by the network. 

\clearpage

\subsection{Pulse Wave Propagation: Reduced Order Model for Blood Flow}
\label{sec: appendix pwp}

The reduced order Navier-Stokes model considers a system of discrete compliant tubes and arterial segments, connected at points \cite{alastruey2012arterial}. The length of the vessels is considered much larger than the local curvature and the wave propagation happens on the axial direction. Therefore, their properties can be described using a Cartesian coordinate $x$ \cite{sherwin2003computational}. The luminal cross-sectional area is defined as $A(x, t) = \int_A d \sigma$, the average velocity is defined as $U(x,t) = \frac{1}{A} \int_S \mathbf{u} (\mathbf{x} ,t ) \cdot \mathbf{n} d \sigma$ and the volume flux at a given cross-section as $Q(x,t) = U A$ \cite{alastruey2012arterial}. The arteries are allowed to deform in the radial direction due to the internal pressure $P(x,t)$ which is considered constant over a cross-section and the wall is considered impermeable. The blood is considered as an incompressible Newtonian fluid, with viscosity $\mu = 2.5  mPa \ s$ and density $\rho = 1060 kg \ m^{-3}$ \cite{ formaggia2010cardiovascular}. All the gravitational effects are ignored assuming a patient in a supine position \cite{alastruey2012arterial}. Using the above assumptions, the pulse wave propagation is formulated as the following system of hyperbolic constitutive laws \cite{alastruey2012arterial,formaggia2010cardiovascular}:
\begin{equation}
    \begin{split}
        &\frac{\partial A}{\partial t} + \frac{\partial AU}{\partial x} = 0, \\
        &\frac{\partial U}{\partial t} + (2\alpha - 1) U \frac{\partial U }{\partial x} + (\alpha -1) \frac{U^2}{A} \frac{\partial A}{\partial x} + \frac{1}{\rho} \frac{\partial P}{ \partial x} = \frac{f}{\rho A},
    \end{split}
\end{equation}
where $a(x,t) = \frac{1}{A U^2} \int_A \mathbf{u}^2 d \sigma$ the Coriolis coefficient that accounts for the non-linearity of the integration over the cross-section and describes the shape of the velocity profile. The velocity profile is commonly considered as axisymmetric, constant, and satisfying the non-slip boundary conditions on the arterial walls. An example of such profile is \cite{alastruey2012arterial}:
\begin{equation}
    u(x, r, t) = U \frac{\zeta + 2}{\zeta} \Big [ 1 - (\frac{r}{R})^\zeta \Big]
\end{equation}
where $r$ is the radial coordinate, $R(x,t)$ the lumen radius and $\zeta = \frac{2 - \alpha}{\alpha -1}$ a constant. A flat profile is defined by $\alpha=1$ \citep{charlton2019modeling}. Integrating the three-dimensional Navier-Stokes equations for the axi-symmetric flow provides the friction force per unit length $f (x,t) = -2 (\zeta + 2) \mu \pi U$. To account for the fluid-structure interaction of the problem, a pressure area relation is derived by considering the tube law. For a thin isotropic homogenous and incompressible vessel that deforms axisymmetrically at each circular cross-section the following Voigt-type visco-elastic law \cite{vcanic2006modeling} is typically used:
\begin{equation}
    \begin{split}
        & P = P_e(A;x) + \frac{\Gamma(x)}{A_0(x) \sqrt{A} } \frac{\partial A}{\partial t}, \\
        & P_e(A,x) = P_{\text{ext}} + \frac{\beta(x)}{A_0(x)} \Big ( \sqrt{A} - \sqrt{A_0 (x)},\\
        & \beta(x) = \frac{4}{3} \sqrt{\pi} E(x) h(x), \quad \Gamma(x) = \frac{2}{3} \sqrt{\pi} \phi(x) h(x),
    \end{split}
\end{equation}
where $P_e$ the elastic pressure component, $h(x)$ the wall thickness, $E(x)$ the Young's modulus, $\phi(x)$ the wall viscosity, $\beta(x)$ a wall stiffness coefficient, and $\Gamma(x)$ the wall viscosity. The scalars $P_{ext}$ and $A_0$ correspond to the external pressure and equilibrium cross-sectional area, respectively. The parameters $\beta(x)$ and $\Gamma(x)$ are computed using empirical relations found in the literature \citep{charlton2019modeling}. For this problem, a pulse wave from the heart is considered as a boundary condition for the inlet of the artery. For merging and splitting of arteries, the conservation of mass and the continuity of dynamic and kinematic pressure are considered, as well as, no energy losses. For the outlet of the arterial system, boundary conditions that model the downstream circulation are considered. In \citep{charlton2019modeling}, the three-element Windkessel model \cite{segers2008three} is used:
\begin{equation}
    Q(1 + \frac{R_1}{R_2}) + C R_1 \frac{\partial Q}{\partial t} = \frac{P_e - P_{\text{out}}}{R_2} + C \frac{P_e}{t}
\end{equation}

In \cite{charlton2019modeling}, the authors consider algebraic relations from the literature that allows them to include different parameters, e.g. the volume flux, in the pulse wave propagation problem. For example, the digital PPG measurements can be related to the volume flux at the extremities as:
\begin{equation}
    \text{PPG}(t) = \int_0^T Q_{W}(t) - Q_{\text{out}}(t) dt,
\end{equation}
where $T$ is the duration of a cardiac cycle, $Q_W$ the flow entering the Windkessel model, and $Q_{\text{out}}$ the flow in the outlet. 

\begin{table}[ht]
\centering
\caption{Cardiovascular parameters used in the reduced order model for the blood flow pulse wave propagation.}
\begin{tabular}{l c l}
\toprule
\textbf{Parameter} & \textbf{Units} & \textbf{Description} \\
\midrule
Age & years & Age of the individual \\
HR & bpm & Heart Rate \\
SV & ml & Stroke Volume \\
CO & l/min & Cardiac Output \\
LVET & ms & Left Ventricular Ejection Time \\
dp/dt & mmHg/s & Rate of pressure change in the heart \\
PFT & ms & Pulse Transit Time \\
RFV & ml & Residual Filling Volume \\
SBP$_a$ & mmHg & Systolic Blood Pressure (arterial) \\
DBP$_a$ & mmHg & Diastolic Blood Pressure (arterial) \\
MAP$_a$ & mmHg & Mean Arterial Pressure \\
PP$_a$ & mmHg & Pulse Pressure (arterial) \\
SBP$_b$ & mmHg & Systolic Blood Pressure (brachial) \\
DBP$_b$ & mmHg & Diastolic Blood Pressure (brachial) \\
MBP$_b$ & mmHg & Mean Blood Pressure (brachial) \\
PP$_b$ & mmHg & Pulse Pressure (brachial) \\
PP$_{amp}$ & ratio & Pulse Pressure Amplification \\
AP & mmHg & Augmentation Pressure \\
AIx & \% & Augmentation Index \\
Tr & ms & Reflection Time \\
PWV$_a$ & m/s & Pulse Wave Velocity (arterial) \\
PWV$_{cf}$ & m/s & Pulse Wave Velocity (carotid-femoral) \\
PWV$_{br}$ & m/s & Pulse Wave Velocity (brachial-radial) \\
PWV$_{fa}$ & m/s & Pulse Wave Velocity (femoral-ankle) \\
dia$_{asca}$ & mm & Diameter of Ascending Aorta \\
dia$_{dta}$ & mm & Diameter of Descending Thoracic Aorta \\
dia$_{abda} $& mm & Diameter of Abdominal Aorta \\
dia$_{car}$ & mm & Diameter of Carotid Artery \\
Len & mm & Length \\
drop$_{fin}$ & mmHg & Pressure Drop at Finger \\
drop$_{ankle}$ & mmHg & Pressure Drop at Ankle \\
SVR & $10^6$ Pa s / m$^3$ & Systemic Vascular Resistance \\
\bottomrule
\end{tabular}
\label{tab:cv_parameters}
\end{table}
\begin{table}[ht]
\centering
\caption{Artery locations used in the reduced order model for the blood flow pulse wave propagation.}
\begin{tabular}{l l}
\toprule
\textbf{Artery Location} & \textbf{Description} \\
\midrule
Aortic Root & The portion of the aorta that is attached to the heart \\
Carotid & Major arteries in the neck leading to the brain, neck, and face \\
Thoracic Aorta & The part of the aorta that runs through the chest \\
Brachial & The major artery of the upper arm \\
Radial & Artery located in the forearm, commonly used to measure the pulse \\
Abdominal Aorta & The part of the aorta that runs through the abdomen \\
Iliac Bifurcation & The point where the aorta splits into the common iliac arteries \\
Common Iliac & Arteries that supply the pelvic organs and lower limbs \\
Femoral & The major artery supplying blood to the thigh and lower limbs \\
Anterior Tibial & Artery located in the lower leg \\
Sup. Middle Cerebral & Artery that supplies blood to the cerebral hemispheres \\
Sup. Temporal & Artery that supplies blood to parts of the face and scalp \\
Digital & Arteries that supply blood to the fingers\\
\bottomrule
\end{tabular}
\label{tab:artery_locations}
\end{table}

\begin{sidewaysfigure}[ht]
    \centering
    \begin{subfigure}[b]{0.3\textwidth}
        \centering
        \includegraphics[width=1.03\textwidth, trim=0 30 0 0, clip]{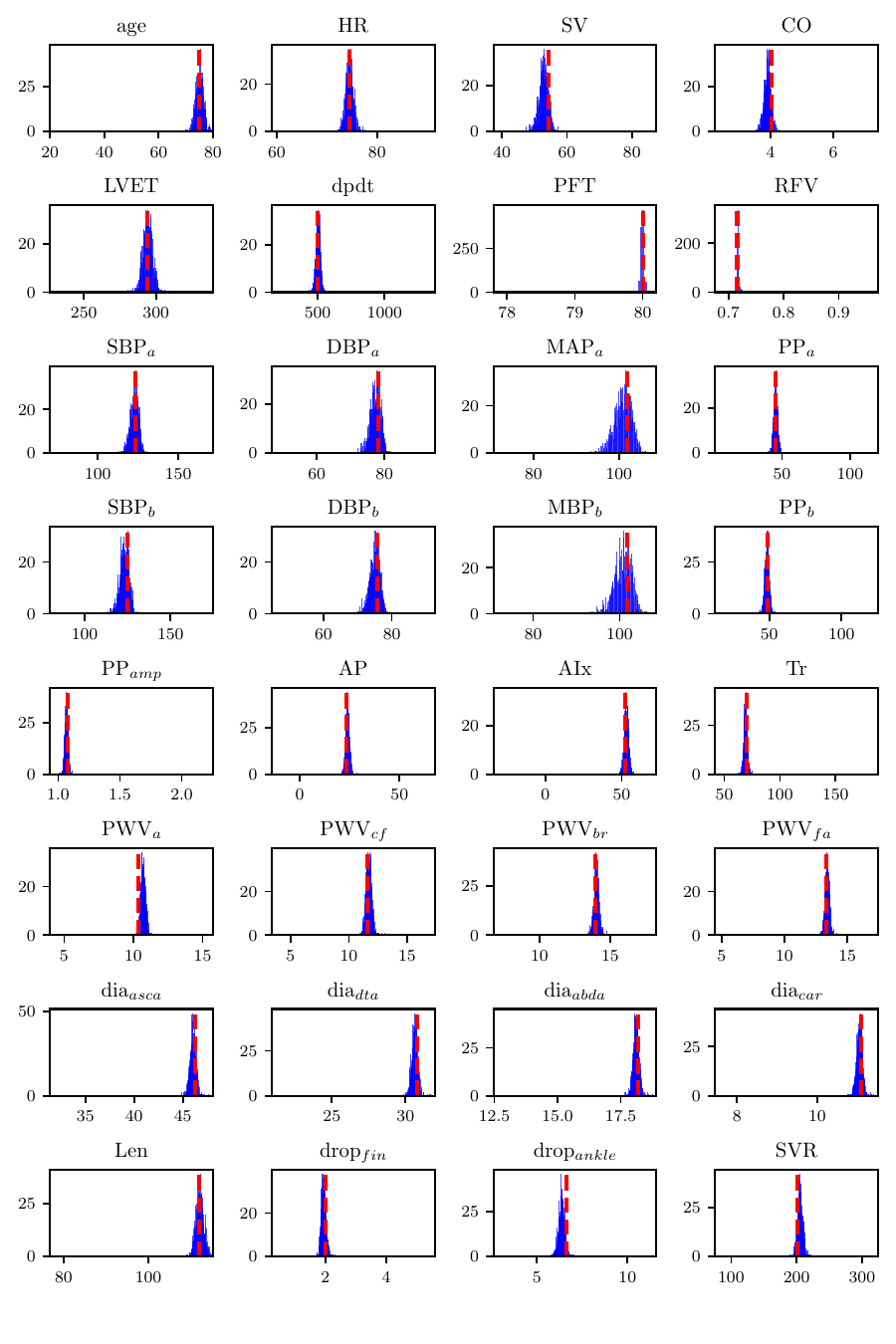}
        \caption{Information level 1}
    \end{subfigure}
    \hfill
    \begin{subfigure}[b]{0.3\textwidth}
        \centering
        \includegraphics[width=\textwidth]{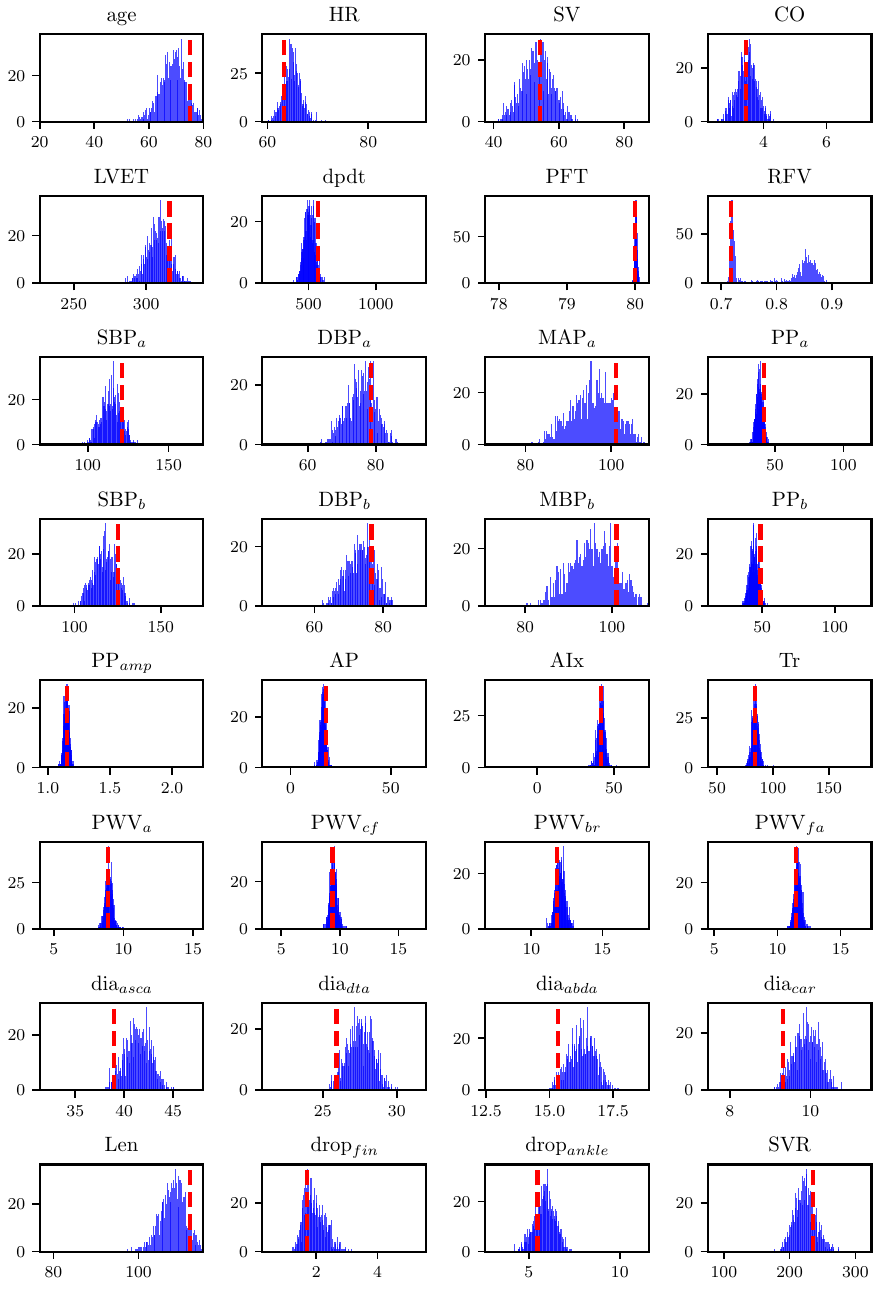}
        \caption{Information level 2}
    \end{subfigure}
    \hfill
    \begin{subfigure}[b]{0.3\textwidth}
        \centering
        \includegraphics[width=\textwidth]{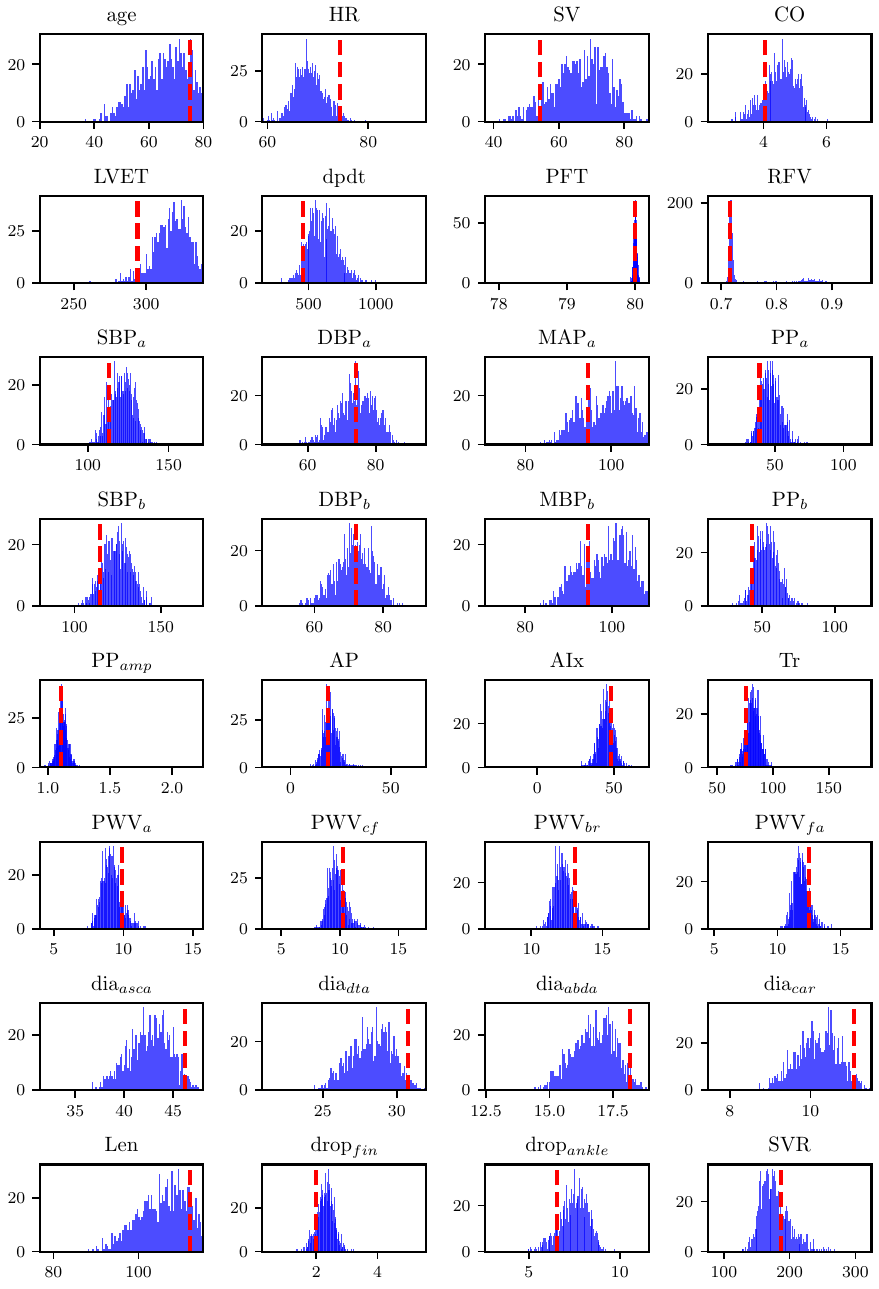}
        \caption{Information level 3}
    \end{subfigure}
    \caption{PWP, inverse problem: Histograms for the sample with the median CRPS for each level of available input information.}
    \label{fig:pwp median histograms}
\end{sidewaysfigure}

\begin{sidewaysfigure}[ht]
    \centering
    \begin{subfigure}[b]{0.3\textwidth}
        \centering
        \includegraphics[width=1.0\textwidth, trim=0 0 0 0, clip]{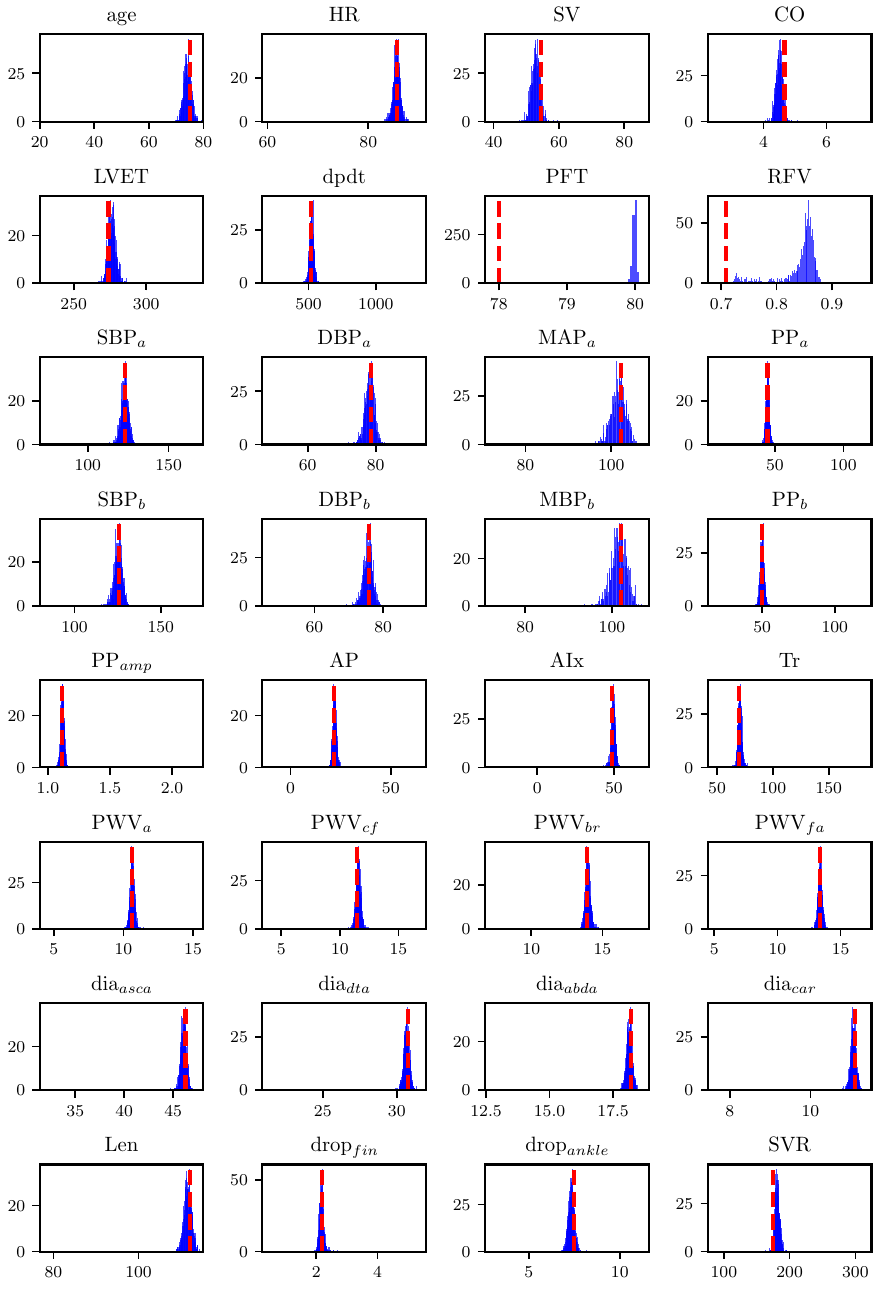}
        \caption{Information level 1}
    \end{subfigure}
    \hfill
    \begin{subfigure}[b]{0.3\textwidth}
        \centering
        \includegraphics[width=\textwidth]{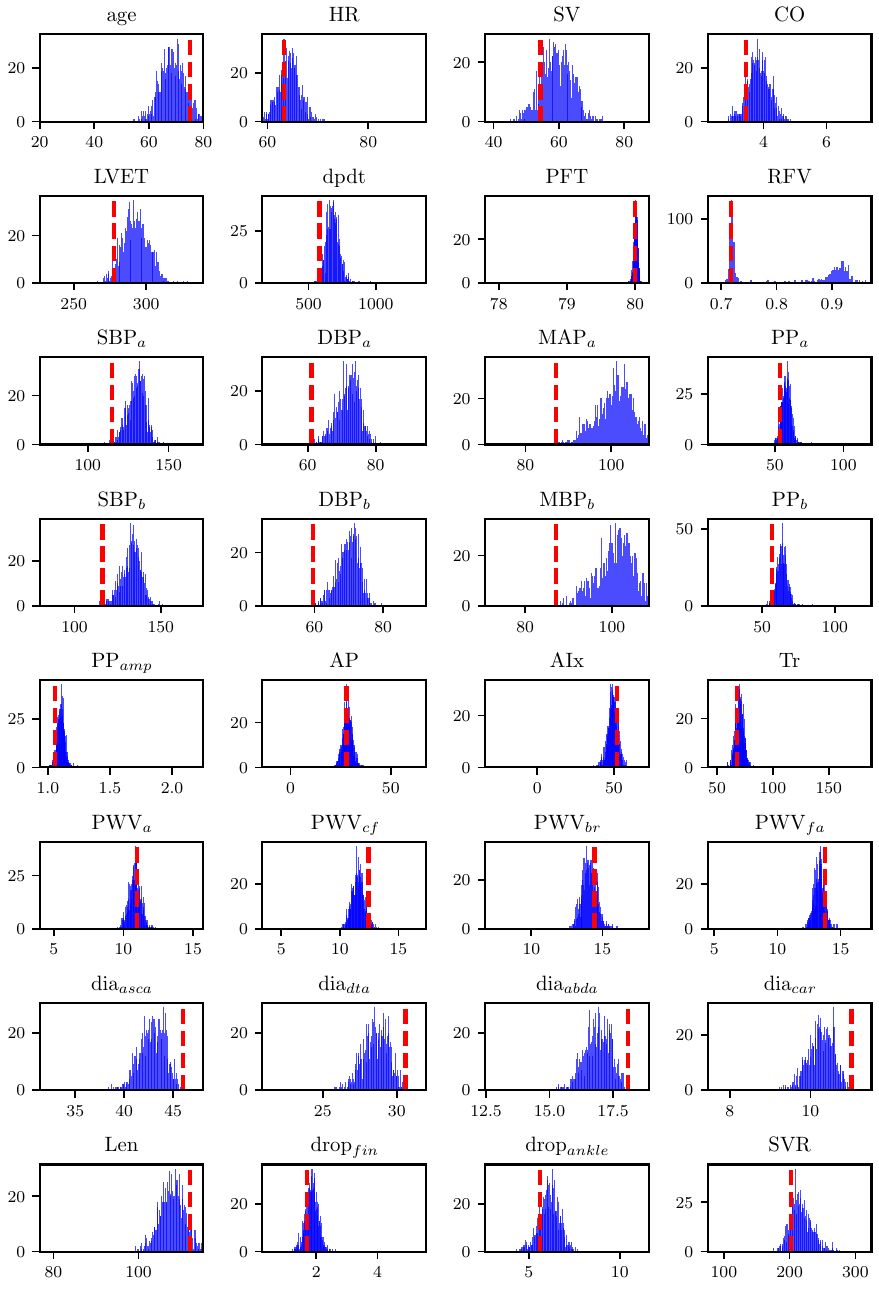}
        \caption{Information level 2}
    \end{subfigure}
    \hfill
    \begin{subfigure}[b]{0.3\textwidth}
        \centering
        \includegraphics[width=\textwidth]{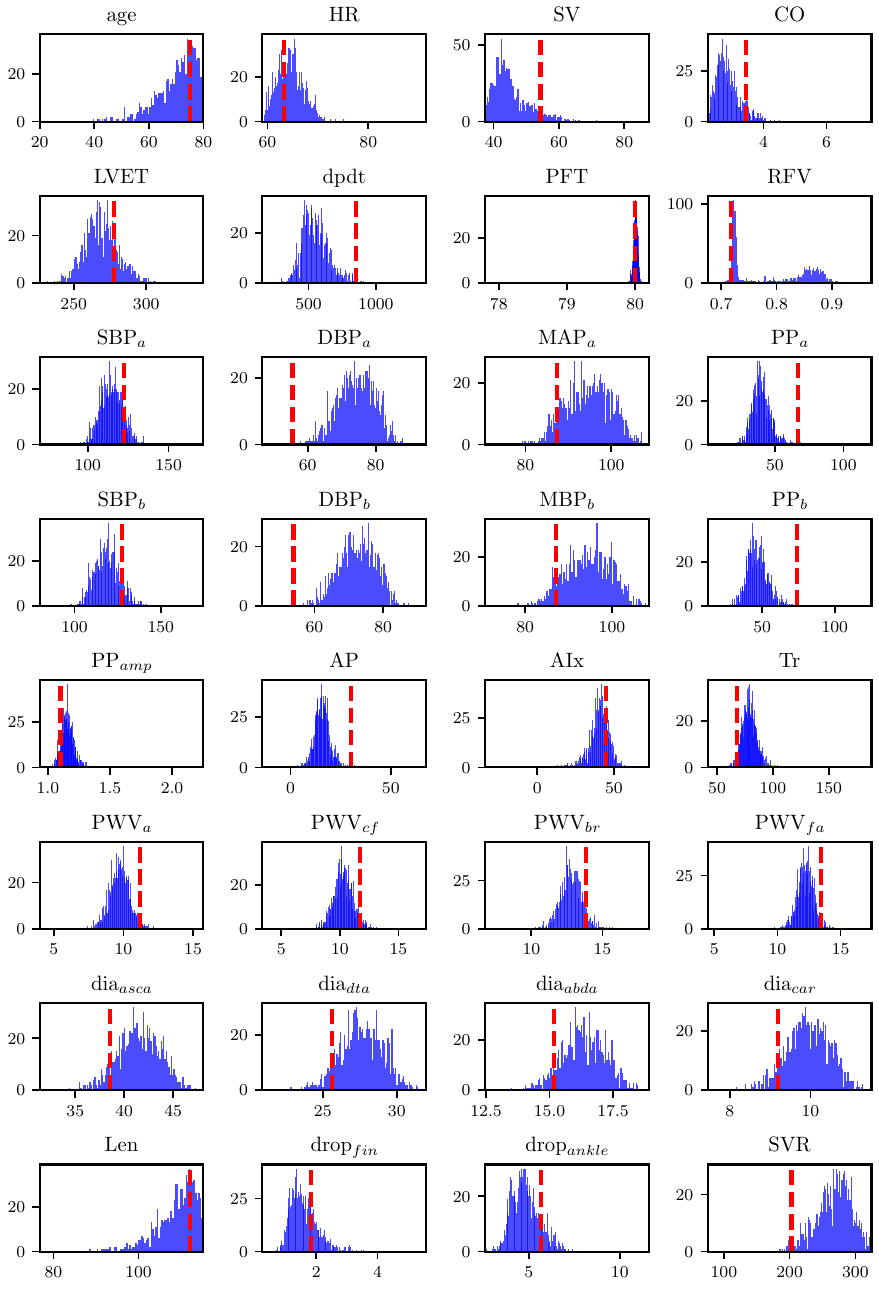}
        \caption{Information level 3}
    \end{subfigure}
    \caption{PWP, inverse problem: Histograms for the sample with the greatest CRPS (worst case) for each level of available input information.}
    \label{fig:pwp worst histograms}
\end{sidewaysfigure}

\begin{sidewaysfigure}[ht]
    \centering
    \begin{subfigure}[b]{0.3\textwidth}
        \centering
        \includegraphics[width=1.0\textwidth, trim=0 0 0 0, clip]{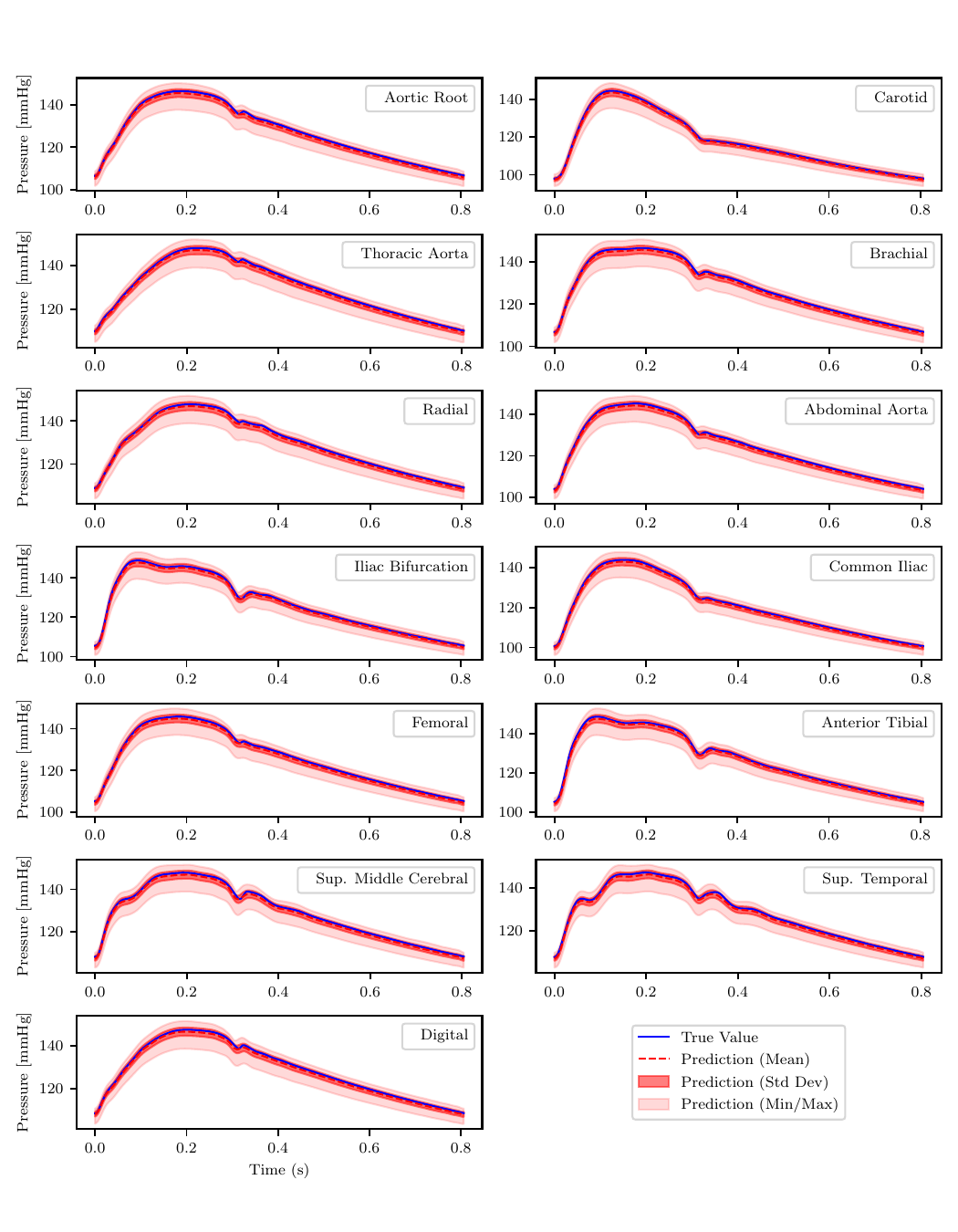}
        \caption{Information level 1}
    \end{subfigure}
    \hfill
    \begin{subfigure}[b]{0.3\textwidth}
        \centering
        \includegraphics[width=\textwidth]{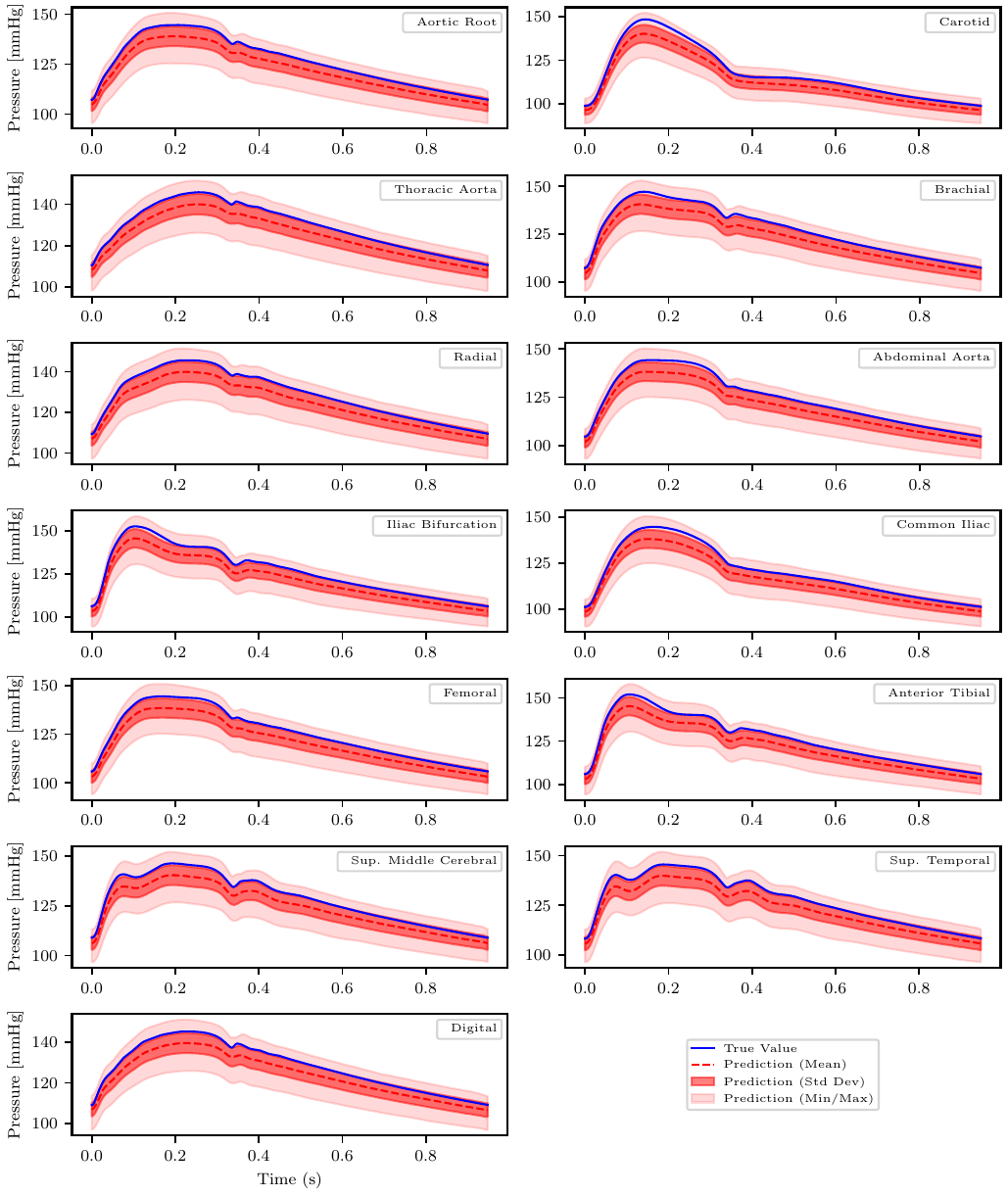}
        \caption{Information level 2}
    \end{subfigure}
    \hfill
    \begin{subfigure}[b]{0.3\textwidth}
        \centering
        \includegraphics[width=\textwidth]{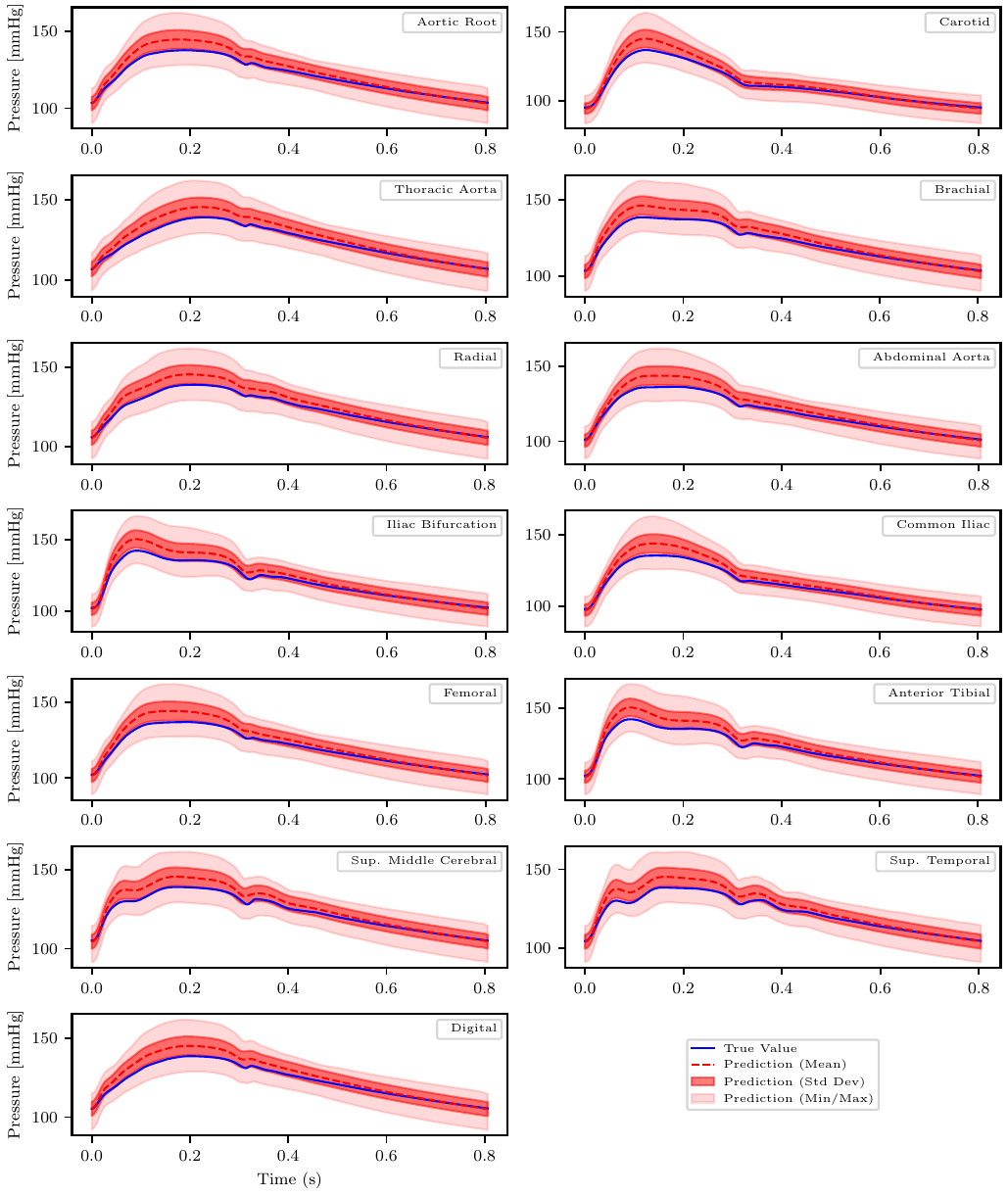}
        \caption{Information level 3}
    \end{subfigure}
    \caption{PWP, propagated uncertainty: Pressure time series for the sample with the median rel. $\mathcal{L}_1$ Error for each level of available input information.}
    \label{fig:pwp median waveforms}
\end{sidewaysfigure}

\begin{sidewaysfigure}[ht]
    \centering
    \begin{subfigure}[b]{0.3\textwidth}
        \centering
        \includegraphics[width=1.03\textwidth, trim=0 30 0 0, clip]{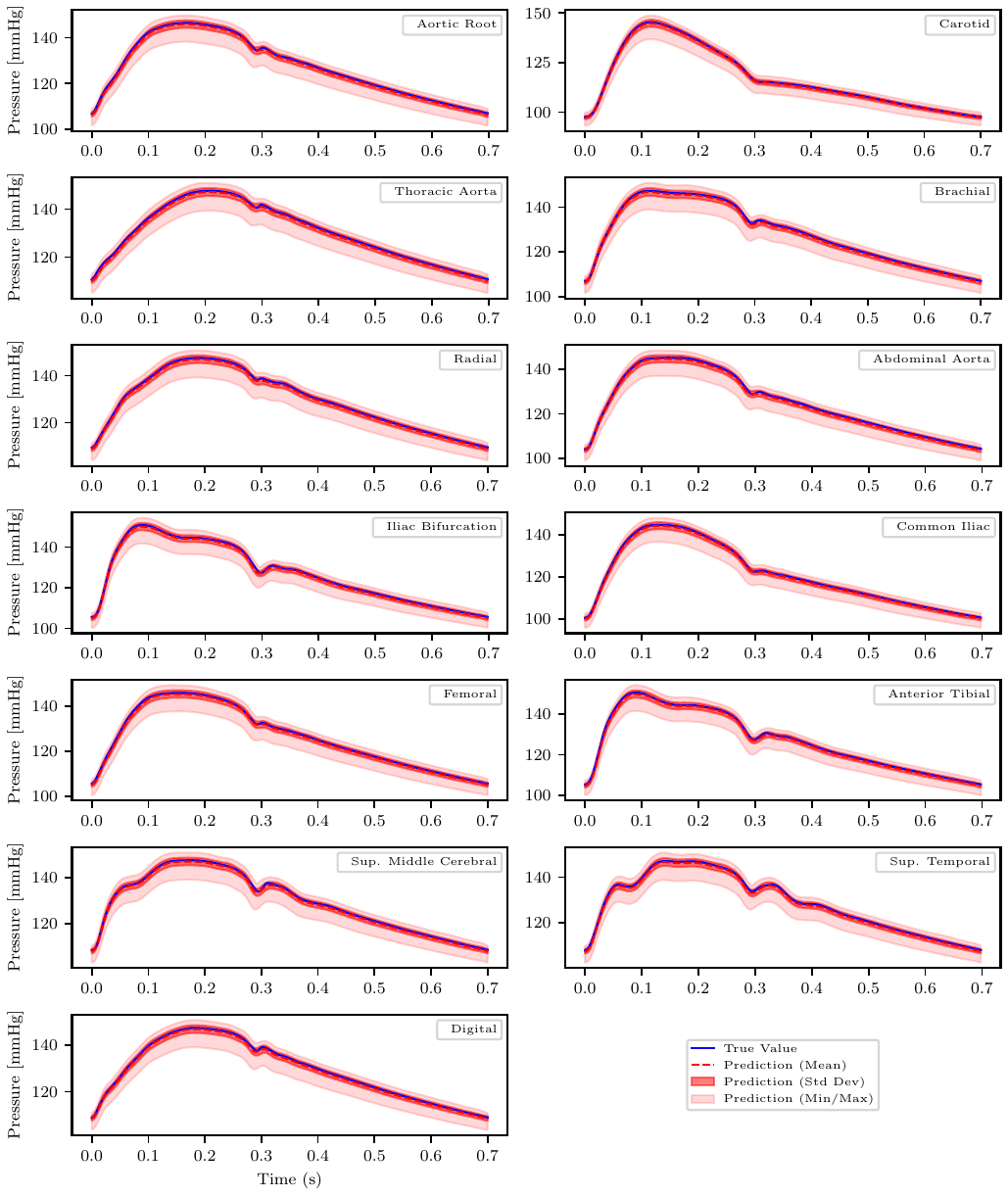}
        \caption{Information level 1}
    \end{subfigure}
    \hfill
    \begin{subfigure}[b]{0.3\textwidth}
        \centering
        \includegraphics[width=\textwidth]{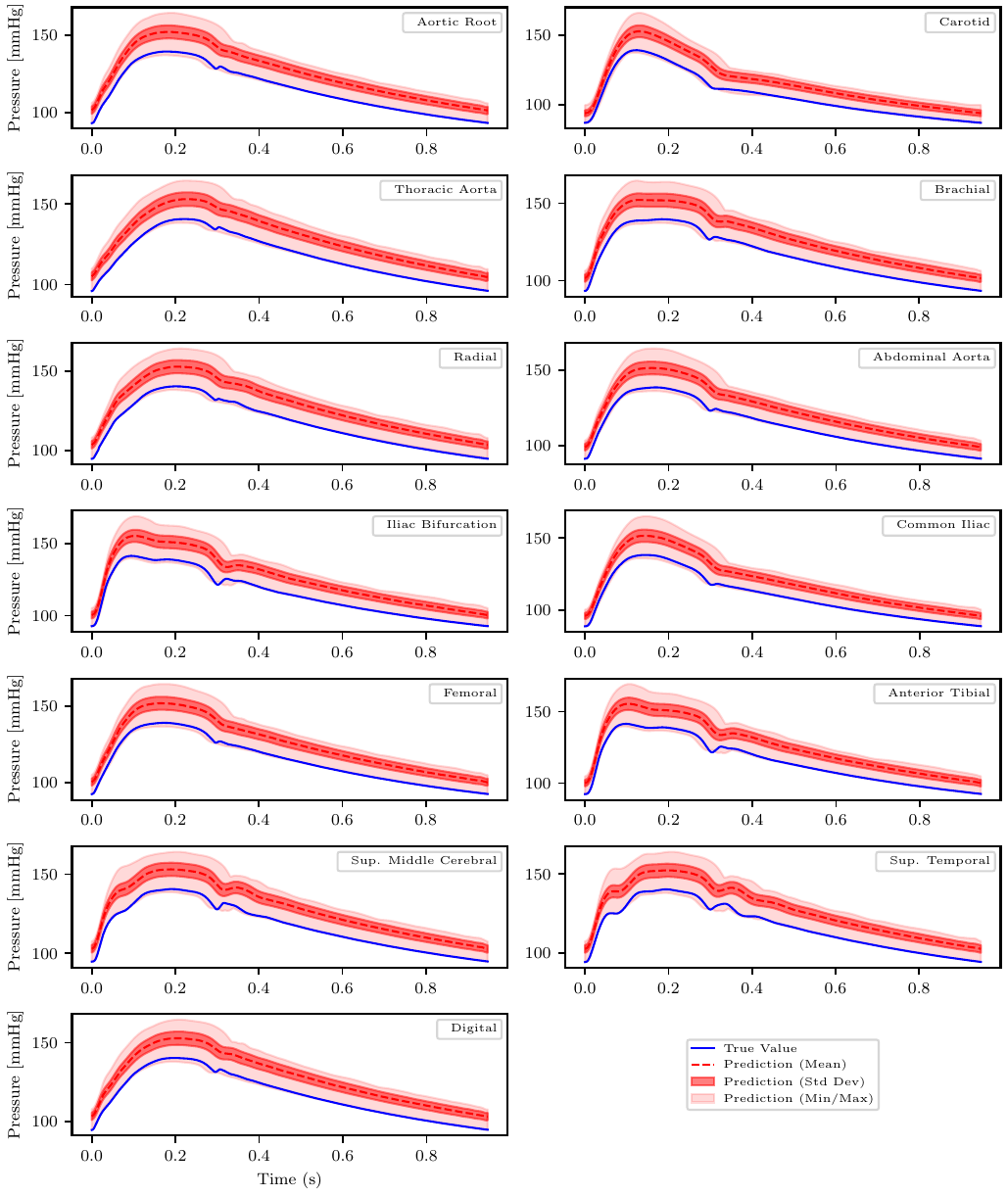}
        \caption{Information level 2}
    \end{subfigure}
    \hfill
    \begin{subfigure}[b]{0.3\textwidth}
        \centering
        \includegraphics[width=\textwidth]{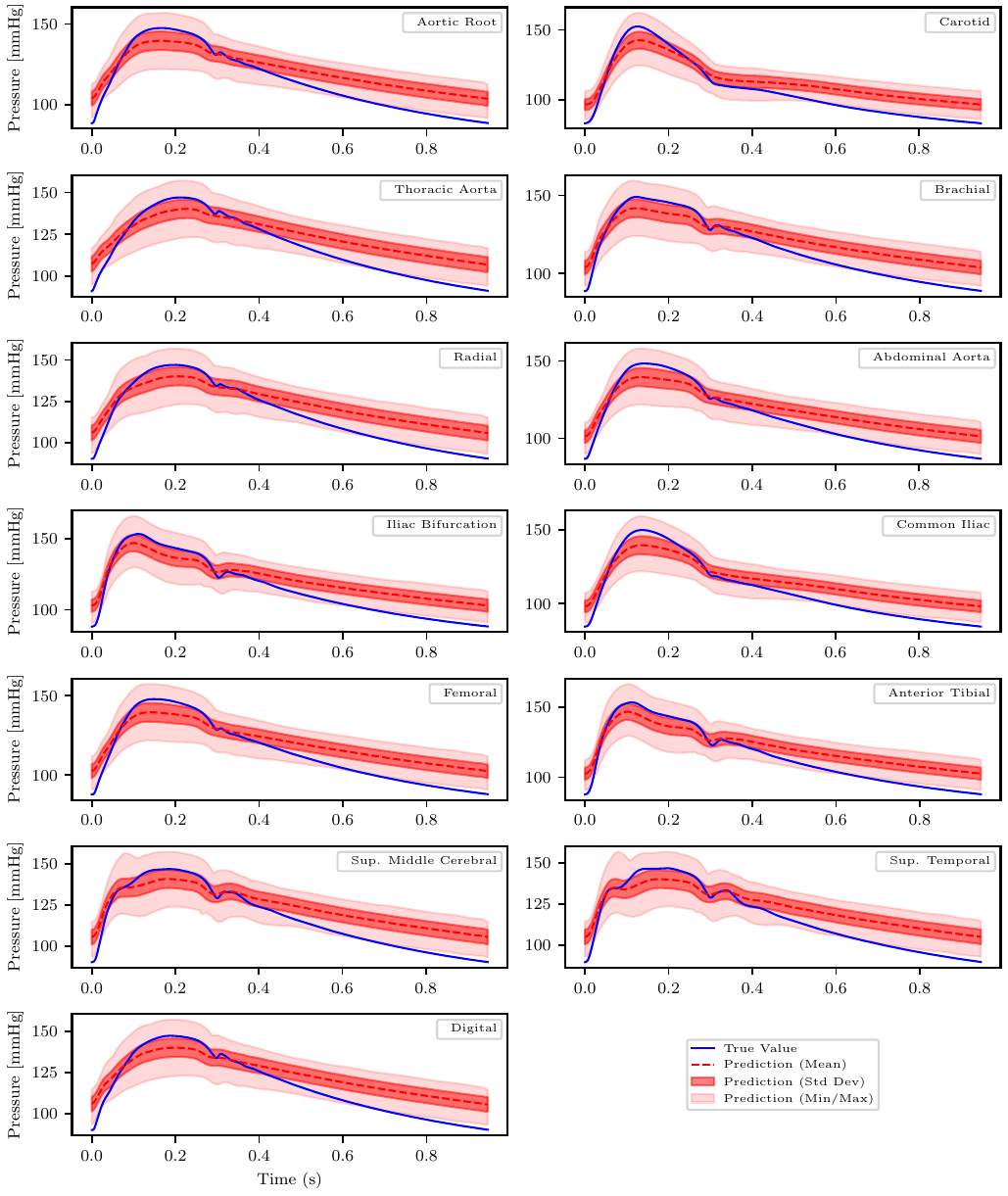}
        \caption{Information level 3}
    \end{subfigure}
    \caption{PWP, propagated uncertainty: Pressure time series for the sample with the greatest rel. $\mathcal{L}_1$ Error (worst case) each level of available input information.}
    \label{fig:pwp worst waveforms}
\end{sidewaysfigure}

\begin{figure}
	\centering
    \begin{subfigure}[b]{0.35\textwidth}
       \includegraphics[height=3.5cm]{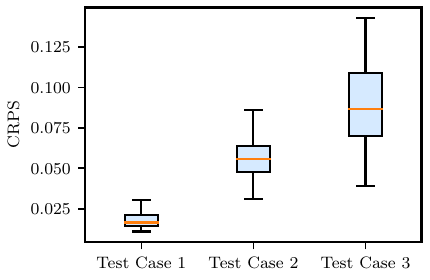}
        \caption{CRPS on the inverse problem.}
    \end{subfigure} \hfill
    \begin{subfigure}[b]{0.6\textwidth}
        \includegraphics[height=3.5cm]{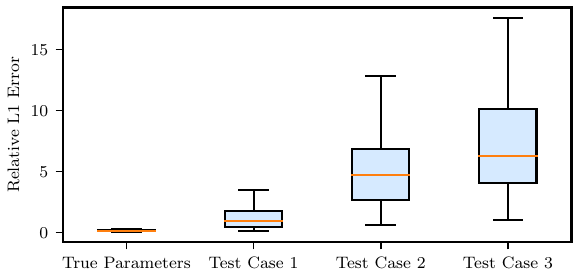}
        \caption{$L^1$ error on the forward and unified problem.}
        \label{fig:pwp box plot l1 errors}
    \end{subfigure}
    \caption{PWP: Box plots of the errors of the forward and inverse components of FUSE, as reported in Table \ref{tab:results inv for both}, for different levels of available input information ("test cases"). For the inverse problem, parameter samples $\xi_i$ are sampled from $\rho^{\phi}(\xi^i | u)$. For the forward problem, the output function $s$ is predicted based on the true parameter values $\xi^* \sim \rho(\xi | u)$. For the unified problem evaluating both the inverse and forward model parts, the mean $\Bar{s}$ of the ensemble prediction $s_i$ from inferred parameters $\xi_i \sim \rho^{\phi}(\xi^i | u)$ is compared to the true output time series $s$. As information is removed from the input in the different cases, it becomes more difficult to estimate $s$.}
    \label{fig:pwp box plot errors}
\end{figure}

\begin{figure}
	\includegraphics[width=\linewidth]{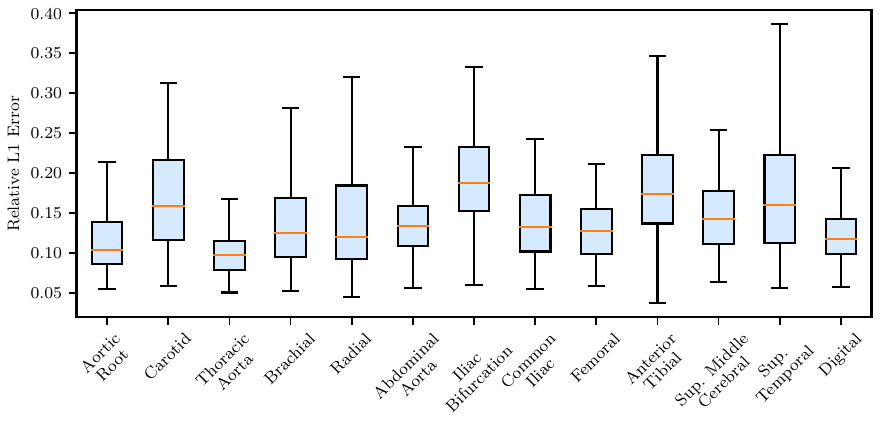}
     \caption{PWP, propagated uncertainty: Relative $L^1$ Error per vessel predicting pressure time series from the true parameters.}
     \label{fig:pwp true parameters box plot}
\end{figure}

\begin{figure}[htb]
  \hfill
  \rotatebox{90}{%
    \begin{minipage}{0.6\linewidth}
    	\includegraphics[width=\textwidth]{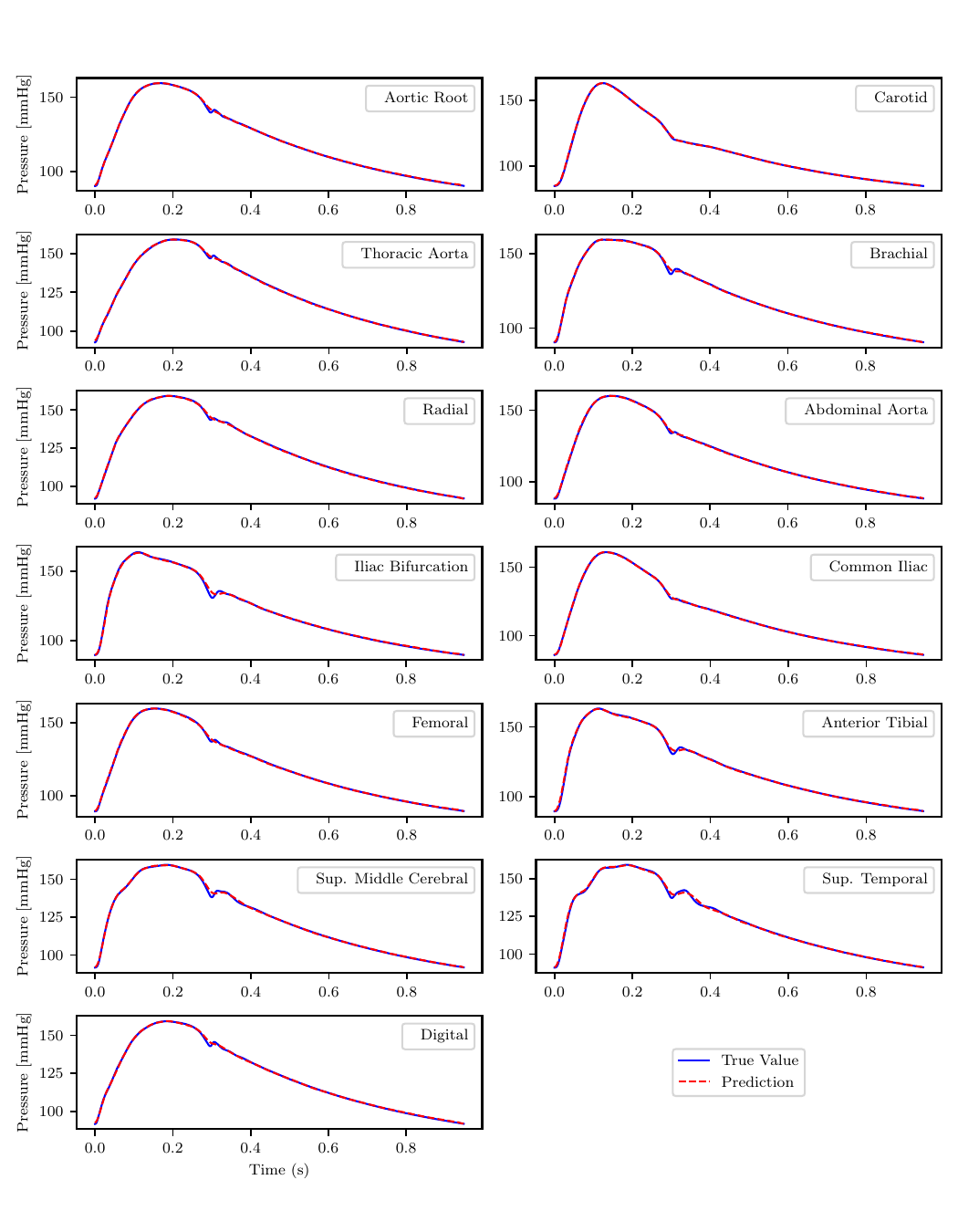}
        \caption{PWP, forward problem: Predictions with the largest $L^1$ error using the true parameters of the cardiovascular model.}
      \label{fig:pwp worst case predictions from true parameters}
    \end{minipage}
  }\hfill \strut
\end{figure}

\begin{figure}[htb]
  \hfill
  \rotatebox{90}{%
    \begin{minipage}{0.6\linewidth}
    	\includegraphics[width=\textwidth]{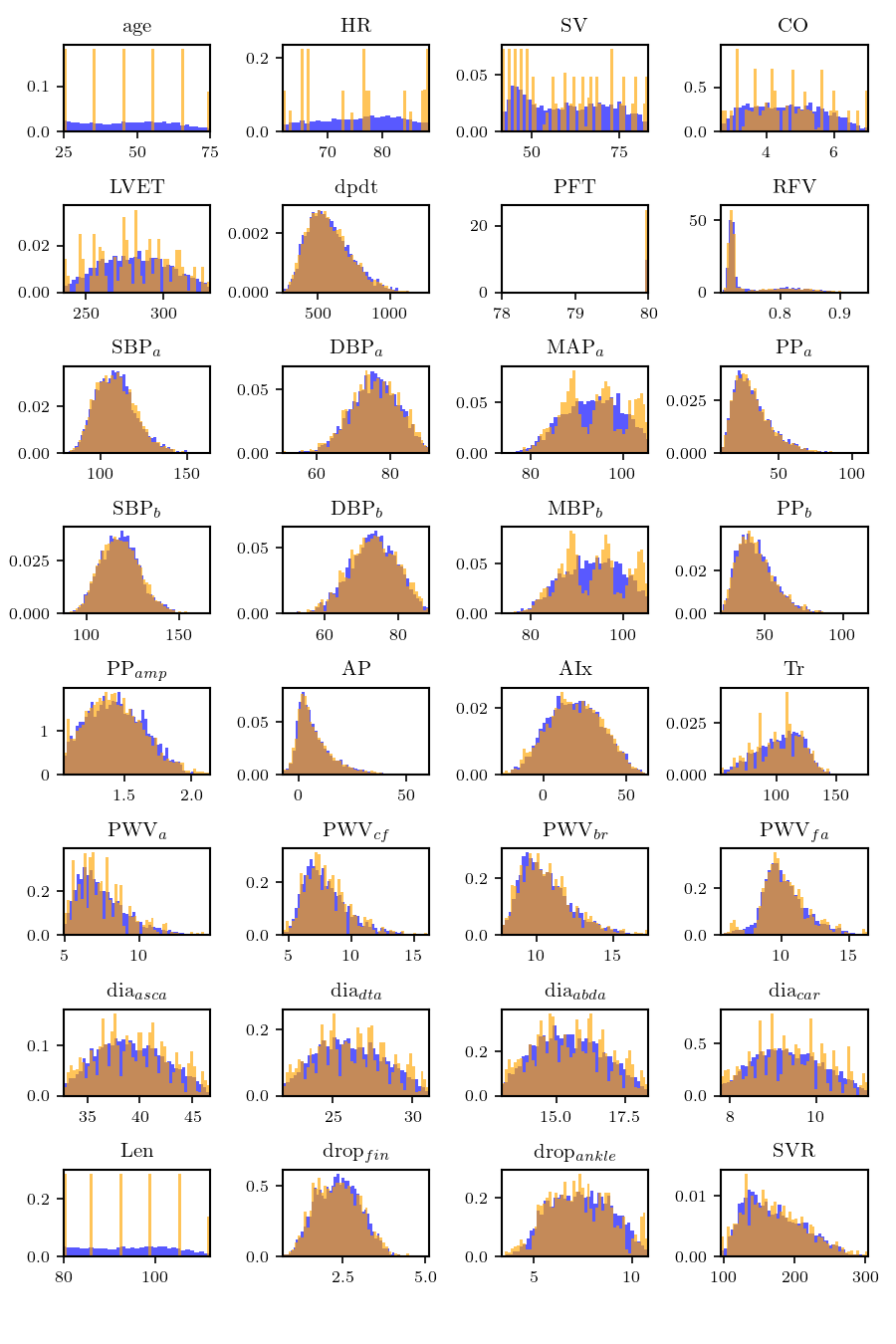}
        \caption{PWP, inverse problem: In the case of a fully masked input (all waveforms are zero), FUSE returns distributions which closely match the prior distributions. Shown in blue are parameter samples for a completely masked input, and in orange the prior distributions.}
      \label{fig:prior vs predicted}
    \end{minipage}
  }\hfill \strut
\end{figure}

\begin{figure}[htb]
  \hfill
  \rotatebox{90}{%
    \begin{minipage}{0.6\linewidth}
        \includegraphics[width=0.9\textwidth]{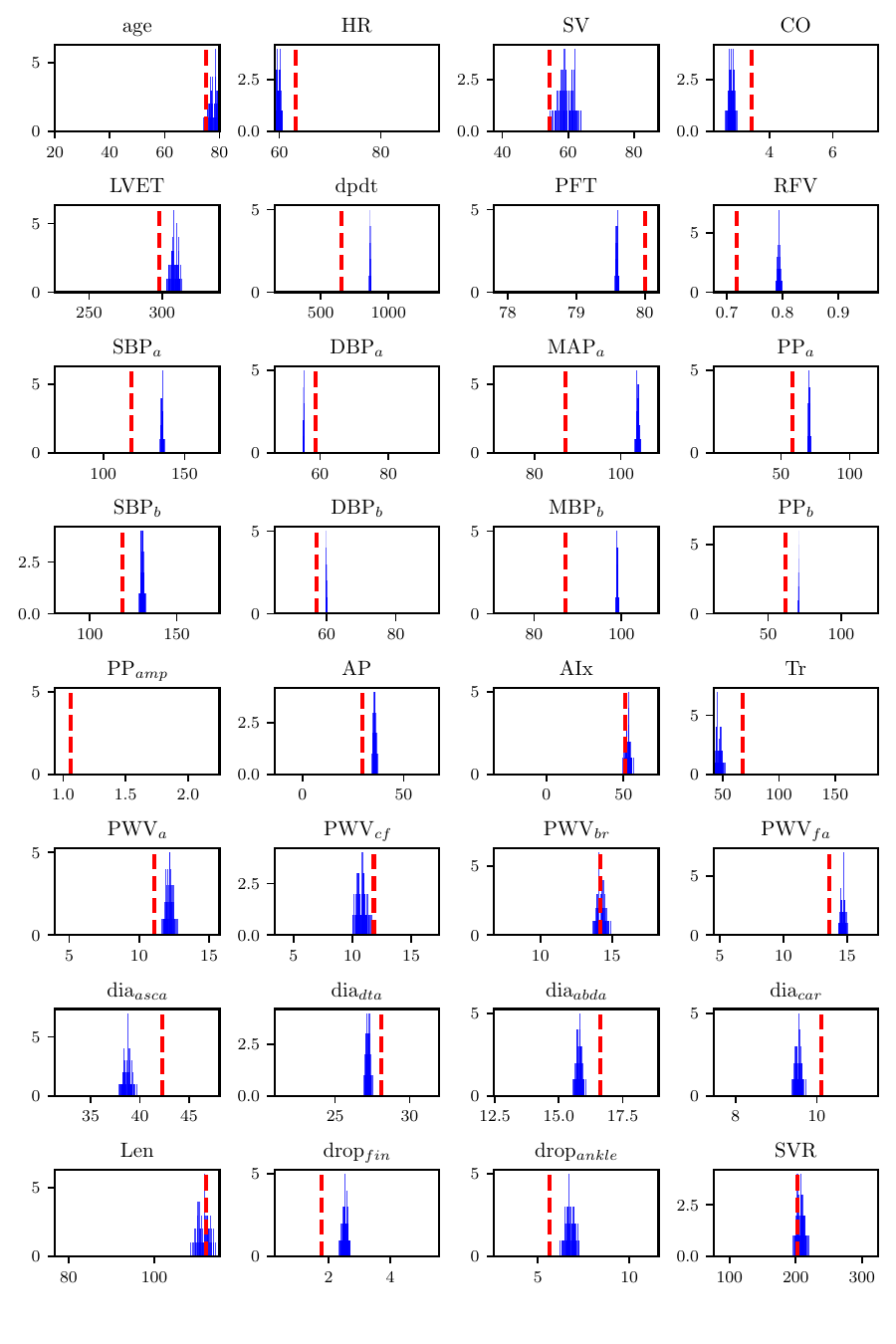}
        \caption{PWP, inverse problem: InVAErt predictions over parameters, the median sample with ful input information (level 1). InVAErt is particularly susceptible to posterior collapse, i.e., the predictions over the parameters become overconfident and the uncertainty bounds fail to capture the true parameter.}
      \label{fig:invaert posterior collapse}
    \end{minipage}
  }\hfill \strut
\end{figure}

\begin{figure}
    \centering
    \begin{subfigure}{0.45\textwidth}
        \includegraphics[width=\linewidth, trim = 0 560 245 0, clip]{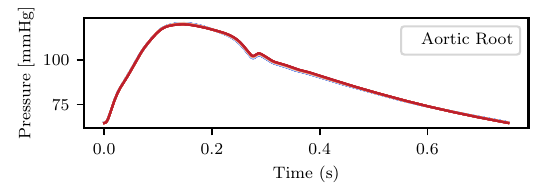}
        \caption{Age: 25 (blue) to 65 (red)}
    \end{subfigure}
    \hfill
    \begin{subfigure}{0.45\textwidth}
        \includegraphics[width=\linewidth, trim = 0 560 245 0, clip]{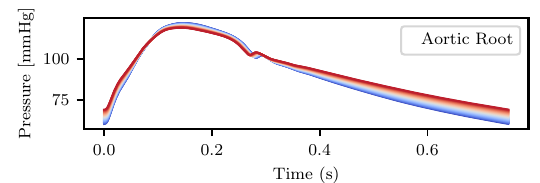}
        \caption{DBP: 55 (blue) to 90 (red) }
    \end{subfigure}
    
        \begin{subfigure}{0.45\textwidth}
        \includegraphics[width=\linewidth, trim = 0 560 245 0, clip]{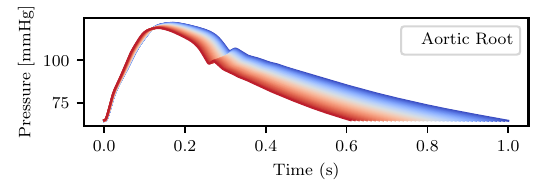}
        \caption{HR: 60 (blue) to 90 (red)}
    \end{subfigure}
    \hfill
    \begin{subfigure}{0.45\textwidth}
        \includegraphics[width=\linewidth, trim = 0 560 245 0, clip]{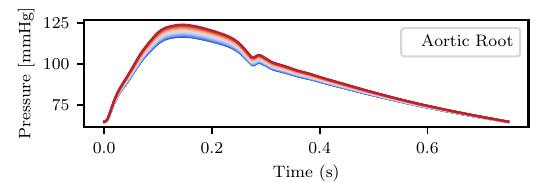}
        \caption{SBP: 80 (blue) to 165 (red) }
    \end{subfigure}
    
        \begin{subfigure}{0.45\textwidth}
        \includegraphics[width=\linewidth, trim = 0 555 245 0, clip]{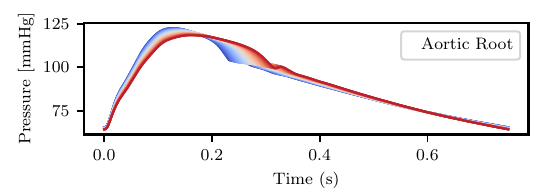}
        \caption{LVET: 230 (blue) to 360 (red)}
    \end{subfigure}
    \hfill
    \begin{subfigure}{0.45\textwidth}
        \includegraphics[width=\linewidth, trim = 0 560 245 0, clip]{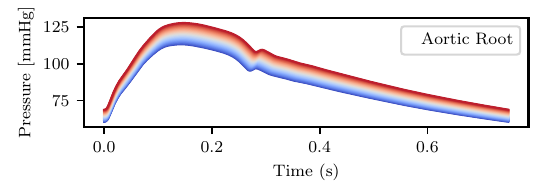}
        \caption{MAP: 75 (blue) to 125 (red)}
    \end{subfigure}
    \caption{PWP, sensitivity analysis: Pressure in the Aortic Root. We study how the pressure changes due to parameters of the system. Blue represents the lower bound of possible values, and red represents the upper bound, with colors of varying hue at intermediate values.}
    \label{fig:pwp fingerprints blood flow}
\end{figure}

\begin{figure}[htbp]
    \centering
        \includegraphics[width=0.8\textwidth]{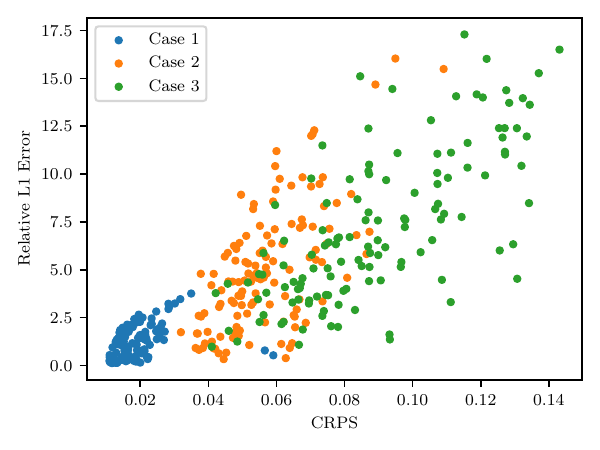}
    \caption{PWP: Correlation of the errors between continuous ($L^1$) and discrete (CRPS) parameters for different levels of available input information ("cases").}
    \label{fig:pwp l1 crps correlation}
\end{figure}

\clearpage

\subsection{Atmospheric Cold Bubble} \label{sec:appendix acb}

In atmospheric modelling, large-eddy simulations (LES) help to understand the complicated dynamics within convective storms \citep{orf2012} that are usually observed only by scalar local measurements of wind (anemometers), other scalar measurements obtained at surface-level weather stations (e.g., pressure, humidity), and, if available, three-dimensional wind lidar records.
A particularly destructive phenomenon associated with storms are \emph{downbursts}, sudden local downdrafts that cause extreme winds when hitting the surface. Manual tuning of a numerical model is usually required to fit a simulation to match an observed downburst (e.g., \cite{parodi2019}), due to the model complexity involving several parameterizations such as cloud microphysics. The model can then be configured to explore the sensitivity of the extreme winds both to the numerical model parameters and to the physical environmental conditions \citep{moore2024}. In addition to improving the scientific understanding and forecasts for more precise warnings, simulations of downbursts help to design robust infrastructure, such as wind turbines. 
However, the amount of simulations performed for model tuning (inverse problem) and sensitivity analysis (forward problem) is limited by the high computational cost to typically tens to hundreds of simulations. The affordable number of numerical simulations is either used for strategic expert tuning 
or to train an emulator for the parameter-to-data map, which is cheaper to evaluate and can then be used for in-depth sensitivity analysis (e.g., Gaussian Processes \cite{cleary2021}; quadratic regression \cite{liu2022}).
We would like to highlight that the presented test case is conceptually similar to downbursts, but far from simulating real-world scenarios due to its idealized nature and reduced complexity. However, we believe that test cases like this one will help to pave the way for handling realistic cases with methodologies such as FUSE. 

\paragraph{Model Description}

The numerical data for the cold bubble simulation is obtained with the PyCLES model \citep{pressel2015}. It solves the anelastic Navier-Stokes equations as formulated by \cite{pauluis2008}, evolving around a hydrostatic reference state,
\begin{equation}
    \begin{array}{ccc}
        \alpha_0 \frac{\partial p_0}{\partial x_3} &=& -g,
    \end{array}
\end{equation}
with reference profiles $\alpha_0(z)$ for specific volume and $p_0(z)$ for pressure, and $(x_1,x_2,x_3)=(x,y,z)$ and $(u_1,u_2,u_3)=(u,v,w)$ for simplicity.
The anelastic equations of motion are given by 
\begin{subequations}
\begin{empheq}[left=\empheqlbrace]{align}
    \frac{\partial u_i}{\partial t} + \frac{1}{\rho_0} \frac{\partial (\rho_0 u_i u_j)}{\partial x_i} 
    &=
    -\frac{\partial \alpha_0 p'}{\partial x_i} + b\delta_{13} 
    -\frac{1}{\rho_0} \frac{\partial (\rho_0 \tau_{ij})}{\partial x_j} + \Sigma_i,
    \\
    \frac{\partial s}{\partial t} + \frac{1}{\rho_0} \frac{\partial (\rho_0 u_i s)}{\partial x_i}
    &=
    \frac{Q}{T} 
    -\frac{1}{\rho_0} \frac{\partial (\rho_0 \gamma_{s,i})}{\partial x_i} + \Dot{S},  \\
    \frac{\partial \rho_0 u_i}{\partial x_i} &= 0,
\end{empheq}
\end{subequations}
encompassing the momentum equations (a), the entropy equation (b), and the continuity equation (c).
We use standard conventions for summing and the Kronecker delta $\delta_{ij}$, and left out the terms and further equations referring to the Coriolis force and moisture, as they do not apply to the small-scale and dry cold bubble simulation.
The buoyancy term $b$ depends on the specific volume, moisture, gravity, and the reference pressure profile, $p'$ is the dynamic pressure perturbation over the hydrostatic reference state, and $\Sigma_i$ is an additional momentum source term, which is expected to be small for the cold bubble experiment. In the entropy equation, $Q$ is the diabatic heating rate, $T$ is the temperature, and $\Dot{S}$ is a source term of irreversible entropy sources.
Finally, $\tau_{ij}$ and $\gamma_{s,i}$ are the sub-grid scale (SGS) stresses acting on velocity and entropy, which model the effect of turbulence smaller than the grid scale in a diffusive way as
\begin{equation}
    \tau_{ij} = -2\nu_t S_{ij},
\end{equation}
relating the sub-grid stress to the strain rate $S_{ij}=1/2 (\partial_i u_j + \partial_j u_i)$ of the resolved flow. The eddy viscosity $\nu_t$ may be related to the diffusivity $D_t$ for heat and other scalars $\phi$, and ultimately to the corresponding SGS stresses $\gamma_{\phi,i}$ by the turbulent Prandtl number $\mathrm{Pr}_t$ by
\begin{equation}
    \begin{array}{ccc}
        D_t &=& \nu_t / \mathrm{Pr}_t, \\
        \gamma_{\phi,i} &=& -D_t \frac{\partial \phi}{\partial x_i}.
    \end{array}
\end{equation}
For three-dimensional simulations, one may choose a sub-grid scale parameterization for $\nu_t$ such as a Smagorinsky-type first-order closure or a higher-order closure based on a prognostic equation for turbulent kinetic energy. For the two-dimensional cold bubble case, however, we stick to uniform values for both $\nu_t$ and $D_t$, independent of the turbulent Prandtl number.

\begin{figure}
    \centering
    \includegraphics[height=9cm, trim = 0 0 0 30, clip]{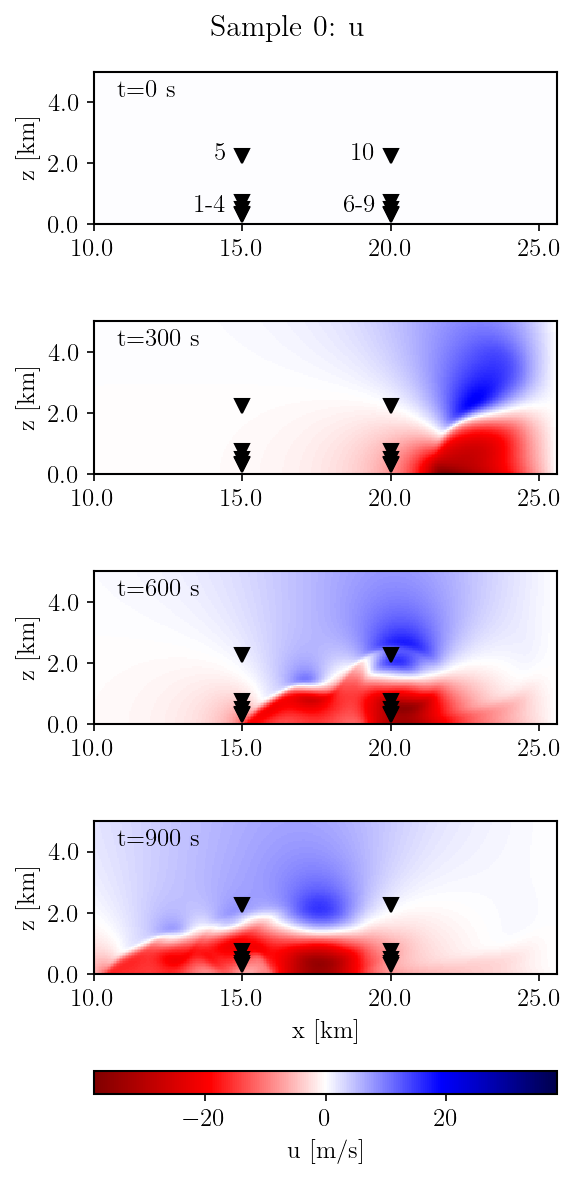}
    \includegraphics[height=9cm, trim = 20 0 0 30, clip]{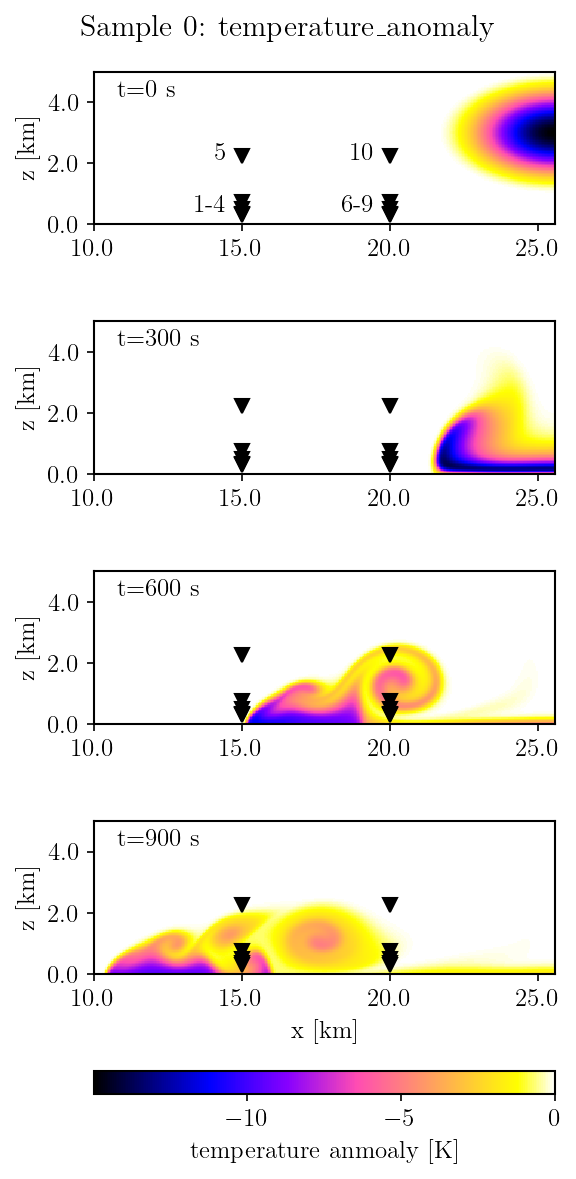}
    \includegraphics[height=9cm, trim = 20 0 0 30, clip]{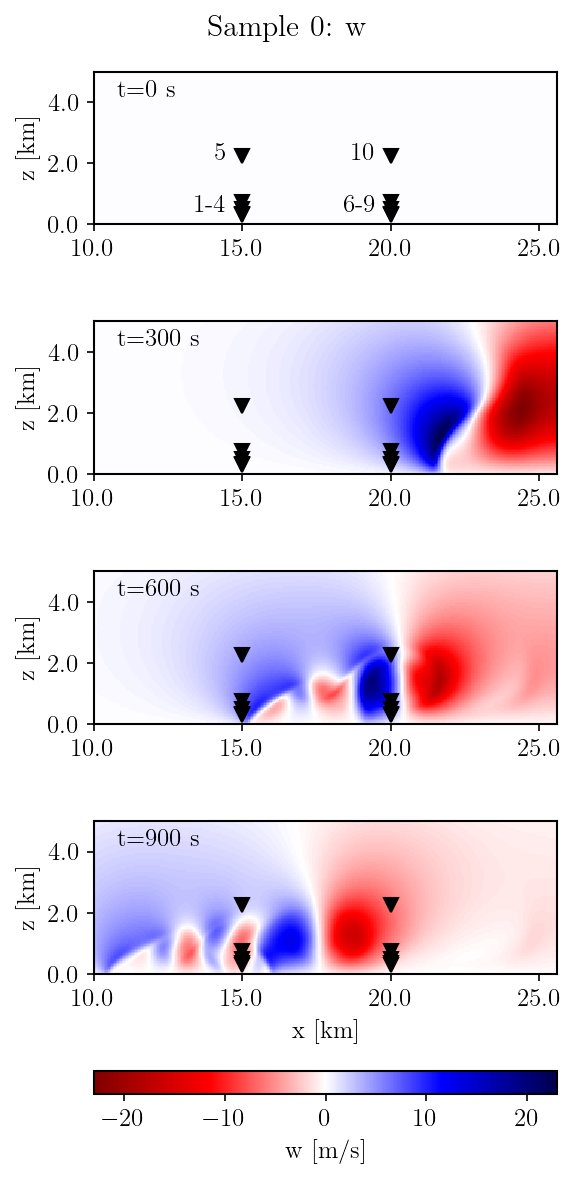}
    \caption{ACB: Time evolution of horizontal velocity (left), temperature anomaly (middle), and vertical velocity (right), for the default values of the parameters given in Table \ref{tab:tt params}, showing three main vortices that develop until $T=900$ s. Measurement locations for the time series data are marked with triangles, with their location indices indicated in the first row.}
    \label{fig:bubble_fields}
\end{figure}

The domain of size $(L_x,~L_z)=(51.6~\mathrm{km},~6.4~\mathrm{km})$ exhibits a neutrally stratified background profile ($\partial_z \theta =0$, where $\theta$ denotes potential temperature), and no background winds. An elliptic cold air anomaly is prescribed by
\begin{equation}
    \begin{array}{ccc}
        \Delta T &=& -a \left( \cos (\pi L) + 1\right)/2, \\[\medskipamount]
        L &=&  \left|\left| \left( (x-x_c)x_r^{-1},~ (z-z_c)z_r^{-1}\right) \right|\right|_2,
    \end{array}
    \label{eq:bubble}
\end{equation}
where default values and ranges for all parameters can be found in Table \ref{tab:tt params}, with the resulting flow fields shown in Figure \ref{fig:bubble_fields}. 

\paragraph{Numerical Setup}
The PyCLES model by \citep{pressel2015} relies on a WENO finite volumes scheme that exhibits exceptional stability, implemented in Cython. It is used here in the setup described by the authors, at resolution $\Delta x = \Delta z = 50$ m and an adaptive time stepping bound to $\Delta t \leq 5$ s in order to record time series measurements at a temporal resolution of five seconds.

The data is recorded at the eight locations marked in Figure \ref{fig:bubble_fields}, at horizontal locations $x=15,~20$ km (about $10$ and $5$ km from the center of the anomaly), and vertical heights of $z=50,~100,~250,~500,$ and $2000$ m, each being part of the numerical grid.

In terms of computational complexity, a PyCLES simulation in the given setup takes about half an hour on eight cores.

\begin{table}
    \centering
    \caption{ACB: Discrete parameters with their ranges used for uniform sampling of the training data, as well as default values used by \cite{straka1993}. The parameters either encode the initial condition (IC) or are part of the sub-grid scale (SGS) model (M). The horizontal location $x_c$ is kept fixed to ensure the horizontal symmetry of the domain.}
    \begin{tabular}{c l l c c c c}
    \toprule
        Parameter & Name             & Unit & Type & min & default     & max \\ \midrule
        $x_c$     & horizontal location& km & IC   & -   & \textbf{26.2} & -  \\
        $x_r$     & horizontal radius& km   & IC   & 2   & \textbf{4}  & 8  \\
        $z_c$     & vertical location& km   & IC   & 2.5 & \textbf{3}  & 3.5\\
        $z_r$     & vertical radius  & km   & IC   & 1   & \textbf{2}  & 2.5\\
        $a$       & amplitude        & K    & IC   & 5   & \textbf{15} & 25 \\
        $\nu_t$       & eddy viscosity    & m$^2$/s& M    & 0   & \textbf{75} & 75 \\
        $D_t$       & eddy diffusivity  & m$^2$/s& M    & 0   & \textbf{75} & 75 \\
    \bottomrule
    \end{tabular}
    \label{tab:tt params}
\end{table}

\clearpage

\begin{figure}
    \centering
    \includegraphics[height=9cm, trim = 0 0 0 30, clip]{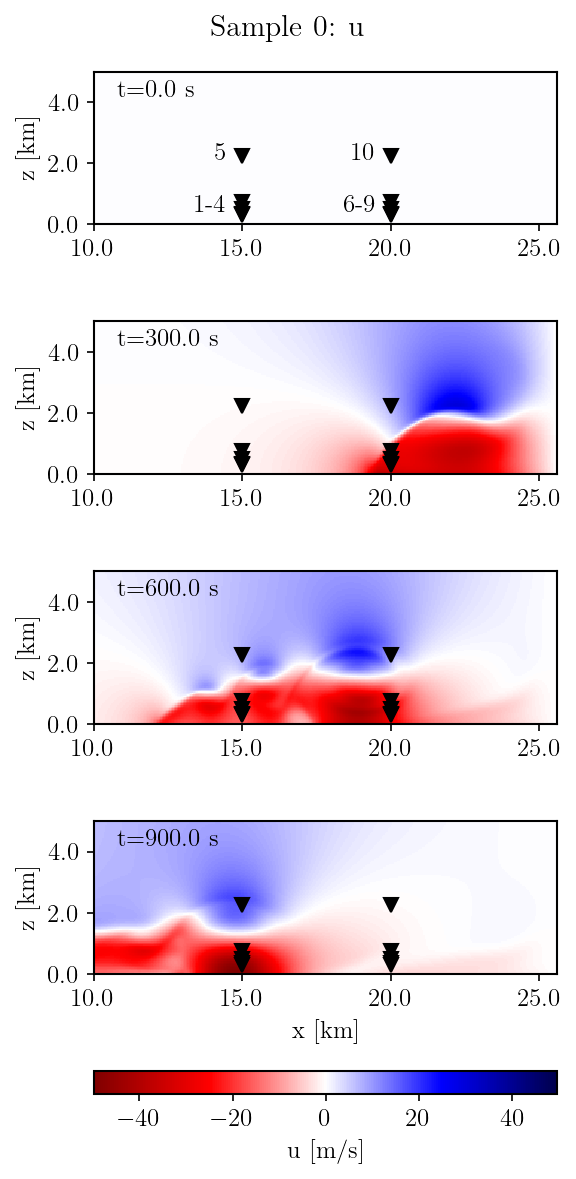}
    \includegraphics[height=9cm, trim = 20 0 0 30, clip]{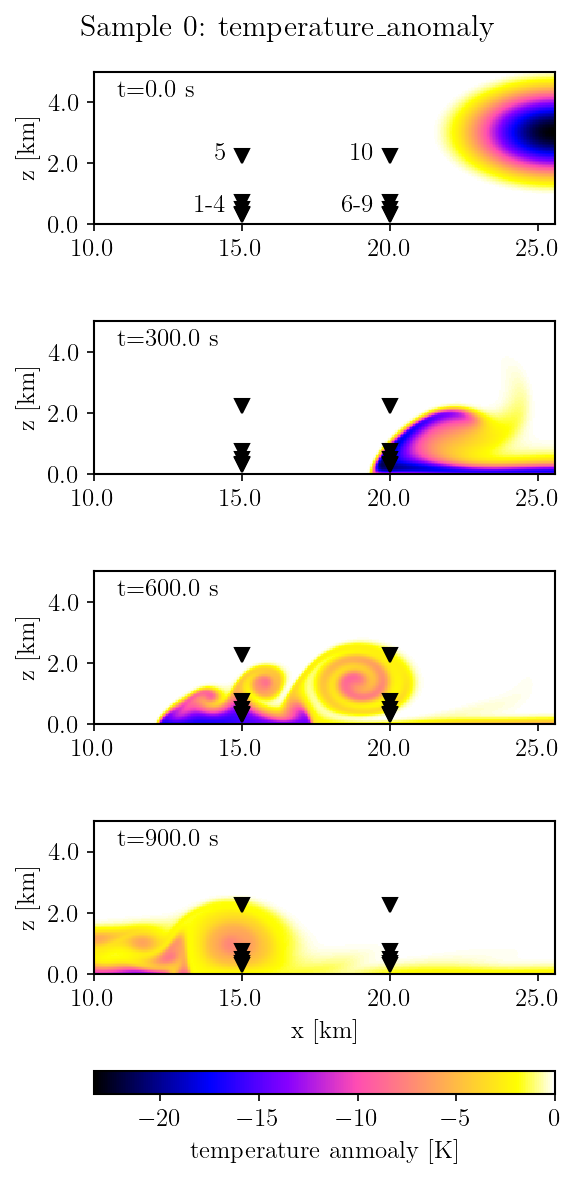}
    \includegraphics[height=9cm, trim = 20 0 0 30, clip]{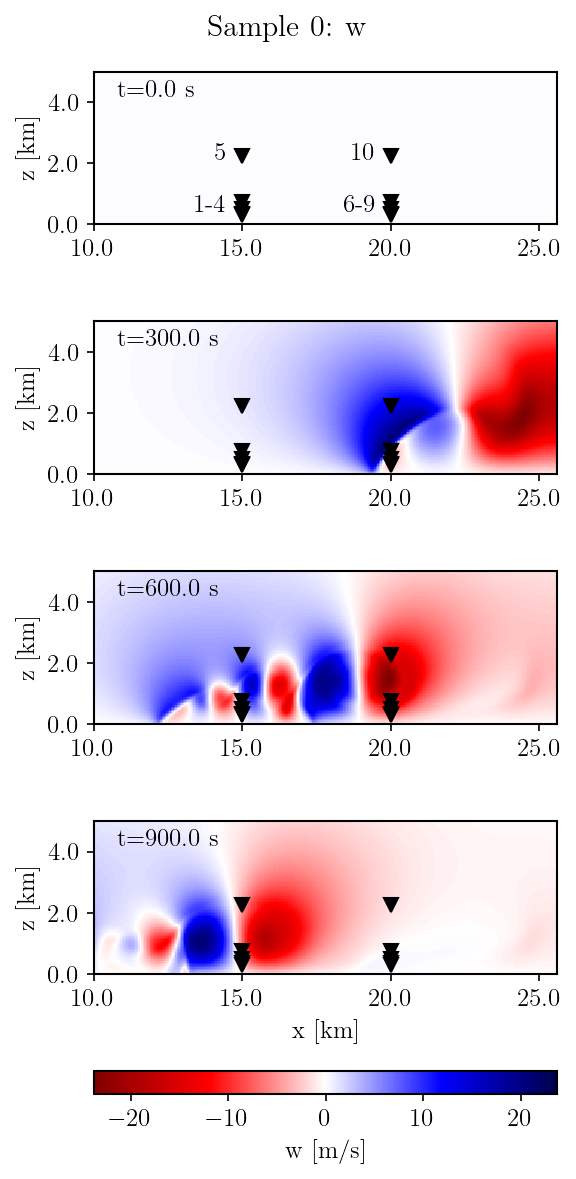}
    \caption{ACB: As Figure \ref{fig:bubble_fields}, for a sample with a stronger anomaly, reaching the state of three main vortices around $t=600$ s instead of $t=900$ s.}
    \label{fig:interesting_fields}
\end{figure}

\begin{sidewaysfigure}[ht]
    \centering
    \begin{subfigure}{0.45\linewidth}
    \includegraphics[width=\linewidth]{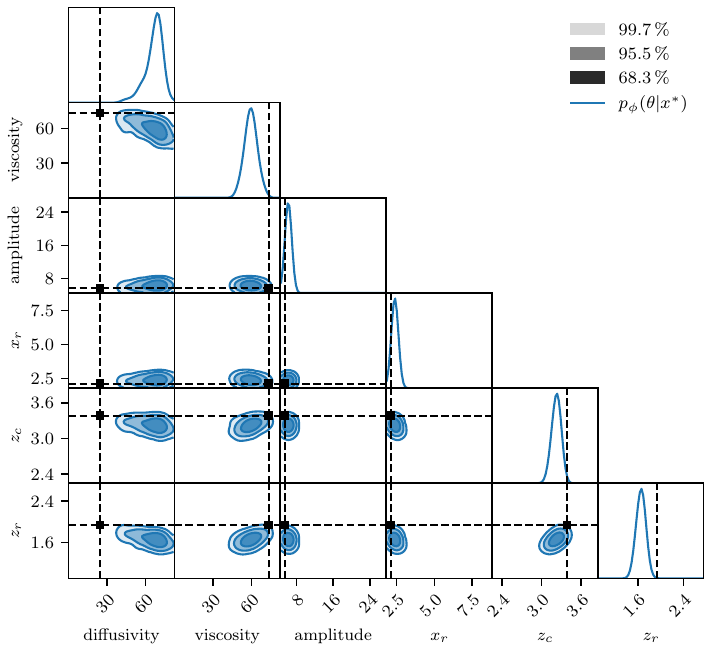}
        \caption{Parameter estimates from the test sample with greatest (worst) CRPS.}
    \end{subfigure} \hfill
    \begin{subfigure}{0.45\linewidth}
        \includegraphics[width=\linewidth]{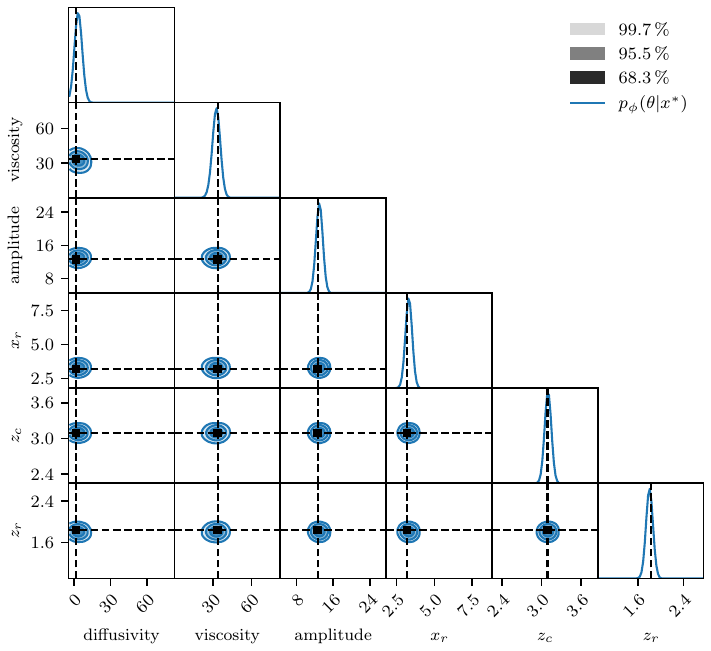}
        \caption{Parameter estimates from the test sample with median CRPS.}
    \end{subfigure}
    \caption{ACB, inverse problem: Samples drawn from the parameter distributions given continuous time series data. The true parameter values are marked with dashed lines, while the predicted distributions are given in blue shading.}
    \label{fig:acb worst med crps}
\end{sidewaysfigure}

\begin{sidewaysfigure}[ht]
    \centering
    \begin{subfigure}{0.3\linewidth}
        \includegraphics[width=\linewidth]{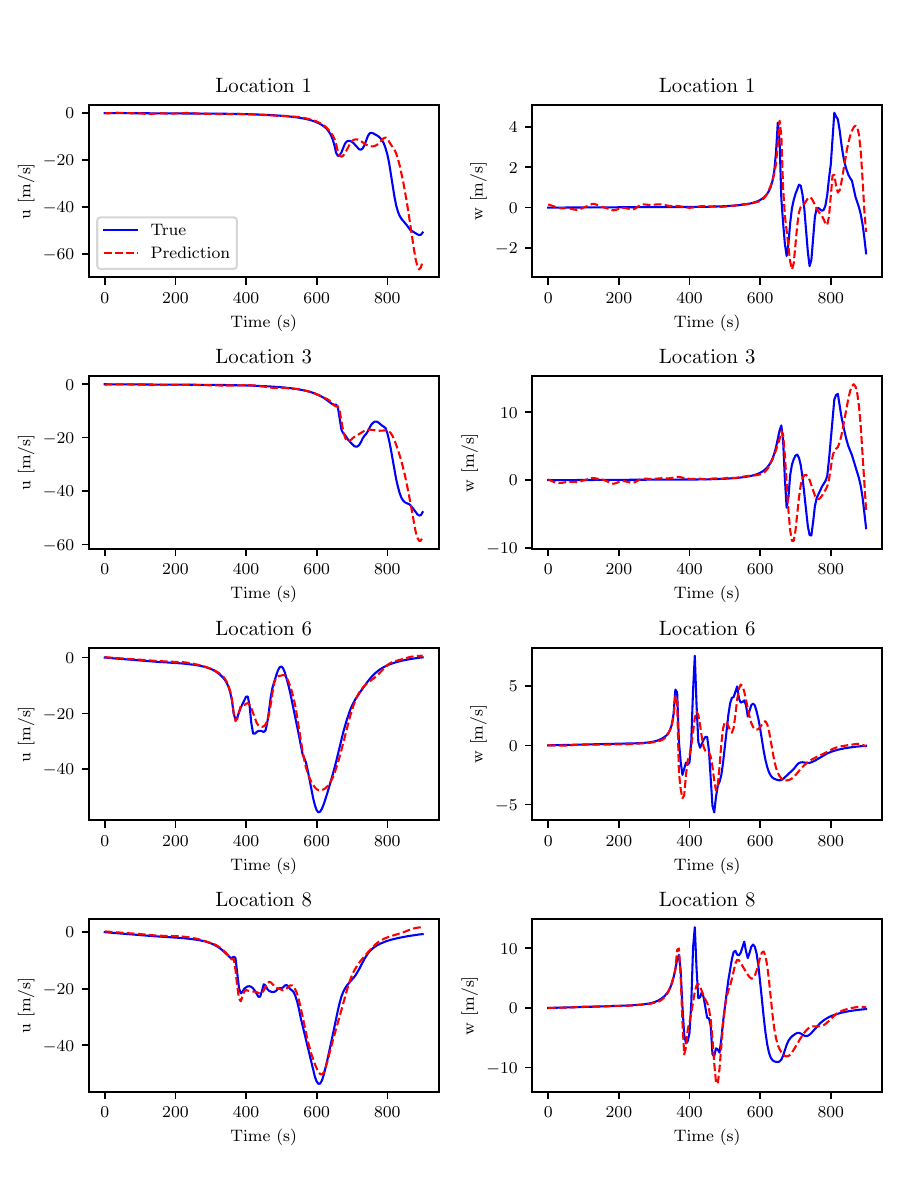}
        \caption{Sample with greatest (worst) error of FUSE.}
    \end{subfigure} \hfill
    \begin{subfigure}{0.3\linewidth}
        \includegraphics[width=\linewidth]{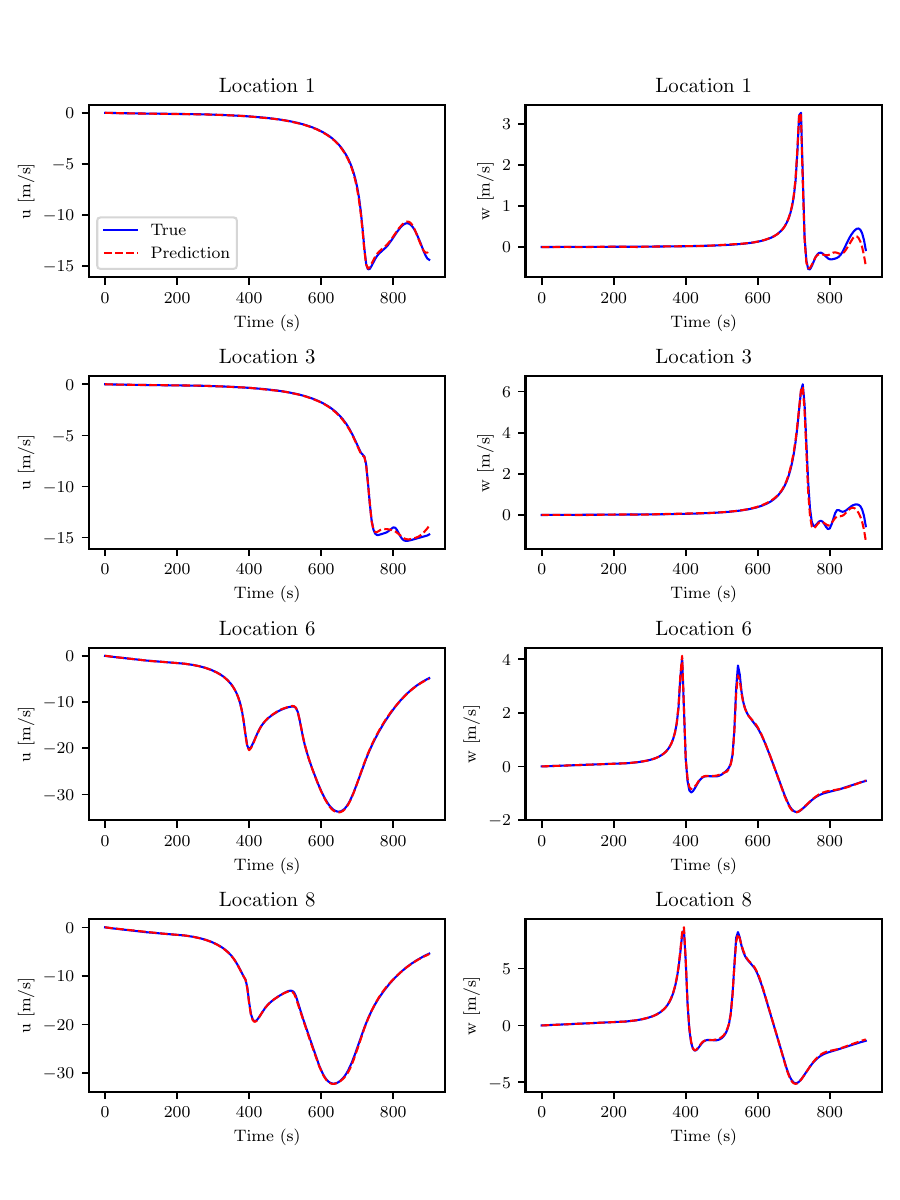}
        \caption{Sample with median error of FUSE.}
    \end{subfigure} \hfill
    \begin{subfigure}{0.3\linewidth}
        \includegraphics[width=\linewidth]{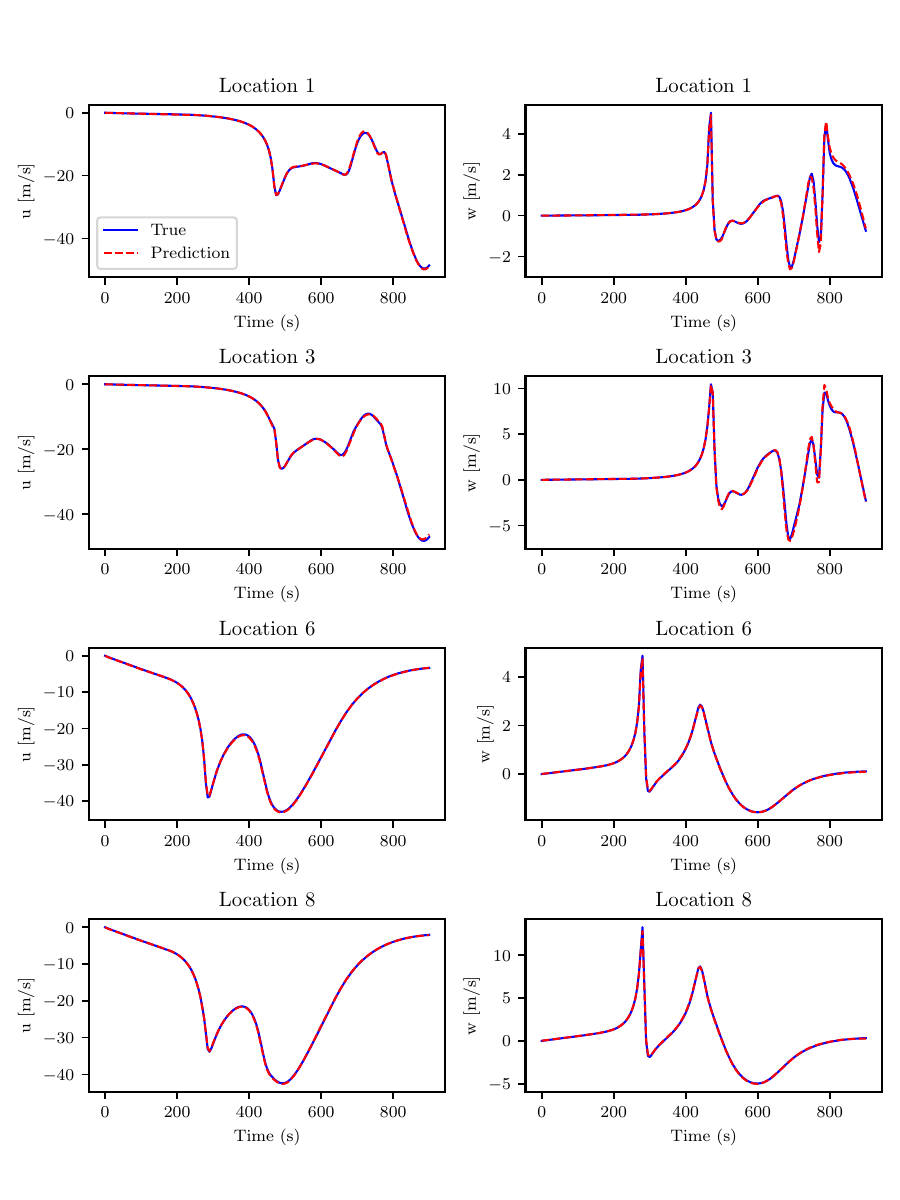}
        \caption{Sample with strong velocity gradients (Fig. \ref{fig:interesting_fields}).}
    \end{subfigure}
    \caption{ACB, forward problem: Velocity predictions with the FUSE model using the true parameters for different samples from the test set.}
    \label{fig:acb time series for different parameters}
\end{sidewaysfigure}

\begin{figure}[ht]
    \centering
    \begin{subfigure}{0.9\linewidth}
    \includegraphics[width=\linewidth]{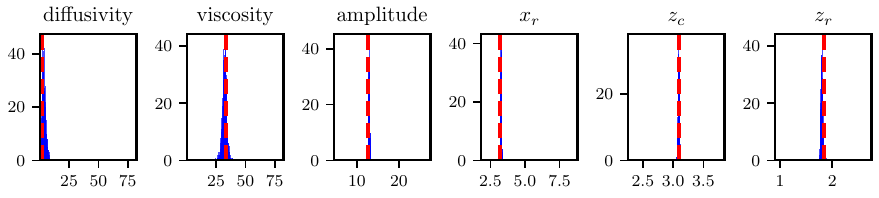}
        \caption{Sample on which FUSE has the median CRPS.}
        \label{fig:tt cd 1}
    \end{subfigure}
    
    \begin{subfigure}{0.9\linewidth}
        \includegraphics[width=\linewidth]{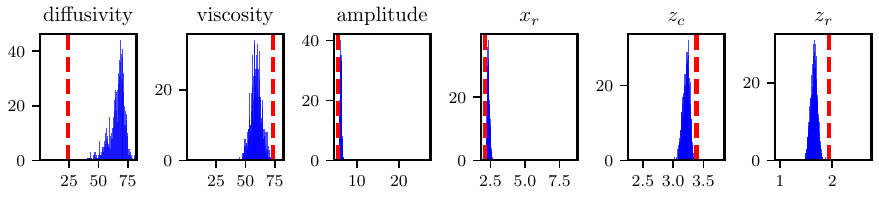}
        \caption{Sample on which FUSE has worst CRPS.}
        \label{fig:tt cd 2}
    \end{subfigure}
    \caption{ACB, inverse problem: Histograms of the parameters inferred from continuous time series measurements $u$, for different test samples. Given the very small amplitude and horizontal extend of the worst-case sample, this perturbation hardly reaches the sensor and the parameters are hence difficult to infer from the weak velocities measured.}
    \label{fig:acb worst med histograms}
\end{figure}

\begin{figure}[ht]
    \centering
        \begin{subfigure}{0.48\textwidth}
        \includegraphics[width=\linewidth]{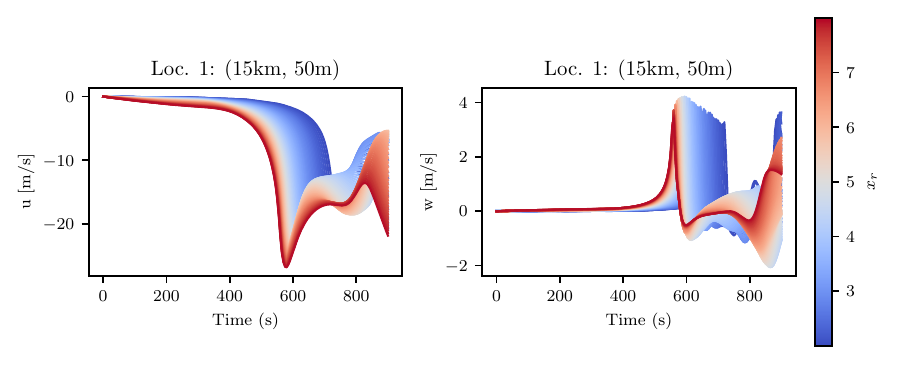}
        \caption{$x_r$}
    \end{subfigure}
    \hfill
    \begin{subfigure}{0.48\textwidth}
        \includegraphics[width=\linewidth]{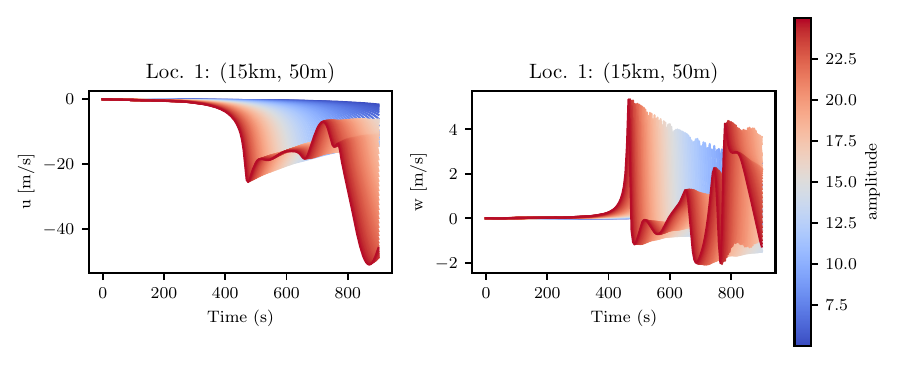}
        \caption{Amplitude}
    \end{subfigure}
    \begin{subfigure}{0.48\textwidth}
        \includegraphics[width=\linewidth]{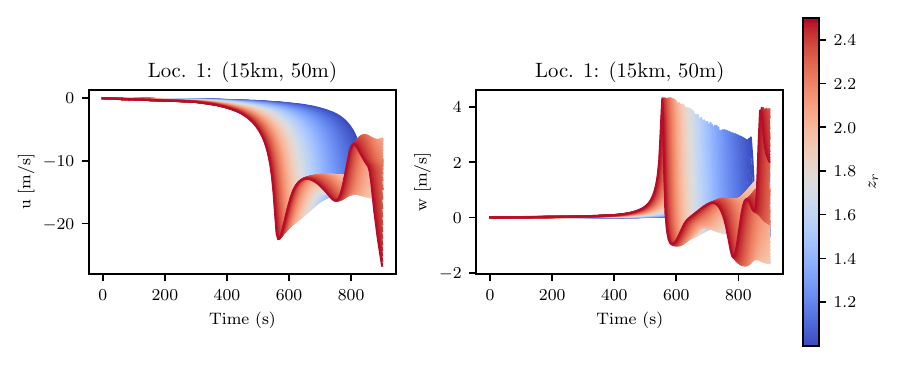}
        \caption{$z_r$}
    \end{subfigure}    
    \hfill
    \begin{subfigure}{0.48\textwidth}
        \includegraphics[width=\linewidth]{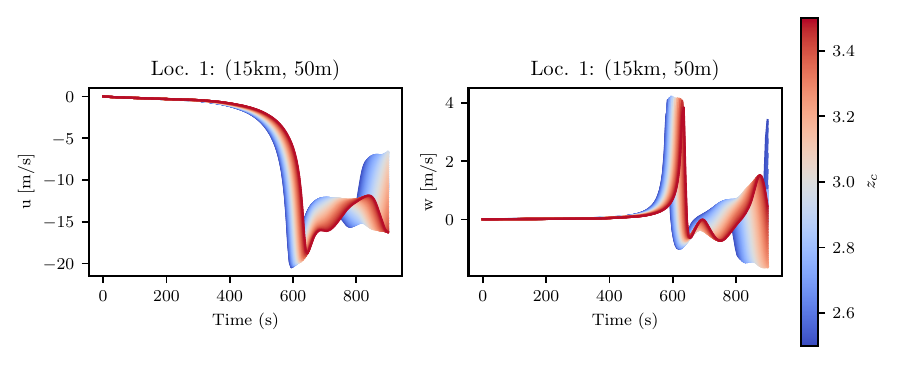}
        \caption{$z_c$}
    \end{subfigure}
    \begin{subfigure}{0.48\textwidth}
        \includegraphics[width=\linewidth]{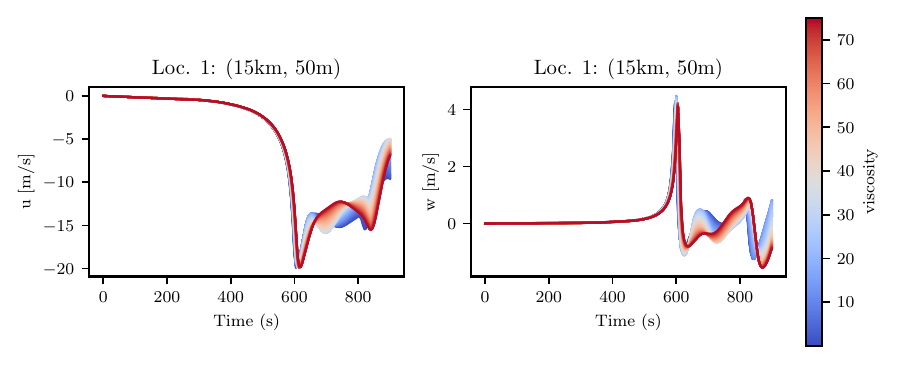}
        \caption{Viscosity}
    \end{subfigure}
    \hfill
    \begin{subfigure}{0.48\textwidth}
        \includegraphics[width=\linewidth]{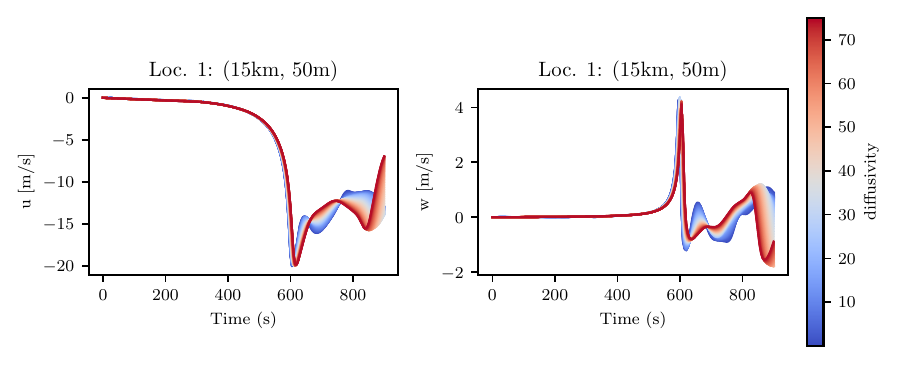}
        \caption{Diffusivity}
    \end{subfigure}
    
    \caption{ACB, sensitivity analysis: Fingerprints showing the model sensitivity to one parameter at a time at location 1, while keeping all others at their default value.}
    \label{fig:acb forward sensitivity 1}
    \label{fig:acb forward sensitivity}
\end{figure}

\begin{figure}[ht]
    \centering
    \begin{subfigure}{0.48\textwidth}
        \includegraphics[width=\linewidth]{images/results/TurbulenceTower/pairwise_validation/pairwise_u_max_loc0_error_rel.png}
    \end{subfigure}
    \hfill
    \begin{subfigure}{0.48\textwidth}
        \includegraphics[width=\linewidth]{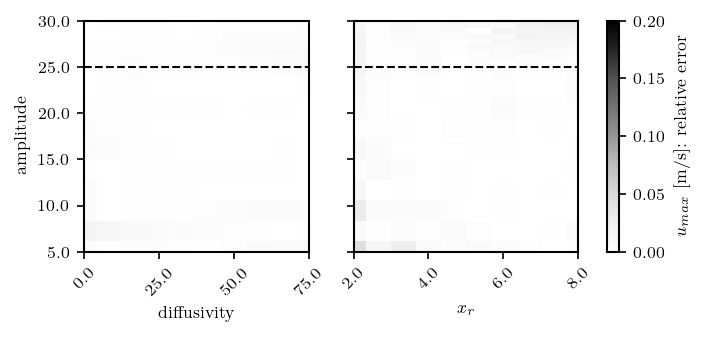}
    \end{subfigure}    

    \begin{subfigure}{0.48\textwidth}
        \includegraphics[width=\linewidth]{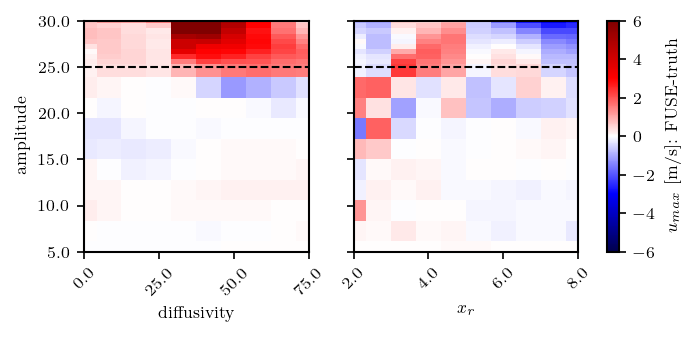}
    \end{subfigure}
    \hfill
    \begin{subfigure}{0.48\textwidth}
        \includegraphics[width=\linewidth]{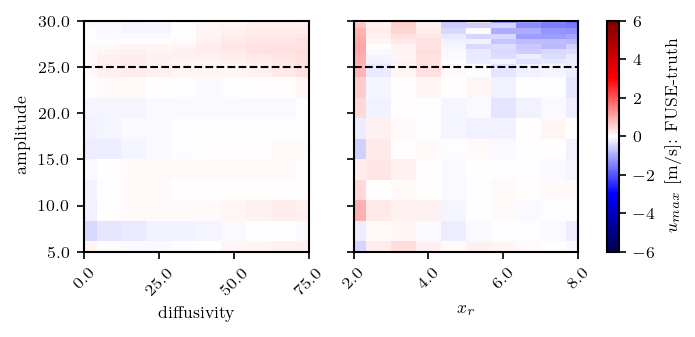}
    \end{subfigure}    

    \begin{subfigure}{0.48\textwidth}
        \includegraphics[width=\linewidth]{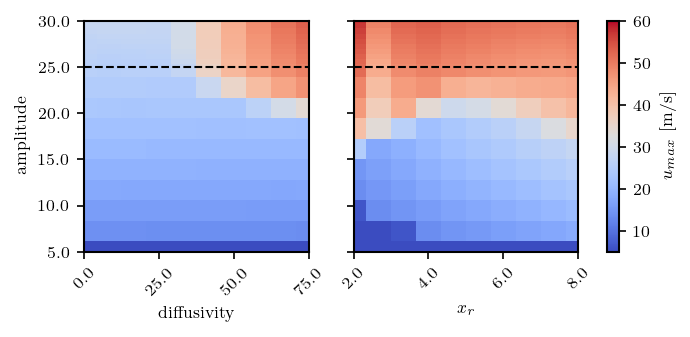}
    \end{subfigure}
    \hfill
    \begin{subfigure}{0.48\textwidth}
        \includegraphics[width=\linewidth]{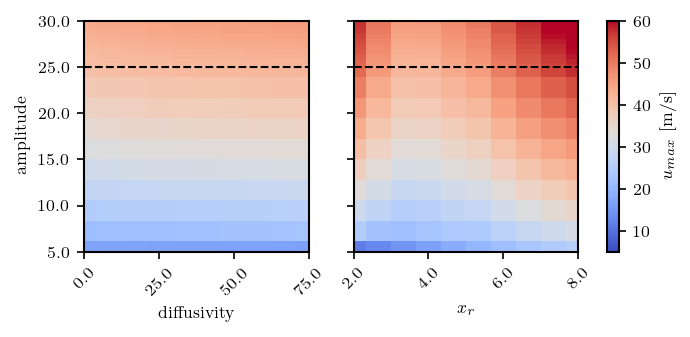}
    \end{subfigure}    
        
    \begin{subfigure}{0.48\textwidth}
        \includegraphics[width=\linewidth]{images/results/TurbulenceTower/pairwise_validation/pairwise_u_max_loc0_neural.png}
        \caption{Location 1  ($x=15$ km, $z=50$ m)}
    \end{subfigure}
    \hfill
    \begin{subfigure}{0.48\textwidth}
        \includegraphics[width=\linewidth]{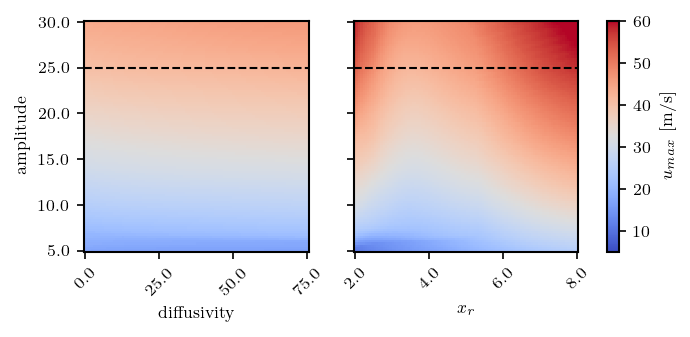}
        \caption{Location 5  ($x=20$ km, $z=50$ m)}
    \end{subfigure}    
    
    \caption{ACB, sensitivity analysis, continuation of Fig. \ref{fig:tt pariwise validation}: Validation of the FUSE model against numeric simulations on peak horizontal velocities $u$ at location 1 (left, further away from the perturbation center) and 5 (right, closer). From top to bottom: relative error between FUSE and the numerical model, difference between FUSE and the numerical model, maximum velocity calculated by the numerical model, maximum velocity calculated by FUSE.}
    \label{fig:acb ood}
\end{figure}

\begin{figure}[ht]
    \centering
    \begin{subfigure}{0.49\linewidth}
    \includegraphics[height=7cm, trim = 0 0 85 0, clip]{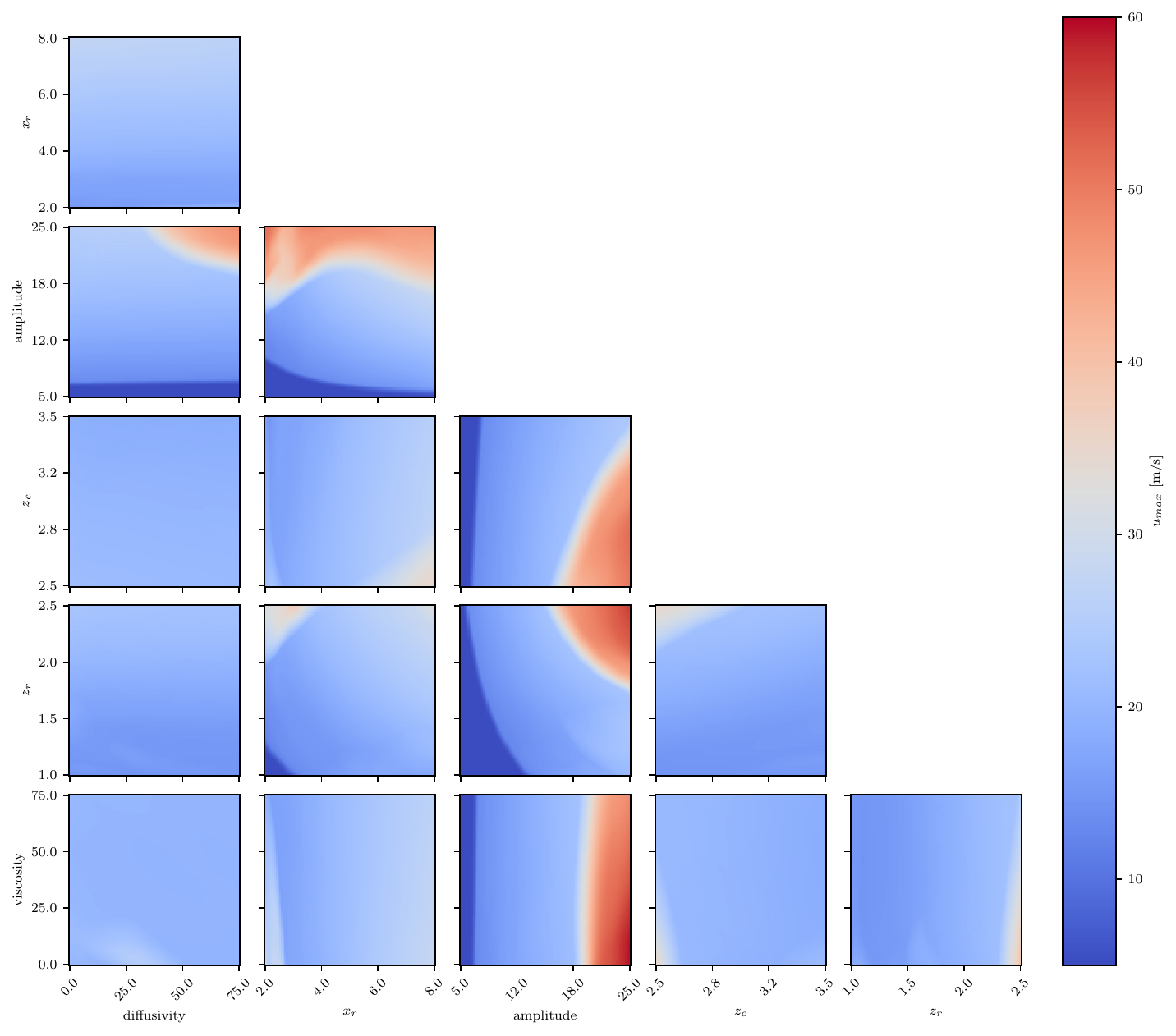}
        \caption{FUSE model.}
    \end{subfigure} \hfill
    \begin{subfigure}{0.49\linewidth}
    \includegraphics[height=7cm]{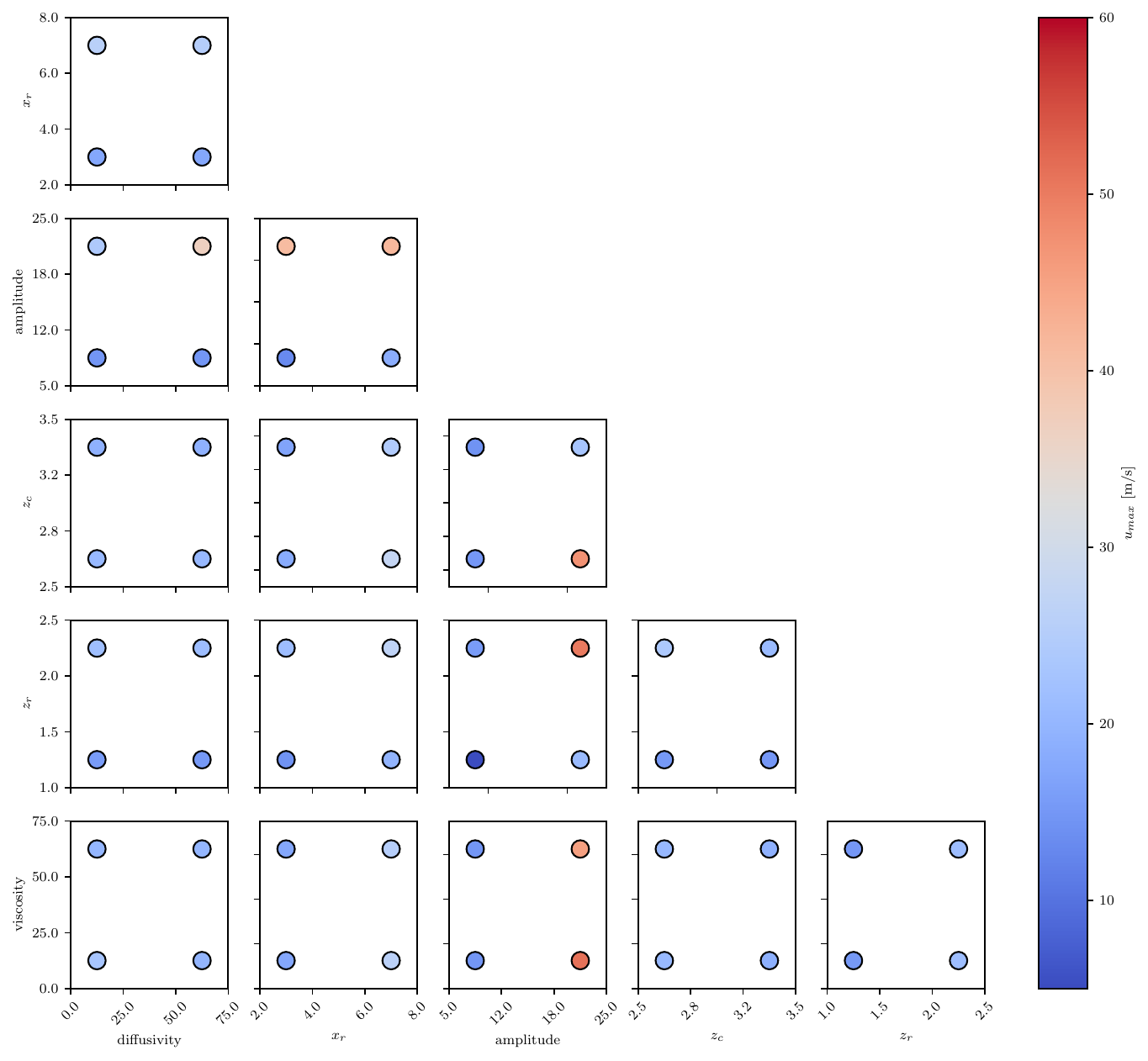}
        \caption{Numerical ground truth.}
    \end{subfigure}
    \caption{ACB, sensitivity analysis: Peak horizontal velocities $u$ at location 1 ($x=15$ km, $z=50$ m), sampled for pairwise combinations of the parameters, while keeping all others at their default value. For the neural model (a), $100$ samples are drawn for each parameter, corresponding to 150,000 evaluations. For the numerical model (b), four simulations were run per pair of parameters, with each taking values corresponding to $1/6$ and $5/6$ of the parameter range, corresponding to 60 model evaluations.
    }
    \label{fig:acb acb pairwise validation z50}
\end{figure}

\begin{figure}[ht]
    \centering
    \begin{subfigure}{0.49\linewidth}
    \includegraphics[height=7cm, trim = 0 0 85 0, clip]{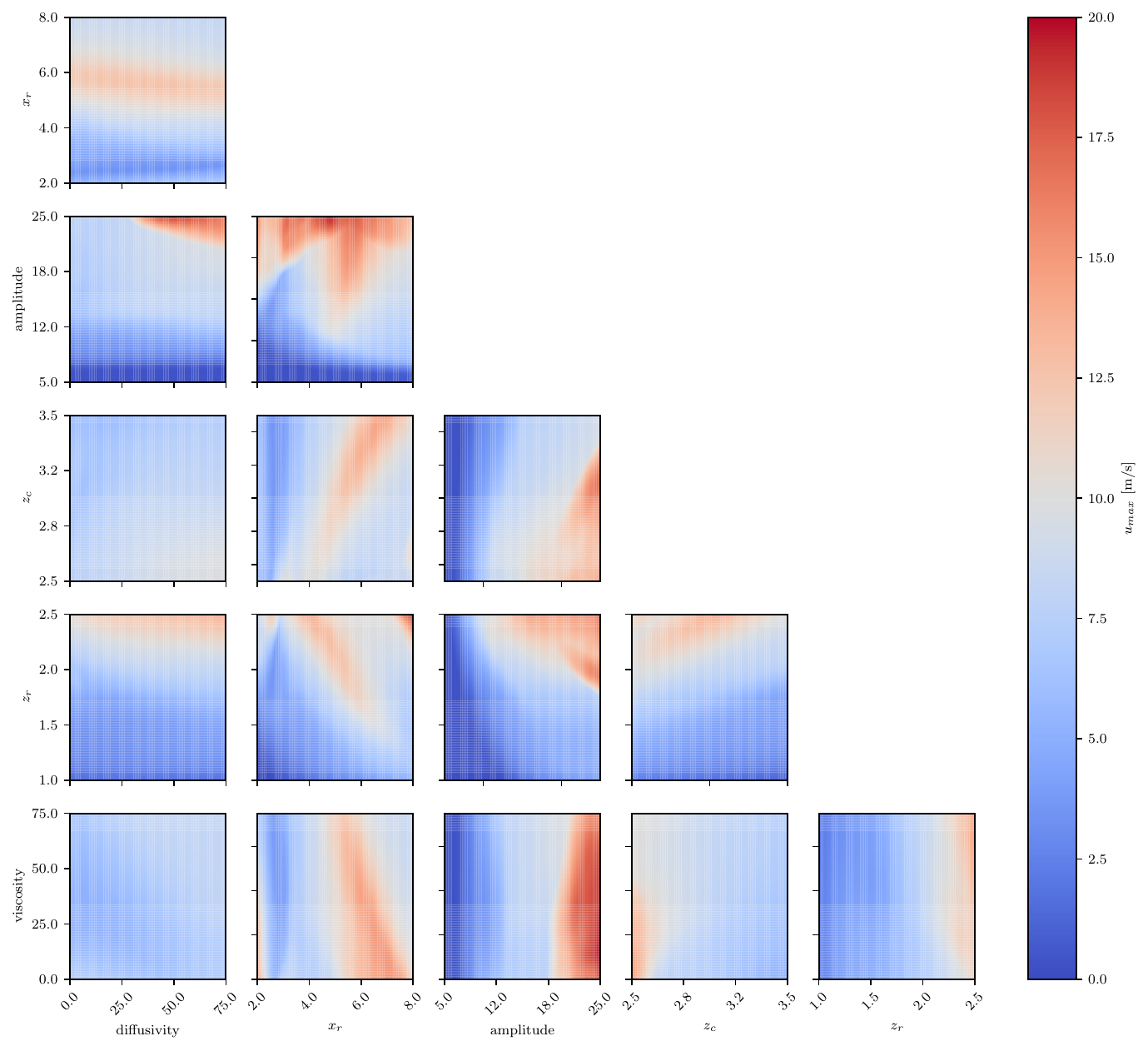}
        \caption{FUSE model.}
    \end{subfigure} \hfill
    \begin{subfigure}{0.49\linewidth}
    \includegraphics[height=7cm]{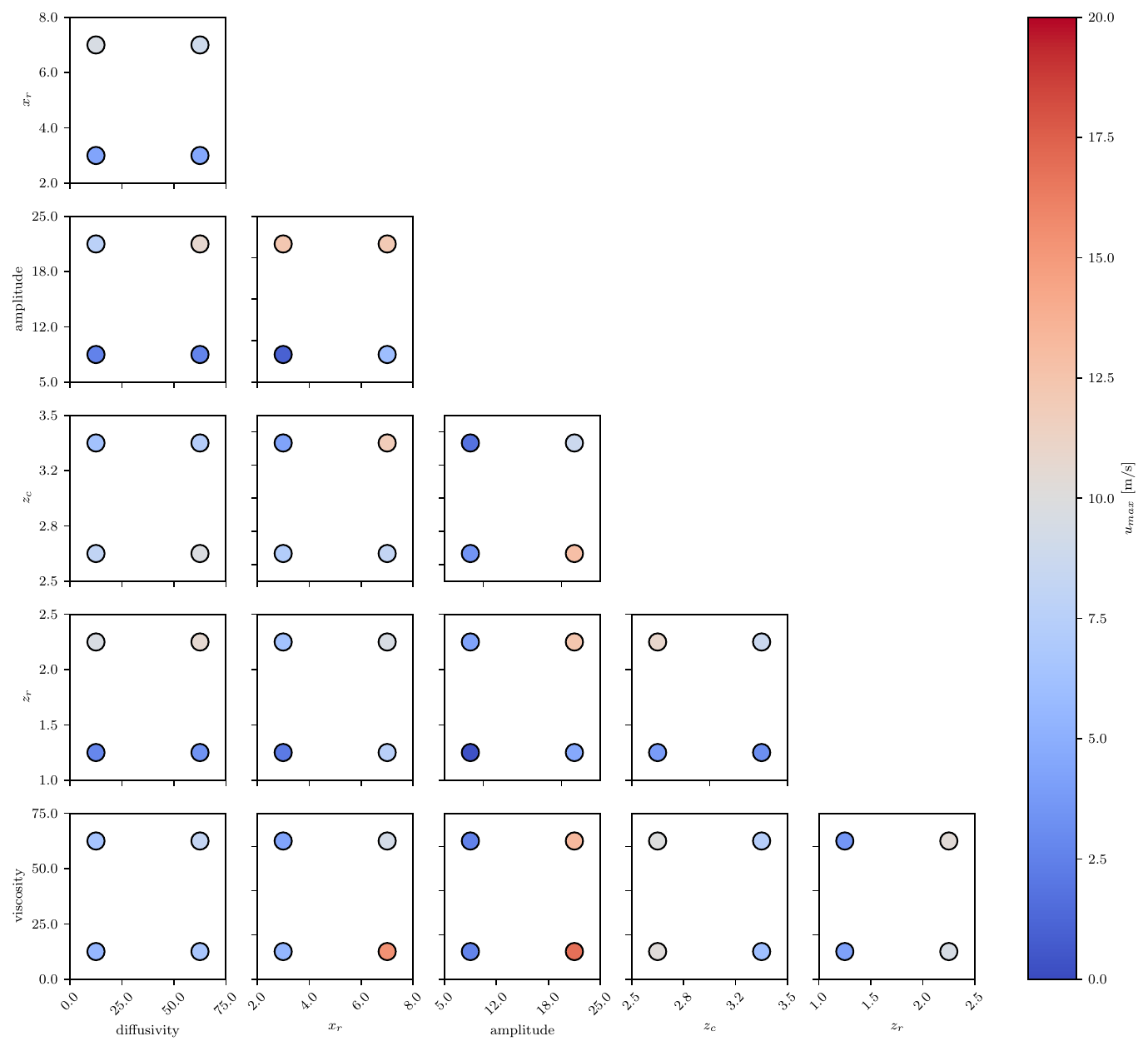}
        \caption{Numerical ground truth.}
    \end{subfigure}
    \caption{ACB, forward problem: Same as Figure \ref{fig:acb acb pairwise validation z50}, at location 5 ($x=20$ km, $z=2000$ m).}
    \label{fig:acb acb pairwise validation z2000}
\end{figure}

\begin{figure}[ht]
    \centering
    \begin{subfigure}{\linewidth}
        \includegraphics[width=\textwidth]{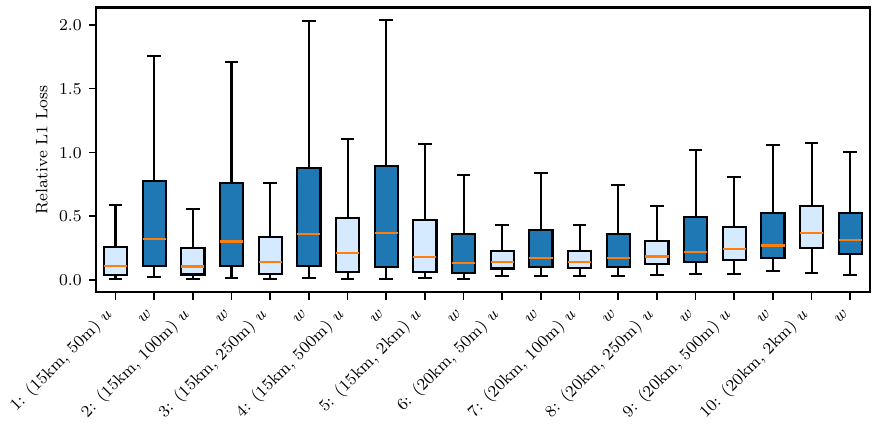}
        \caption{Forward problem: Error between the prediction of $s$ based on the true parameters $\xi^*$ and the true output time series $s$.}
        \label{fig:acb dc box plot errors each location}
    \end{subfigure}
    \begin{subfigure}{\linewidth}
        \includegraphics[width=\textwidth]{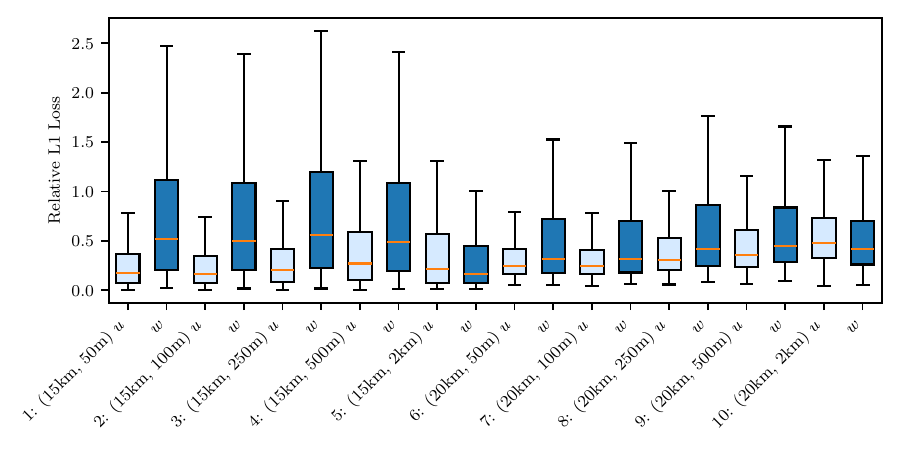}
        \caption{Unified prediction: Error between the predicted ensemble mean $\Bar{s}$ based on the input time series $u$, and the true output time series $s$.}
        \label{fig:acb cc box plot errors each location}
    \end{subfigure}
    \caption{ACB, propagated uncertainty: Box plots of relative errors in the time series predictions at each location.}
\end{figure}

\clearpage

\subsection{Properties of the Total Variation Metric}
\label{sec:total variation inf equals disc}
Consider the total variation distance $d(\mu, \mu^*$), where $\mu$ and $\mu^*$ are measures over $u \in \mathcal{U}$, which is defined as
\begin{equation}
    \begin{split}
    d (\mu, \mu^*) &= \sup_{A \subseteq \mathcal{U}} | \mu(A) - \mu^*(A)|, \\
                   &= \sup_{A \subseteq \mathcal{U}} | \int_{\mathcal{U}} \mu(u) du - \int_{\mathcal{U}} \mu^*(u) du|.
    \end{split}
\end{equation}
Further, consider a random vector $\xi \in \Xi \subseteq \mathbb{R}^m$, the function $h:\Xi \rightarrow \mathcal{U}$, and the probability measures $\rho$ and $\rho^*$ over $\Xi$. Since $u=h(\xi)$, the measures $\mu$ and $\mu^*$ can be considered as the pushforward measures induced by $\rho$ through the function $h$, $\mu = h_{\#\rho}$ and $\mu^* = h_{\#\rho^*}$. 
We assume all measures admit densities and use the same notation for measures $\mu(A),~A\subseteq \mathcal{U},$ and their densities $\mu(u),~u\in\mathcal{U},$ since the distinction is clear from the arguments.

We consider the conditional probability measures $\rho(\xi | u), \rho^*(\xi | u)$. Then, $\mu$ and $\mu^*$ are defined as:

\begin{equation*}
    \mu(u) = \int_{\Xi} \rho(\xi) \rho(\xi|u) d\xi, \quad \mu^*(u) = \int_{\Xi} \rho^*(\xi) \rho^*(\xi|u) d\xi,
\end{equation*}
and the measure of set $A$ as:
\begin{equation*}
    \begin{split}
        \mu(A) &= \int_A \mu(u) du = \int_A \int_{\Xi} \rho(\xi) \rho(\xi|u) d\xi du \\
        \mu^*(A) &= \int_A \mu^*(u) du = \int_A \int_{\Xi} \rho^*(\xi) \rho^*(\xi|u) d\xi du  
    \end{split}
\end{equation*}

Assume $u$ is uniquely determined by $\xi$
through the map $u = h(\xi)$ and that $\rho(\xi|u)$ is hence highly concentrated at $\xi = h^{-1}(u)$ and therefore:

\begin{equation*}
    \begin{split}
        \rho(\xi | u) &= \delta (\xi - h^{-1}(u)) \\
        \rho^*(\xi | u) &= \delta (\xi - h^{-1}(u)) 
    \end{split}
\end{equation*}
where $\delta$ is the Dirac delta function and 
\begin{equation*}
    \begin{split}
        \mu(u)  &= \int_\Xi \rho(\xi) \delta (\xi - h^{-1}(u)) d\xi = \rho(h^{-1}(u))\\
        \mu^*(u)  &= \int_\Xi \rho^*(\xi) \delta (\xi - h^{-1}(u)) d\xi = \rho^*(h^{-1}(u)).
    \end{split}
\end{equation*}

The definition of the total variation is then written as:
\begin{equation*}
    d ( \mu, \mu^*) = \sup_{A \subseteq \mathcal{U}} \left| \int_A \rho(h^{-1} (u))du - \int_A \rho^* (h^{-1} (u)) du \right|.
\end{equation*}

The terms $\rho(h^{-1}(u))$ and $\rho^*(h^{-1}(u))$ represent the pullback operators and therefore we can write:

\begin{equation}
    d (\mu, \mu^*) = \sup_{A \subseteq \mathcal{U}} \left| \rho(h^{-1} (A)) - \rho^*(h^{-1}(A)) \right| = d (\rho, \rho^*)
\end{equation}

Therefore due to the unique definition of $u$ from $\xi$ via $h$, the two definitions of the total variation are equivalent. The relation between the total variation of the original and the conditional measures is then given by:

\begin{equation}
   d (\mu, \mu^*)  =  d (\rho, \rho^*) \leq \int_{\mathcal{U}} d (\rho(\xi | u), \rho^*(\xi | u)) d \mu^* = d (\rho(\xi | u), \rho^*(\xi | u)) .
\end{equation}

\subsection{Detailed Derivation of the Forward Objective}
\label{sec:appendix derive L1}

We start with $d ( \Gcttilde_{\# \mu}, \Gctilde_{\# \mu})$ and our goal is to bound this metric by the supervised operator learning objective. For a function $u$, its corresponding images $\Gctilde(u)$ and $\Gcttilde(u)$, and for any set $A \in \mathcal{B}(\mathcal{S})$, where $\mathcal{B}$ is a Borel $\sigma$-algebra over $\mathcal{S}$:

\begin{equation*}
     \Gctilde_{\# \mu}(A) = \int_{\mathcal{S}} w_A(s) d \Gctilde_{\# \mu} (s), \quad \Gcttilde_{\# \mu} (A) = \int_{\mathcal{S}} w_A(s) d \Gcttilde_{\# \mu}(s),
\end{equation*}
where $w_A(s)$ is the indicator function that is one if $s\in A$ and zero otherwise.
Let $\mathcal{B}'(\mathcal{U}):=\Gctilde^{-1}(\mathcal{B}(\mathcal{S}))$ be the $\sigma$-algebra on $\mathcal{U}$ generated by the pre-images of sets in $\mathcal{B}(\mathcal{S})$ under $\Gctilde$. Then,

\begin{equation*}
\begin{split}
    d ( \Gctilde_{\# \mu}, \Gcttilde_{\# \mu}) &= \sup_{A \in \mathcal{B}(\mathcal{S})} \left| \int_{\mathcal{S}} w_A(s) d \Gcttilde_{\# \mu} (s) -  \int_{\mathcal{S}} w_A(s) d \Gcttilde_{\# \mu}(s) \right| \\
    &= \sup_{\Tilde{A} \in \mathcal{B}'(\mathcal{U})}  \left|\int_{\mathcal{U}} w_{\Tilde{A}}({\Gctilde}(u)) d \mu(u) -  \int_{\mathcal{U}} w_{\Tilde{A}}(\Gcttilde(u)) d \mu(u)\right|.
\end{split}
\end{equation*}

Further,
\begin{equation}
\label{eq:operator learning cont}
\begin{split}
    d ( \Gctilde_{\# \mu}, \Gct_{\# \mu}) &= \sup_{\Tilde{A} \in \mathcal{B}'(\mathcal{U})} \left| \int_{\mathcal{U}} ( w_{\Tilde{A}}({\Gctilde}(u)) - w_{\Tilde{A}}(\Gcttilde(u) ) d \mu(u) \right|\\
    &\leq \  \left| \int_{\mathcal{U}} ({\Gctilde}(u) - \Gcttilde(u)) d \mu(u) \right|\\
    &\leq \int_{\mathcal{U}} \| {\Gctilde}(u) - \Gcttilde(u)\|_{L^1(Y)} d \mu(u).
\end{split}
\end{equation}
Here, $\| \cdot \|_{L^1(Y)}$ is the $L^1$ norm over the space of function $\mathcal{S}$. We are interested in parametric PDEs where we have access to parameters $\xi \in \Xi \subset \mathbb{R}^m$, the true PDE parameters, sampled from $\rho$, a probability measure on $\mathbb{R}^m$. Operator learning involves a map between continuous functions, so an intermediate step needs to be considered to first lift the parameters $\xi$ to a continuous space before performing the action of the operator. For this purpose, we consider a function $h$ composed using two maps: $l: \mathbb{R}^m \rightarrow \mathcal{B}_{\Omega}$ and $T^{-1}: \mathcal{B}_{\Omega} \rightarrow \mathcal{U}$, the inverse Fourier transform. In this formulation, $l$ maps the parameters to $k$ to the Paley-Wiener space of band-limited functions in Fourier space, $\mathcal{B}_{\Omega} = \{ l \in L^2(\mathbb{R}): \text{supp}\ l \subset [- \Omega, \Omega] \}$, thus truncating the frequencies that can be used to represent the data. For this reason, we consider $\mu = h_{\# \rho(\xi|u)}$, the pushforward of $\rho(\xi|u)$ under $h$, and use a change of variables:
\begin{equation*}
    \int_{\mathcal{U}} f(u) d \mu(u) = \int_{\Xi} f(h(\xi)) d \rho(\xi | u),
\end{equation*}
for any integrable function $f$. We can apply this property to:
\begin{equation*}
\begin{split}
d ( \Gctilde_{\# \mu}, \Gcttilde_{\# \mu}) &= \sup_{\Tilde{A} \in \mathcal{B}'(\mathcal{U})} \left| \int_{\mathcal{U}} ( w_{\Tilde{A}}({\Gctilde}(u)) - w_{\Tilde{A}}({\Gctilde}^{\theta}(u) ) d \mu(u) \right|
\\ &= \sup_{\Tilde{A} \in \mathcal{B}'(\mathcal{U})}  \left| \int_{\Xi} ( w_{\Tilde{A}} (\Gctilde(h(\xi))) - w_{\Tilde{A}} (\Gcttilde(h(\xi))) d \rho(\xi) \right|.
\end{split}
\end{equation*}
We can define the operators $\Gc: \Xi \rightarrow \mathcal{U}$ and $\Gc^{\theta}: \Xi \rightarrow \mathcal{U}$ and write:
\begin{equation*}
\begin{split}
    d (  \Gctilde_{\# \mu}, \Gcttilde_{\# \mu}) &=  \sup_{\Tilde{A} \in \mathcal{B}'(\mathcal{U})}  \left| \int_{\Xi} ( w_{\Tilde{A}} (\Gc(\xi)) - w_{\Tilde{A}} (\Gc^{\theta}(\xi))) d \rho(\xi) \right| \\
    &\leq \int_{\Xi} \| \Gc(\xi) - \Gc^{\theta}(\xi)) \|_{L^1(Y)} d \rho(\xi)
\end{split}
\end{equation*}
Finally, we arrive at
\begin{equation}
    d ( \Tilde{\Gc}_{\# \mu}, \Tilde{\Gc}^{\theta}_{\# \mu}) \leq
      \int_{\Xi} \| \Gc(\xi) - \Gc^{\theta}(\xi)) \|_{L^1(Y)} d \rho(\xi) .
\end{equation}

\subsection{Detailed Derivation of the Inverse Objective}
\label{sec:appendix derive L2}

In the supervised operator learning problem, we have access to samples $u$ of functions but we do not have access to the distribution that generates $u$. Assume that $\Gcttilde$ is Lipschitz continuous: 
\begin{equation*}
     \| \Tilde{\Gc}^{\theta}(u) - \Tilde{\Gc}^{\theta}(v)\|_{L^1(Y)} \leq C \| u - v \|_{L^1(X)}.
\end{equation*}
Assuming that $\mu^\phi$ is an approximation of $\mu$, we can relate these two measures by the total variation of their respective pushforward measures under the forward surrogate $\Gc^\theta$ using the continuity properties of the operator (see Appendix \ref{sec:total variation inf equals disc}),
\begin{equation*}
\label{eq:total variation over continuous measures}
    d ( \Gcttilde_{\#\mu^\phi}, \Gcttilde_{\#\mu}) \leq C \ d(\mu^\phi, \mu),
\end{equation*}
for some constant $C\in\R$, which we generally assume to be equal to one.
This property holds because of the continuity of $\Gcttilde$ and because the total variation is preserved under continuous transformations. We assume that the functions $u$ are generated by solving a parametric PDE, where the parametric uncertainty over $\xi$ is propagated onto $u$ through the likelihood $\mu(u|\xi)$. Performing parameter estimation corresponds to sampling from a conditional distribution $\rho^\phi(\xi | u)$ approximating the posterior $\rho(\xi | u)$. To do so, we first bound the previous distance by the total variation between  $\rho^\phi(\xi|u)$ and $\rho(\xi |u)$ over $\Xi$ (see Appendix \ref{sec:total variation inf equals disc}):
\begin{equation}
\label{eq:bound of d}
     d(\mu^\phi, \mu) \leq \Tilde{d} (\rho^\phi(\xi | u), \rho(\xi | u)).
\end{equation}
Combining the two inequalities, we get
\begin{equation}
\label{eq:bound of d wrt rho}
     d ( \Gcttilde_{\#\mu^\phi}, \Gcttilde_{\#\mu}) \leq d (\rho^\phi(\xi | u), \rho(\xi | u)),
\end{equation}
which means that we can minimize the right hand side of the inequality \eqref{eq:bound of d wrt rho}.

\subsection{Formulation of the FMPE loss}
\label{sec:appendix FMPE}
For our implementation of Flow-Matching Posterior Estimation (FMPE), we closely follow \cite{dax2023flow}.
In the main text (Eqn. \ref{eq:inverse}), we formulated the FMPE objective as
\begin{equation*}
     d (\rho^\phi(\xi|u), \rho(\xi |u)),
\end{equation*}
for the total variation distance $d$ on measures over the parameter space $\Xi \subseteq \mathbb{R}^m$. We will minimize this distance in order to approximate the underlying distribution, $\rho^\phi \approx \rho$, and sample $M$ samples from $\rho^\phi(\xi|u)$ with the help of FMPE. We also defined the transform \eqref{eq:fmpe-conditional-info},
\begin{equation*}
    T^{\phi_1,\phi_2}(u) = \hat{u} = \Tilde{h} \circ \mathcal{K}^{\phi_1} \circ \mathcal{P}^{\phi_2} (u),
\end{equation*}
which transforms an infinite-dimensional input $u \in \mathcal{U}$ to a finite-dimensional conditional information $\hat{u} \in \mathbb{R}^k$ in Fourier space.

The FMPE method learns an invertible transform between the target distribution of parameters, conditional on the additional information, and a standard normal distribution. This transform is modelled as an ordinary differential equation with artificial time dimension $t \in [0,1]$ and solution $\psi_{t, \hat{u}}: \mathbb{R}^m \rightarrow \mathbb{R}^{m}$ as
\begin{equation}
    \frac{d}{dt} \psi_{t, \hat{u}} (\xi) = v_{t, \hat{u}} (\psi_{t, \hat{u}}(\xi)), \quad \psi_{0, \hat{u}}(\xi) = \xi_0,
\end{equation}
where $\xi_0$ follows a standard normal distribution. Trajectories in the artificial time $t$ can be obtained as $\xi_t = \psi_{t, \\hat{u}}(\xi)$. This way, the target probability $\rho(\xi | \hat{u})$ is approximated  as
\begin{equation*}
    \rho^\phi(\xi | \hat{u}) = (\psi_{1,\hat{u}})_{\# \rho_0} (\xi) = \rho_0(\xi) \exp \Big( - \int_0^1 \text{div} \ v_{t,\hat{u}} (\xi_t) dt \Big),
\end{equation*}
by solving the transport equation $\partial_t \rho_t + \text{div} (\rho_t v_{t,\hat{u}})=0$.

During training, $\psi_{0, \hat{u}_i}(\xi_i)$ is fit to resemble a standard normal on the training samples $\{\xi_i,\hat{u}_i\}_{i=1}^n$. During evaluation, a standard normal sample $\xi_0$ is drawn, and transformed to a sample that approximately follows $\rho^*(\xi |u)$ as
\begin{equation}
    \xi' = \psi_{\hat{u}_i, 0}(\xi_0) \sim \rho(\xi|u).
\end{equation}
In order to fully describe $\psi_{t, \hat{u}}$, it hence suffices to learn the trajectory velocity function $v_{t, \hat{u}}$. Let $p_t(\xi | \xi_1)$ be a sample-conditional Gaussian probability path with vector field $l_t(\xi_t | \xi_1)$, and let time be distributed as $t\sim p(t)$.  Then, \cite{dax2023flow} formulate the FMPE loss as
\begin{equation}
\label{eq:fmpe objective}
        \mathcal{L}^{FMPE}(\phi) = \mathbb{E}_{t\sim p(t), \xi \sim \rho^*(\xi | u), \hat{u} \sim p(\hat{u}|\xi_1), \xi_t \sim p_t(\xi_t| \xi_1)}||v^{\phi_0}_{t, T^{\phi_1,\phi_2}(u)}(\xi_t) - l_t(\xi_t|\xi)||_2^2.
\end{equation}

\clearpage

\end{document}